\documentclass[12pt]{elsarticle}
\usepackage[margin=1in]{geometry}
\AtBeginDocument{\hypersetup{linkcolor=Blue,urlcolor=Maroon,citecolor=Fuchsia}}
\usepackage{newtxtext} 
\biboptions{sort&compress}

\usepackage[colorlinks=true,hypertexnames=false]{hyperref}
\usepackage[utf8]{inputenc}
\usepackage{amsmath, amssymb, amsthm}
\usepackage[T1]{fontenc}
\usepackage{mathtools}
\usepackage{bm}
\usepackage[usenames,dvipsnames]{xcolor}
\usepackage[normalem]{ulem}
\usepackage{caption}
\usepackage{placeins}

\newcommand{\fref}[1]{\hyperref[#1]{Figure~\ref*{#1}}}
\newcommand{\Fref}[1]{\hyperref[#1]{Figure~\ref*{#1}}}
\newcommand{\sref}[1]{\hyperref[#1]{Section~\ref*{#1}}}
\newcommand{\Sref}[1]{\hyperref[#1]{Section~\ref*{#1}}}
\newcommand{\tref}[1]{\hyperref[#1]{Table~\ref*{#1}}}
\newcommand{\Tref}[1]{\hyperref[#1]{Table~\ref*{#1}}}
\newcommand{\aref}[1]{\hyperref[#1]{\ref*{#1}}}
\newcommand{\Aref}[1]{\hyperref[#1]{\ref*{#1}}}
\newcommand{\thref}[1]{\hyperref[#1]{Theorem~\ref*{#1}}}
\newcommand{\Thref}[1]{\hyperref[#1]{Theorem~\ref*{#1}}}

\usepackage{refcount}
\newcommand{\footref}[1]{\hyperref[#1]{\footnotemark[\getrefnumber{#1}]}}

\theoremstyle{definition}

\newtheorem{theorem}{Theorem}
\newtheorem*{theorem*}{Theorem}

\newtheorem*{lemma*}{Lemma}

\newtheorem*{corollary*}{Corollary}

\usepackage[ruled,vlined,linesnumbered]{algorithm2e}

\allowdisplaybreaks 

  \usepackage{array}
  \usepackage{longtable}
  \usepackage{booktabs}
  
  \newcolumntype{P}[1]{>{\centering\arraybackslash}p{#1}} 
  \newcolumntype{M}[1]{>{\centering\arraybackslash}m{#1}} 
  \newcolumntype{B}[1]{>{\centering\arraybackslash}b{#1}} 
  
  \usepackage{dcolumn}
  \newcolumntype{.}{D{.}{.}{-1}} 

\usepackage{tikz}
\usetikzlibrary{positioning}
\usetikzlibrary{shapes}
\usetikzlibrary{calc}

\makeatletter
\newcommand{\vast}{\bBigg@{3}}
\newcommand{\Vast}{\bBigg@{4}}
\makeatother

\renewcommand{\P}{{\mathcal{P}}}
\newcommand{\Q}{{\mathcal{Q}}}
\newcommand{\X}{{\mathcal{X}}}

\DeclareMathOperator*{\esssup}{ess\,sup}

\newcommand{\raindances}{\texttt{RainDancesVI}}
\newcommand{\raindancesurl}{https://prasanthcakewalk.gitlab.io/raindancesvi}

\journal{Econometrics and Statistics}

\begin{document}

\begin{frontmatter}



\title{\texorpdfstring{InClass Nets: Independent Classifier Networks for Nonparametric Estimation of Conditional Independence Mixture Models and Unsupervised Classification}{InClass Nets: Independent Classifier Networks for Nonparametric Estimation of Conditional Independence Mixture Models and Unsupervised Classification}}


\author{Konstantin T. Matchev}
\ead{matchev@ufl.edu}

\author{Prasanth Shyamsundar}
\ead{prasanths@ufl.edu}

\address{Institute for Fundamental Theory, Physics Department, University of Florida, Gainesville, FL 32611, USA}

\date{August 31, 2020}

\begin{abstract}
We introduce a new machine-learning-based approach, which we call the Independent Classifier networks (InClass nets) technique, for the \emph{nonparameteric} estimation of conditional independence mixture models (CIMMs). We approach the estimation of a CIMM as a multi-class classification problem, since dividing the dataset into different categories naturally leads to the estimation of the mixture model. InClass nets consist of multiple independent classifier neural networks (NNs), each of which handles one of the variates of the CIMM. Fitting the CIMM to the data is performed by simultaneously training the individual NNs using suitable cost functions. The ability of NNs to approximate arbitrary functions makes our technique nonparametric. Further leveraging the power of NNs, we allow the conditionally independent variates of the model to be individually high-dimensional, which is the main advantage of our technique over existing non-machine-learning-based approaches. We derive some new results on the nonparametric identifiability of bivariate CIMMs, in the form of a necessary and a (different) sufficient condition for a bivariate CIMM to be identifiable. We provide a public implementation of InClass nets as a Python package called \href{\raindancesurl}{\raindances{}} and validate our InClass nets technique with several worked out examples. Our method also has applications in unsupervised and semi-supervised classification problems.
\end{abstract}




\end{frontmatter}

\tableofcontents

\section{Introduction}

\subsection{Conditional Independence Mixture Models}
In many fields of science one encounters multivariate models which consist of several distinct sub-populations or components, say $C$ in number. Each component $i\in\{1,\dots,C\}$ has its own characteristic probability density function $f^{(i)}(\mathcal{X})$ of the relevant multi-dimensional variable $\X$. Such models are referred to as multivariate finite mixture models \cite{McLaghlan2019}, and the probability density of $\X$ under such a model is given by
\begin{equation}
 \P(\X) = \sum_{i=1}^C~w_i\,f^{(i)}(\X)\,,\quad \text{with} \quad \sum_{i=1}^C~w_i = 1\,,\quad \text{and} \quad w_i \geq 0~~~\forall i\in\{1,\dots,C\}\,, \label{eq:fmm}
\end{equation}
where the non-negative weights $w_i$ parameterize the mixing proportions of the individual components. An important special case of these finite mixture models is that of the conditional independence multivariate finite mixture models\footnote{In the literature, these models are also referred to as finite mixtures of product measures \cite{teicher1967}.} \cite{chauveau2015,Zhu2016}, which for brevity we will simply refer to as conditional independence mixture models (CIMMs). Under this special case, the variable $\X$ is parameterized using $V$ ``variates'' as $\X \equiv (x_1,\dots,x_V)$ such that for each component $i$, the density function $f^{(i)}$ factorizes into a product of distributions for the individual variates as
\begin{equation}
 f^{(i)}(\X) = \prod_{v=1}^V~f^{(i)}_v(x_v)\,,\qquad \forall i\in\{1,\dots,C\}\,, \label{eq:component_f}
\end{equation}
so that \eqref{eq:fmm} becomes
\begin{equation}
 \P(\X) = \sum_{i=1}^C~w_i~\prod_{v=1}^V~f^{(i)}_v(x_v)\,.\label{eq:cifmm}
\end{equation}
Here $f^{(i)}_v(x_v)$ is the unit-normalized probability density of $x_v$ within component $i$---the top index $(i)$ denotes the component and the bottom index $v$ denotes the variate. In our treatment, the individual variates $x_v$ are themselves allowed to be multi-dimensional. In other words, $V$ is not the dimensionality of the data, but rather the number of groups the attributes in $\X$ can be partitioned into so that they are (conditionally) independent of each other within the given component $i$ a datapoint belongs to. This is similar to the treatment, for example, in \cite{Hall2005,chauveau2015,CHAUVEAU20161}. In particular, the technique we develop below will be applicable in situations where the variates $x_v$ are high-dimensional (dim($x_v$) $\gg 1$). Unless otherwise stated, henceforth a ``mixture model'' shall refer to the conditional independence mixture model of \eqref{eq:cifmm}.

\subsubsection{Applications of Conditional Independence Mixture Models}

Conditional independence mixture models have applications in situations where the correlations and dependence between different variables in the data are explained in terms of a latent or hidden \emph{confounding} variable which influences the observed variables. This is referred to as Latent Structure Analysis (LSA) \cite{lazarsfeld_henry_1968,Andersen1982}. In particular, when the confounding variable is discrete or categorical, it can be interpreted as representing the \emph{class} a given datapoint belongs to. Such models are referred to as latent class models and their study and analysis is referred to as Latent Class Analysis (LCA) \cite{Clogg1995}.

The connection between CIMMs and LCMs can be seen in a straightforward manner as follows. We can sample a datapoint as per the mixture model in \eqref{eq:cifmm} by first generating the component index $i\in\{1,\dots,C\}$ as per the multinomial probability distribution induced by the weights $w_i$, and then sampling $(x_1,\dots,x_V)$ as per the distribution $f^{(i)}(\X)$ within component $i$. Now, the component index $i$ can be interpreted as the latent variable that explains the dependence between the random variates $\{x_1,\dots,x_V\}$ in the mixture.

CIMMs, LCA, and mixture models in general have applications in a wide range of fields, including econometrics \cite{Compiani2016,SpecialIssue2017}, social sciences
\cite{Vermunt2003,VermuntMagidson2004,Porcu2017,Petersen2019},
bioinformatics \cite{Yu2010}, 
astronomy and astrophysics
\cite{Nemec1991,Bovy:2011xm,Lee2012,Melchior:2016asy,Kuhn2017,Necib:2018iwb,Jones:2019njd}, high energy physics \cite{Stepanek:2015jqa,Cranmer:2015bka,Cranmer:2016swd}, and many others.

\subsubsection{Nonparametric Estimation of Conditional Independence Mixture Models}
Estimation of a CIMM is simply the process of estimating the weights $w_i$ and functions $f^{(i)}_v$ (assuming the number of components $C$ is known) from a dataset sampled from the joint distribution $\P(\X)$ of the variates under the mixture model, see \eqref{eq:cifmm}.

Under \emph{parametric} estimation of mixture models (conditionally independent or otherwise), one assumes that each of the distributions $f^{(i)}(\X)$ is from an appropriately chosen parametric class of distributions, e.g. multivariate Gaussians. The choice of the class of functions assumed to contain the true $f^{(i)}(\X)$ is informed by practitioner's prior knowledge of the problem at hand. This reduces the problem of estimating the mixture to the more tractable problem of estimating the weights $w_i$ and the parameter values corresponding to the true $f^{(i)}(\X)$. This weight and parameter estimation from the data is typically approached as a maximum likelihood estimation problem \cite{doi:10.1111/insr.12315}, often tackled using the expectation--maximization algorithm \cite{10.2307/2984875}.

\emph{Semi-parametric} estimation of mixture models has been studied in many works, including \cite{doi:10.1080/00949650214669,doi:10.1198/jcgs.2009.07175,chauveau2015,Xiang2019}. In this paper, however, we are interested in the \emph{nonparametric} estimation of conditional independence mixture models, i.e., no parametric forms will be assumed for the functions $f^{(i)}_v(x_v)$. Nonparametric estimation has been addressed by several works recently \cite{Hall2003,Hall2005,doi:10.1198/jcgs.2009.07175,JSSv032i06,10.5555/3023638.3023695,Levine2011,Zhu2016,Kasahara2014,CHAUVEAU20161,Zheng2019}. In this paper, we introduce a novel machine-learning-based-approach, called the InClass nets technique, to split the dataset into its different components in a nonparametric way. This splitting naturally leads to the estimation of the mixture model. In order to perform the splitting, the InClass nets technique directly exploits the fact that the variates $x_v$ are mutually independent of each other within each component. The biggest advantage offered by our machine-learning-based-technique over existing approaches is the possibility of tackling situations where the individual variates $x_v$ are high-dimensional. This opens up the possibility of using CIMMs for hitherto unfeasible applications.

The estimation of mixture models is closely tied to the concept of \emph{identifiability} of mixture models. A statistical model is said to be identifiable if it is theoretically possible to estimate the model (i.e., uniquely identify the parameters and functions that describe it) based on an infinite dataset sampled from it. A model will not be identifiable if two or more parameterizations of the model are observationally indistinguishable under those circumstances.

In the situations where the CIMM is identifiable, our technique estimates the true $w_i$ and $f^{(i)}_v$. On the other hand, when the model is not identifiable, our technique will yield one of the parameterizations that best fits the available data. In \sref{sec:identifiability}, we provide some new results on the (nonparametric) identifiability of bivariate ($V=2$) conditional independence mixture models, to supplement existing results on nonparametric mixture model identifiability \cite{teicher1967,yakowitz1968,Gyllenberg1994,Hall2003,Hall2005,Elmore2005,Allman2009,Kasahara2009,kovtun_akushevich_yashin_2014,Tahmasebi2018}.


\subsection{Unsupervised Classification in Machine Learning}
In this paper, we approach the estimation of CIMMs as a classification problem---classifying the datapoints in a given dataset into the different categories will lead to the estimation of the mixture model in a straightforward way.

This classification needs to be performed in an unsupervised manner since the dataset being analyzed does not contain labels for the component each datapoint belongs to. In this way, our method has connections to unsupervised clustering techniques like k-means clustering \cite{1056489} and other density-based clustering techniques \cite{10.5555/3001460.3001507}. However, our approach does not rely on the different components being spatially clustered to perform the classification.

The intuition behind our method can be understood as follows: In supervised classification, the target class labels associated with the training datapoints serve as the supervisory signal for training the classifier. In the absence of target labels, a quantity that is dependent on or shares mutual information with, the (unavailable) target label can be used as the supervisory signal. Now, lets say we are training a classifier that bases its decision or output only on the first variate $x_1$. The other variates $\{x_2,\dots,x_V\}$ can serve as the supervisory signal, since they contain information regarding the component $i$ the datapoint belongs to. Our approach can be thought of as training $V$ classifiers, one for each of the $V$ variates, with each classifier relying on the other $V-1$ variates to act as the supervisory signal for training.

There are a few ways of interpreting and actualizing this intuition \cite{10.1145/956750.956764,10.5555/647235.720080,ji2019invariant}. For example, in \cite{ji2019invariant}, neural networks were trained without supervision to classify images, using a training dataset consisting of pairs of images, where the images in a given pair are from the same category. In the InClass nets approach developed in this paper, the neural network architecture and training cost functions we develop will primarily be geared towards estimating conditional independence mixture models, which, as shown below in \sref{sec:MNIST}, can nevertheless be used for classification similar to the technique of \cite{ji2019invariant}. In \sref{sec:multilabel}, we also discuss a straightforward extension of the technique in \cite{ji2019invariant} to handle n-tuples of data, where the components of the n-tuple could be from different sample spaces (as opposed to pairs of images from the same sample space of images).

The idea of using a quantity that shares information with the true labels as the supervisory signal has been employed previously in weakly supervised classification techniques like ``Learning from Label Proportions'' (LLP) \cite{10.1145/1390156.1390254,NIPS2014_5453,Yu2014OnLW} and ``Classification Without Labels'' (CWoLa) \cite{Metodiev:2017vrx}. LLP and CWola learn to distinguish between different classes of datapoints, using multiple mixed datasets which differ in the mixing proportions of the classes---the identity of the mixed dataset a given datapoint belongs to serves as the supervisory signal, since it contains information about the class the datapoint belongs to (due to the mixing proportions in different mixtures being different). While CWola and LLP are not fully unsupervised techniques (since they still require a label indicating which mixture a training datapoint belongs to), they are applicable even in situations where the distribution of the feature $\X$ within a given class $i$ does not factorize as $\displaystyle\prod_{v=1}^V\,f^{(i)}_v$.

\subsection{Other Related Work}
The idea of separating a mixture into its components using a mutual-information-based technique is similar in spirit to Independent Component Analysis (ICA) \cite{hyvarinen2001independent}. However, ICA solves a signal separation problem where multiple mixtures with different mixing weights for the components are provided---this is different from the problem of separating data from a single CIMM into its components.


Mixture models have applications in data analysis in high energy physics \cite{Stepanek:2015jqa,Cranmer:2015bka,Cranmer:2016swd}, where datasets are mixtures of ``events'' (datapoints) produced under different ``processes'' (categories). The ${}_s\mathcal{P}lot$ technique \cite{Pivk:2004ty}, which is popular in data analysis in high energy physics, is used to analyze bivariate conditional independence mixture models where the distribution of one of the variables (referred to as the discriminating variable) is known \textit{a priori}. In such situations, the ${}_s\mathcal{P}lot$ technique can estimate the distribution of the other variable, referred to as the control variable. On the other hand, the InClass nets approach introduced in this paper is capable of estimating the mixture model \emph{without any knowledge of the distributions of any of the variables}. In \sref{subsec:prior_knowledge}, we describe how the InClass nets approach can be modified to incorporate additional information about the distributions of some of the variates.

\section{Independent Classifier Networks (InClass Nets)} \label{sec:inclass}

For the purpose of nonparametric estimation of conditional independence mixture models, we introduce a new neural network architecture which we shall call ``Independent Classifier networks'' or ``InClass nets'' for short. Under InClass nets, the $V$ variates $\{x_1,\dots,x_V\}$ of the input $\X$ are fed into $V$ independent neural networks---one variate for each independent network. Each of the $V$ networks returns a multi-class classifier output. More explicitly, for each $v\in \{1,\dots,V\}$, the $v$-th classifier network returns a vector $\left(\eta^{(1)}_v(x_v),\dots,\eta^{(C)}_v(x_v)\right)$, whose $i$-th component can \emph{roughly} be interpreted as the probability that a datapoint belongs to category $i$, conditional only on its $x_v$ value. 

\begin{figure}[t]
 \centering
 \tikzstyle{block} = [rectangle, draw, text width=13em, text centered, rounded corners, minimum height=5em]
 \tikzstyle{line} = [draw, -latex]
\begin{tikzpicture}[node distance = 8.2em, auto]
\node [block, anchor=center] (classifier1) {Classifier 1\smallbreak Maps $x_1$ to a multi-class classifier output (softmax)};
\node [block, below of=classifier1, yshift=1.3em] (classifier2) {Classifier 2\smallbreak Maps $x_2$ to a multi-class classifier output (softmax)};
\node [below of=classifier2, yshift=4em] (spacer) {\LARGE\vdots};
\node [block, below of=spacer, yshift=3.5em] (classifierV) {Classifier $V$\smallbreak Maps $x_V$ to a multi-class classifier output (softmax)};
\node [block, dashed, align=center, text width=19em, anchor=center, minimum height=25em] at (0em,-8.8em) (a) {\vskip 22em InClass Net};

\path [line] ($(classifier1)-(14em,-2em)$) -- ($(classifier1)-(11em,-2em)$);
\path [line,-] ($(classifier1)-(11.3em,-2em)$) -- ($(classifier1)-(6.85em,-2em)$);
\path [line] ($(classifier1)-(14em,-1em)$) -- ($(classifier1)-(11em,-1em)$);
\path [line,-] ($(classifier1)-(11.3em,-1em)$) -- ($(classifier1)-(6.85em,-1em)$);
\node at ($(classifier1)-(13em,+0.2em)$) {\LARGE\vdots};
\node at ($(classifier1)-(15.5em,+0em)$) {$x_1$};
\path [line] ($(classifier1)-(14em,+2em)$) -- ($(classifier1)-(11em,+2em)$);
\path [line,-] ($(classifier1)-(11.3em,+2em)$) -- ($(classifier1)-(6.85em,+2em)$);

\path [line,-] ($(classifier1)-(-14em,-2em)$) -- ($(classifier1)-(-11em,-2em)$);
\path [line] ($(classifier1)-(-6.85em,-2em)$) -- ($(classifier1)-(-11.3em,-2em)$);
\path [line,-] ($(classifier1)-(-14em,-1em)$) -- ($(classifier1)-(-11em,-1em)$);
\path [line] ($(classifier1)-(-6.85em,-1em)$) -- ($(classifier1)-(-11.3em,-1em)$);
\node at ($(classifier1)-(-13em,+0.2em)$) {\LARGE\vdots};
\node at ($(classifier1)-(-17em,+0em)$) {$\left\{\eta^{(i)}_1(x_1)\right\}$};
\path [line,-] ($(classifier1)-(-14em,+2em)$) -- ($(classifier1)-(-11em,+2em)$);
\path [line] ($(classifier1)-(-6.85em,+2em)$) -- ($(classifier1)-(-11.3em,+2em)$);

\path [line] ($(classifier2)-(14em,-2em)$) -- ($(classifier2)-(11em,-2em)$);
\path [line,-] ($(classifier2)-(11.3em,-2em)$) -- ($(classifier2)-(6.85em,-2em)$);
\path [line] ($(classifier2)-(14em,-1em)$) -- ($(classifier2)-(11em,-1em)$);
\path [line,-] ($(classifier2)-(11.3em,-1em)$) -- ($(classifier2)-(6.85em,-1em)$);
\node at ($(classifier2)-(13em,+0.2em)$) {\LARGE\vdots};
\node at ($(classifier2)-(15.5em,+0em)$) {$x_2$};
\path [line] ($(classifier2)-(14em,+2em)$) -- ($(classifier2)-(11em,+2em)$);
\path [line,-] ($(classifier2)-(11.3em,+2em)$) -- ($(classifier2)-(6.85em,+2em)$);

\path [line,-] ($(classifier2)-(-14em,-2em)$) -- ($(classifier2)-(-11em,-2em)$);
\path [line] ($(classifier2)-(-6.85em,-2em)$) -- ($(classifier2)-(-11.3em,-2em)$);
\path [line,-] ($(classifier2)-(-14em,-1em)$) -- ($(classifier2)-(-11em,-1em)$);
\path [line] ($(classifier2)-(-6.85em,-1em)$) -- ($(classifier2)-(-11.3em,-1em)$);
\node at ($(classifier2)-(-13em,+0.2em)$) {\LARGE\vdots};
\node at ($(classifier2)-(-17em,+0em)$) {$\left\{\eta^{(i)}_2(x_2)\right\}$};
\path [line,-] ($(classifier2)-(-14em,+2em)$) -- ($(classifier2)-(-11em,+2em)$);
\path [line] ($(classifier2)-(-6.85em,+2em)$) -- ($(classifier2)-(-11.3em,+2em)$);

\path [line] ($(classifierV)-(14em,-2em)$) -- ($(classifierV)-(11em,-2em)$);
\path [line,-] ($(classifierV)-(11.3em,-2em)$) -- ($(classifierV)-(6.85em,-2em)$);
\path [line] ($(classifierV)-(14em,-1em)$) -- ($(classifierV)-(11em,-1em)$);
\path [line,-] ($(classifierV)-(11.3em,-1em)$) -- ($(classifierV)-(6.85em,-1em)$);
\node at ($(classifierV)-(13em,+0.2em)$) {\LARGE\vdots};
\node at ($(classifierV)-(15.5em,+0em)$) {$x_V$};
\path [line] ($(classifierV)-(14em,+2em)$) -- ($(classifierV)-(11em,+2em)$);
\path [line,-] ($(classifierV)-(11.3em,+2em)$) -- ($(classifierV)-(6.85em,+2em)$);

\path [line,-] ($(classifierV)-(-14em,-2em)$) -- ($(classifierV)-(-11em,-2em)$);
\path [line] ($(classifierV)-(-6.85em,-2em)$) -- ($(classifierV)-(-11.3em,-2em)$);
\path [line,-] ($(classifierV)-(-14em,-1em)$) -- ($(classifierV)-(-11em,-1em)$);
\path [line] ($(classifierV)-(-6.85em,-1em)$) -- ($(classifierV)-(-11.3em,-1em)$);
\node at ($(classifierV)-(-13em,+0.2em)$) {\LARGE\vdots};
\node at ($(classifierV)-(-17em,+0em)$) {$\left\{\eta^{(i)}_V(x_V)\right\}$};
\path [line,-] ($(classifierV)-(-14em,+2em)$) -- ($(classifierV)-(-11em,+2em)$);
\path [line] ($(classifierV)-(-6.85em,+2em)$) -- ($(classifierV)-(-11.3em,+2em)$);
\end{tikzpicture}
 \caption{Basic architecture of Independent Classifier Networks (InClass nets). The $V$ variates $\{x_1,\dots,x_V\}$ of the input $\X$ are fed into $V$ independent neural networks, each of which returns a multi-class classifier output $\eta^{(i)}_v(x_v)$ for $v\in \{1,\dots,V\}$.}
 \label{fig:inclass}
\end{figure}
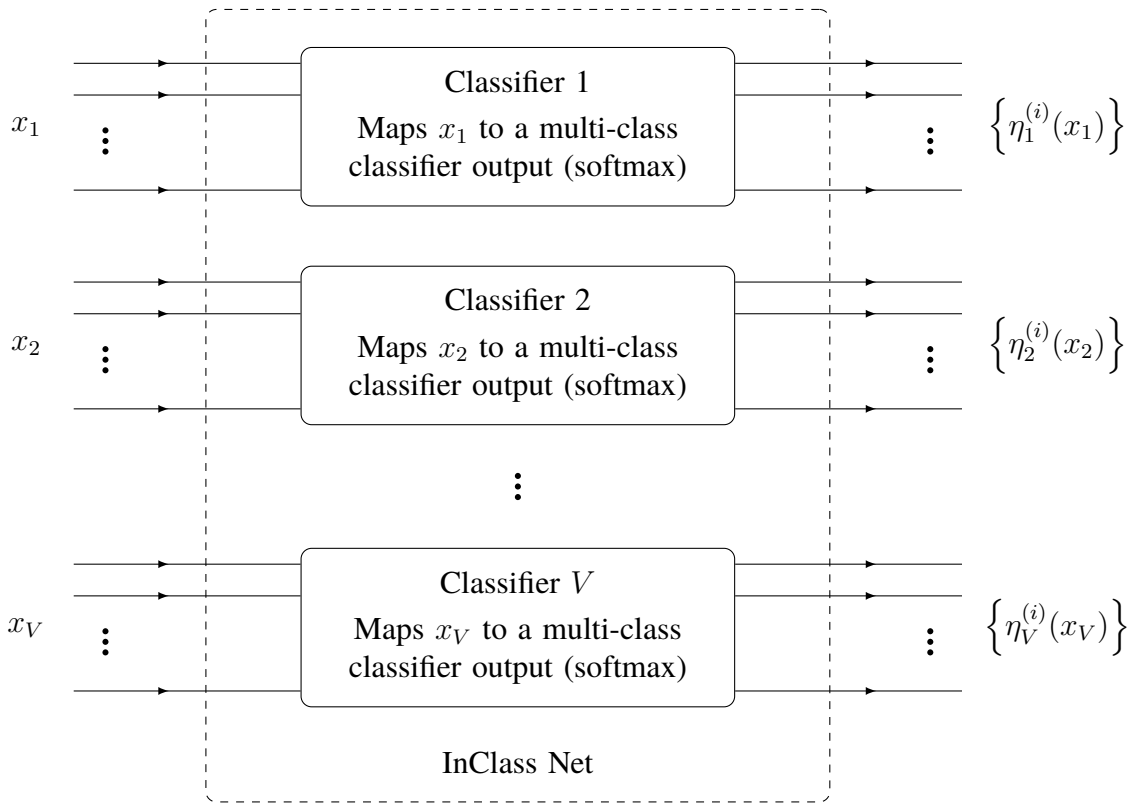

The outputs of the independent classifiers are constrained to obey
\begin{subequations}\label{eq:InClass_constraints}
\begin{alignat}{2}
 \eta^{(i)}_v(x_v) &\geq 0\,,&&\qquad \forall (i,v)\in\{1,\dots,C\}\times\{1,\dots,V\}\,,\\
 \sum_{i=1}^C~\eta^{(i)}_v(x_v) &= 1\,,&&\qquad \forall v\in\{1,\dots,V\}\,,
\end{alignat}
\end{subequations}
possibly using the softmax output layer\footnote{For the case of $C=2$, one can also simply use a one-dimensional output layer constrained to be in $[0,1]$, with $(\texttt{output}, 1-\texttt{output})$ serving as $(\eta^{(1)}, \eta^{(2)})$.} \cite{Goodfellow-et-al-2016} as follows:
\begin{equation}
 \eta^{(i)}_v(x_v) = \texttt{softmax}^{(i)}\left(z^{(1)}_v,\dots,z^{(C)}_v\right) \,,\qquad \forall (i,v)\in\{1,\dots,C\}\times\{1,\dots,V\}\,,
\end{equation}
where the $z^{(i)}$-s are the inputs\footnote{This is assuming that the output layer only performs the softmax operation. If softmax is used as an activation function of the final layer, then the $z^{(i)}$-s represent the outputs of the layer before applying the activation function.} to the final (output) layer of the corresponding network. The $\texttt{softmax}$ function is defined as
\begin{equation}
 \texttt{softmax}^{(i)}\left(z^{(1)}_v,\dots,z^{(C)}_v\right) \equiv \frac{\exp\!{\left(z^{(i)}_v\right)}}{\sum\limits_{j=1}^C~\displaystyle \exp\!{\left(z^{(j)}_v\right)}}\,.
\end{equation}
\Fref{fig:inclass} illustrates the basic architecture of InClass nets. In the next few sections, we will build the framework for estimating conditional independence mixture models using InClass nets.

Recall that the variates $x_v$ can be multi-dimensional. In particular, InClass nets can handle high-dimensional data types like images. The choice of architecture for the individual classifier networks can be influenced by the nature of the input data the classifier will handle.

For the purposes of this paper, we have restricted the output dimensionality of the individual classifiers to be the same (equal to $C$). We have also restricted the inputs $\{x_1,\dots,x_V\}$ of the individual classifiers to form a non-overlapping partition of the features or attributes in $\X$, which is in line with the structure of conditional independence mixture models. However, InClass nets can have applications outside mixture model estimation as well, and for those purposes it may be appropriate to lift the above restrictions. For example, InClass nets can be used to perform unsupervised \emph{multi-label} classification, where the outputs of different classifiers correspond to different labels. In this case, the different classifiers can have different output dimensionalities and the inputs to these networks can also potentially have overlapping features. In \sref{sec:multilabel} we will briefly indicate how the multi-label variant of InClass nets can be trained to perform unsupervised classification by maximizing the mutual information between the classifier outputs.

\subsection{Parameterizing Mixture Models With InClass Nets}
In this section we will show how conditional independence mixture models can be parametrized using InClass nets. The parametrization will be done using the Independent Pseudo Classifiers representation of mixture models which will be introduced in \sref{subsec:ipcr}. But as a useful lead-up, let us first introduce the Constrained Independent Classifiers representation.
\subsubsection{Constrained Independent Classifiers Representation}
\label{subsec:cicr}
The mixture model in \eqref{eq:cifmm} is completely specified by the mixture weights $w_i$ and the distributions $f^{(i)}_v$. Recall that they satisfy
\begin{subequations}
\label{eq:constraint_wf}
\begin{alignat}{2}
 w_i\geq 0\,, ~~f^{(i)}_v(x_v) &\geq 0\,,&&\qquad\forall (i,v)\in \{1,\dots,C\}\times\{1,\dots,V\}\,,\\
 \sum_{i=1}^C~w_i &= 1\,,\\
 \int dx_v~f^{(i)}_v(x_v) &= 1\,,&&\qquad \forall (i,v)\in \{1,\dots,C\}\times\{1,\dots,V\}\,.
\end{alignat}
\end{subequations}
The goal of this paper is to develop a machine learning technique to fit a mixture model to the given data {\bf in an agnostic, nonparametric, manner}. In other words, we will estimate the weights $w_i$ and the distributions $f^{(i)}_v$, without assuming, \textit{a priori}, any parameterized forms (like Gaussians, exponentials, etc.) for $f^{(i)}_v$.
We will approach this as an unsupervised multi-class classification problem---classifying the data into different components will automatically result in an estimation of the mixture model\footnote{Directly modeling the distributions $f^{(i)}_v$ is possible using generative networks, but classifier outputs are more robust quantities, e.g., they are invariant under invertible transformations of the $x_v$-s, and are typically easier to learn in machine learning.}. To this end, let us rewrite the mixture model distribution in terms of the marginal distributions $\P_{\!v}(x_v)$ of the individual variates and multi-class classifiers $\alpha^{(i)}_v(x_v)$ given by
\begin{subequations}
\begin{alignat}{1}
 \P_{\!v}(x_v) &= \left[\,\prod_{\substack{u=1\\u\neq v}}^V\int dx_{u}\right]~\P(\X) = \sum_{i=1}^C~w_i\,f^{(i)}_v(x_v)\,,\qquad \forall v\in\{1,\dots,V\}\,,\\
 \alpha^{(i)}_v(x_v) &= \frac{w_i\,\,f^{(i)}_v(x_v)}{\P_{\!v}(x_v)}\,,\qquad \forall (i,v)\in\{1,\dots,C\}\times\{1,\dots,V\}\,.
\end{alignat}
\end{subequations}
$\P_{\!v}$ is the probability density of the $v$-th variate in the full mixture and can be directly accessed from a dataset sampled from $\P$. $\alpha^{(i)}_v(x_v)$ can be interpreted as the probability that an observed datapoint is from component $i$ conditional on the value of $x_v$. The vector function $\left(\alpha^{(1)}_v(x_v),\dots,\alpha^{(C)}_v(x_v)\right)$ can be interpreted as the output of a multi-class ``classifier'' that returns the probability of a datapoint $\X$ to have come from the different components based only on the $v$-th variate. At this point, one might already notice an emerging connection with InClass nets, which we shall crucially exploit below. The marginals density functions $\P_{\!v}$ and the multi-class classifiers $\alpha^{(i)}_v$ satisfy
\begin{subequations}
\label{eq:constraint_IC}
\begin{alignat}{2}
 \P_{\!v}(x_v)\geq 0\,,~~\alpha^{(i)}_v(x_v) &\geq 0\,,\quad&&\forall(i,v)\in \{1,\dots,C\}\times\{1,\dots,V\}\,, \label{eq:IC_nonnegative}\\
 \int dx_v~\P_{\!v}(x_v) &= 1\,,\quad&&\forall v\in \{1,\dots,V\}\,,\\
 \sum_{i=1}^C \alpha^{(i)}_v(x_v) &= 1\,,\quad&& \forall v\in \{1,\dots,V\} \label{eq:alphaadd1}\,,\\
 \int dx_v~\P_{\!v}(x_v)\,\alpha^{(i)}_v(x_v) &= \int dx_{u}~\P_{\!u}(x_{u})\,\alpha^{(i)}_{u}(x_{u})\,,~~&&\forall(i,v,u)\in \{1,\dots,C\}\times\{1,\dots,V\}^2\,, \label{eq:IC_difficulty}
\end{alignat}
\end{subequations}
where the integrals in \eqref{eq:IC_difficulty} are simply equal to the weight $w_i$ of the $i$-th component. There is a one-to-one map\footnote{This is not a statement on the identifiability of conditional independence mixture models. Identifiability of mixture models will be briefly discussed in \sref{sec:identifiability}.} from the description of the mixture model in terms of the $w_i$-s and $f^{(i)}_v$-s satisfying \eqref{eq:constraint_wf} to the description in terms of $\P_{\!v}$-s and $\alpha^{(i)}_v$-s satisfying \eqref{eq:constraint_IC}. This can be seen from the existence of the inverse transform shown below:
\begin{subequations}
\begin{alignat}{2}
 w_i &= \int dx_v~\P_{\!v}(x_v)\,\alpha^{(i)}_v(x_v) \equiv E_{\P}\left[\alpha^{(i)}_v\right]\,,&&\qquad \forall (i,v)\in \{1,\dots,C\}\times\{1,\dots,V\},\\
 f^{(i)}_v(x_v) &= \frac{\P_{\!v}(x_v)~\alpha^{(i)}_v(x_v)}{w_i}\,,&&\qquad \forall (i,v)\in \{1,\dots,C\}\times\{1,\dots,V\}, \label{eq:f_from_alpha}
\end{alignat}
\end{subequations}
where $E_{\P}[\,\cdots]$ represents the expectation value of $\,\cdots$ under the model. The probability density of $\X$ under the corresponding mixture model is given by
\begin{subequations}\label{eq:IC_pdf}
\begin{align}
 \P(\X) &= \sum_{i=1}^C~w_i~\prod_{v=1}^V~\frac{\P_{\!v}(x_v)~\alpha^{(i)}_v(x_v)}{w_i}\\
 &= \left[\prod_{v=1}^V \P_{\!v}(x_v)\right]~~\left[\sum_{i=1}^C~w_i^{1-V}~\prod_{v=1}^V~\alpha^{(i)}_v(x_v)\right]\,.
\end{align}
\end{subequations}

As mentioned earlier, the marginal distributions $\P_{\!v}$ can be directly estimated from the data. The functions $\alpha^{(i)}_v$ can \emph{potentially} be modeled using InClass nets. The only hurdle is that while the outputs $\alpha^{(i)}_v$ of the $V$ neural networks can be constrained to obey \eqref{eq:IC_nonnegative} and \eqref{eq:alphaadd1} using the softmax output layer (as seen in \eqref{eq:InClass_constraints}), the constraint in \eqref{eq:IC_difficulty} in general will {\bf not} be satisfied by independent classifiers. We will handle this difficulty next in \sref{subsec:ipcr}. We will refer to the description in terms of $\P_{\!v}$-s and $\alpha^{(i)}_v$-s satisfying (\ref{eq:IC_nonnegative}--\ref{eq:IC_difficulty}) as the Constrained Independent Classifiers (CIC) representation of the mixture model.

\subsubsection{Independent Pseudo Classifiers Representation}
\label{subsec:ipcr}
To accommodate the fact that independent classifiers will not obey the constraint \eqref{eq:IC_difficulty} of the Constrained Independent Classifiers representation, we introduce the Independent Pseudo Classifiers (IPC) representation in terms of pseudo marginals $\Q_{v}$ and pseudo classifiers $\beta^{(i)}_v$ which only satisfy the equivalents of constraints (\ref{eq:IC_nonnegative}--\ref{eq:alphaadd1}):
\begin{subequations}
\label{eq:constraint_IPC}
\begin{alignat}{2}
 \Q_{v}(x_v)\geq 0\,,~~\beta^{(i)}_v(x_v) &\geq 0\,,&&\qquad \forall(i,v)\in \{1,\dots,C\}\times\{1,\dots,V\}\,, \label{eq:IPC_nonnegative}\\
 \int dx_v~\Q_{v}(x_v) &= 1\,,&&\qquad \forall v\in \{1,\dots,V\}\,,\\
 \sum_{i=1}^C \beta^{(i)}_v(x_v) &= 1\,,&&\qquad \forall v\in \{1,\dots,V\}\,. \label{eq:IPC_add1}
\end{alignat}
\end{subequations}
The mixture weights under the IPC representation are given by
\begin{equation}
 w_i = \frac{\tilde{w}_i}{\sum\limits_{j=1}^C~\tilde{w}_{j}}\,,\qquad \forall i\in \{1,\dots,C\}\,, \label{eq:tilde_w_norm}
\end{equation}
where the unnormalized weights $\tilde{w}_i$-s are given by
\begin{equation} \label{eq:tilde_w}
\begin{split}
 \tilde{w}_i = \left[\prod_{v=1}^V\int dx_v~\Q_{v}(x_v)\,\beta^{(i)}_v(x_v)\right]^{1/V} &= \left[\prod_{v=1}^V~E_{\Q}\left[\beta^{(i)}_v\right]\right]^{1/V}\\
 &\equiv \left[\prod_{v=1}^V~\varphi^{(i)}_v\right]^{1/V}\,,\qquad \forall i\in \{1,\dots,C\}\,,
\end{split}
\end{equation}
where $\varphi^{(i)}_v \equiv E_{\Q}\left[\beta^{(i)}_v\right]$ represents the expectation value of $\beta^{(i)}_v$ under the distribution $\Q_{v}$. In \eqref{eq:tilde_w_norm} we have used the geometric mean\footnote{It is also possible to use other mean functions, including generalized means, instead of the geometric mean (with appropriate modifications to other relevant quantities and expressions).} $\tilde{w_i}$ of the mixture weights $\varphi^{(i)}_v$ ``proposed'' by the individual pseudo classifiers for component $i$ as its actual mixture weight $w_i$ in the mixture model, after an appropriate scaling to make the weights add up to 1 across all components. We will refer to $\varphi^{(i)}_v$ as the pseudo weight of component $i$ corresponding to variate $v$. In analogy to \eqref{eq:IC_pdf}, we write the probability density function for the mixture model in the IPC representation as
\begin{subequations}
\label{eq:IPC_representation}
\begin{alignat}{2}
 \P(\X) &= \sum_{i=1}^C~w_i~\prod_{v=1}^V~\frac{\Q_{v}(x_v)~\beta^{(i)}_v(x_v)}{\tilde{w}_i} \label{eq:IPC_representation_a}\\
 &= \left[\prod_{v=1}^V \Q_{v}(x_v)\right]~~\frac{\displaystyle\sum_{i=1}^C~\tilde{w}_i^{1-V}~\prod_{v=1}^V~\beta^{(i)}_v(x_v)}{\displaystyle\sum_{i=1}^C~\tilde{w}_i}\,, \label{eq:IPC_representation_b}
\end{alignat}
\end{subequations}
where the $\tilde{w}_i$-s can be written in terms of $\Q_{v}$-s and $\beta^{(i)}_v$-s using \eqref{eq:tilde_w}. We will now make the following observations relevant to our goal of fitting a mixture model to data using InClass nets:
\begin{enumerate}
 \item \textbf{IPC describes a conditional independence mixture model.} This is because the dependences of the distribution on the different $x_v$-s factorize within each component $i$ in \eqref{eq:IPC_representation_a}. The overall probability density in \eqref{eq:IPC_representation_a} is also normalized to 1, since as per \eqref{eq:tilde_w}
 \begin{equation}
  \prod_{v=1}^V \int dx_v~\Q_{v}(x_v)~\beta^{(i)}_v(x_v) = \tilde{w}_i^V\,.
 \end{equation}
 \item \textbf{The pseudo marginals and pseudo classifiers do not necessarily correspond to the true marginals and classifiers.} However, the true marginals and classifiers of the CIC representation can be extracted from the IPC representation as follows
 \begin{subequations}
 \begin{alignat}{2}
 \P_{\!v}(x_v) &= \left[\,\prod_{\substack{u=1\\u\neq v}}^V\int dx_{u}\right]~\P(\X)\\
 &= \frac{\Q_{v}(x_v)}{\sum\limits_{i=1}^C~\tilde{w}_i}~\sum_{i=1}^C~\beta^{(i)}_v(x_v)~\tilde{w}_i^{1-V}~\left[\,\prod_{\substack{u=1\\u\neq v}}^V E_{\Q}\left[\beta^{(i)}_u\right]\right]\\
 &= \frac{\Q_{v}(x_v)}{\sum\limits_{i=1}^C~\tilde{w}_i}~\sum_{i=1}^C~\frac{\beta^{(i)}_v(x_v)~\tilde{w}_i}{E_{\Q}\left[\beta^{(i)}_v\right]} = \frac{\Q_{v}(x_v)}{\sum\limits_{i=1}^C~\tilde{w}_i}~\sum_{i=1}^C~\frac{\beta^{(i)}_v(x_v)~\tilde{w}_i}{\varphi^{(i)}_v}\,, \label{eq:ipc_marginal}\\
 \alpha^{(i)}_v(x_v) 
 &= \left[\sum_{j=1}^C~\frac{\beta^{(j)}_v(x_v)~\tilde{w}_{j}}{\varphi^{(j)}_v}\right]^{-1}~~\frac{\beta^{(i)}_v(x_v)~\tilde{w}_i}{\varphi^{(i)}_v}\,, \label{eq:true_classifier}
 \end{alignat}
 \end{subequations}
 where \eqref{eq:true_classifier} can be deduced from the form of \eqref{eq:ipc_marginal}. 
 Substituting these expressions for $\P_{\!v}(x_v)$ and $\alpha^{(i)}_v(x_v)$ in \eqref{eq:f_from_alpha}, we get the following expression for reconstructing the distributions $f^{(i)}_v$ within the different components:
 \begin{equation}
  f^{(i)}_v(x_v) = \frac{\Q_v(x_v)~\beta^{(i)}_v(x_v)}{\varphi^{(i)}_v} = \frac{\Q_v(x_v)~\beta^{(i)}_v(x_v)}{E_{\Q}\left[\beta^{(i)}_v\right]}\,. \label{eq:IPC_f}
 \end{equation}
 In this way we can reconstruct the mixture model corresponding to a given set of pseudo marginals and pseudo-classifiers of the IPC representation.
 \item \textbf{The IPC representation of a mixture model is not unique.} Unlike the CIC representation, we cannot find a unique map from the weights $w_i$ and $f^{(i)}_v$ to the pseudo marginals and pseudo classifiers. This is because of the additional degrees of freedom due to the removal of the constraints in $\eqref{eq:IC_difficulty}$.
 \item \textbf{Every mixture model has an IPC representation\footnote{Not necessarily unique, even after imposing the constraint that the pseudo marginals match the true marginals.} in which the pseudo marginals match the true marginals of the model.} This can be seen from the fact that the true marginals $\P_{\!v}$ and classifiers $\alpha^{(i)}_v$ from the CIC representation of a mixture model can used as the pseudo marginals $\Q_v$ and pseudo classifiers $\beta^{(i)}_v$ under the IPC representation to get the same model.
\end{enumerate}
Observation (4) means that in order to fit a mixture model to data, we can restrict ourselves to IPC representations of the mixture models with the pseudo marginals set to the marginals of the data. The only remaining unknowns in the IPC representation are the pseudo classifiers $\beta^{(i)}_v(x_v)$ which we can parameterize using an InClass net, identifying $\beta^{(i)}_v$ with the network output $\eta^{(i)}_v$. In the next section we will develop the technique to fit a mixture model parameterized with an InClass net to a given dataset.


\subsection{Fitting Mixture Models to Data With InClass Nets}
In this section we will construct a cost function which can be used to train InClass nets to fit mixture models to the given data. Let the data to which we want to fit a mixture model be sampled from the true underlying distribution $\P^\ast(\X)$ with true marginals $\P^\ast_{\!v}(x_v)$. As per observation (4) in the previous section, we restrict our attention to IPC representations with $\Q_{v}\equiv \P^\ast_{\!v}$. Using \eqref{eq:IPC_representation_b} and \eqref{eq:tilde_w}, we can write the probability density of $\X$ under this restricted class of mixture models as
\begin{equation}
 \P(\X) = \left[\prod_{v=1}^V \P^\ast_{\!v}(x_v)\right]~\frac{\displaystyle\sum_{i=1}^C~\left[\prod_{v=1}^V~\beta^{(i)}_v(x_v)~\left(E_{\P^\ast}\left[\beta^{(i)}_v\right]\right)^{(1-V)/V}\right]}{\displaystyle\sum_{i=1}^C~\left[\prod_{v=1}^V\left(E_{\P^\ast}\left[\beta^{(i)}_v\right]\right)^{1/V}\right]}\,, \label{eq:dist_IPC}
\end{equation}
where $E_{\P^\ast}$ refers to the expectation value under the true distribution of the data. The best-fitting $\P$ can be estimated by minimizing the Kullback--Leibler (KL) divergence from $\P$ to $\P^\ast$ given by
\begin{equation}
 \mathrm{KL}\left[\P^\ast~\big|\big|~\P \right] = \int d\X~\P^\ast(\X)~\log{\left[\frac{\P^\ast(\X)}{\P(\X)}\right]}\,. \label{eq:KL}
\end{equation}
Note that minimizing the KL divergence (over some class of distributions) is equivalent to, and commonly known in some disciplines as, maximizing the likelihood in the large statistics limit. Using the expression for $\P$ from \eqref{eq:dist_IPC}, we can rewrite \eqref{eq:KL} as
\begin{subequations}
\begin{align}
 \mathrm{KL}\left[\P^\ast~\big|\big|~\P \right] &= \int d\X ~\P^\ast(\X)~\log{\left[\frac{\P^\ast(\X)}{\left[\prod\limits_{v=1}^V \P^\ast_{\!v}(x_v)\right]}~\frac{\left[\prod\limits_{v=1}^V \P^\ast_{\!v}(x_v)\right]}{\P(\X)}\right]}\\
 &= \int d\X ~\P^\ast(\X)~\log{\left[\frac{\P^\ast(\X)}{\left[\prod\limits_{v=1}^V \P^\ast_{\!v}(x_v)\right]}\right]} - \int d\X ~\P^\ast(\X)~\log{\left[\frac{\P(\X)}{\left[\prod\limits_{v=1}^V \P^\ast_{\!v}(x_v)\right]}\right]} \label{eq:kl_split}\\
 &= C^\ast(x_1,\dots,x_V) - E_{\P^\ast}\left[~\log\left\{\frac{\displaystyle\sum_{i=1}^C~\left[\prod_{v=1}^V~\beta^{(i)}_v~\left(E_{\P^\ast}\left[\beta^{(i)}_v\right]\right)^{(1-V)/V}\right]}{\displaystyle\sum_{i=1}^C~\left[\prod_{v=1}^V\left(E_{\P^\ast}\left[\beta^{(i)}_v\right]\right)^{1/V}\right]}\right\}~\right], \label{eq:ctc_intro}
\end{align}
\end{subequations}
where $C^\ast(x_1,\dots,x_V)$ is the total correlation \cite{5392532,garner_1962} of the $V$ variates in the data which is one of the generalizations of mutual information to more than two variables. It is given by the KL divergence from the product distribution $\prod\limits_v \P^\ast_{\!v}(x_v)$ to the joint distribution $\P^\ast(\X)$ as
\begin{equation}
 C^\ast(x_1,\dots,x_V) = \int d\X~\P^\ast(\X)~\log{\left[\frac{\P^\ast(\X)}{\P^\ast_{\!1}(x_1) ~\P^\ast_{\!2}(x_2)~\dots~\P^\ast_{\!V}(x_V)}\right]}\,.
 \label{eq:TC}
\end{equation}
Note that the $C^\ast$ term in \eqref{eq:ctc_intro} is independent of the state of the InClass net under consideration. This means that the second term in \eqref{eq:ctc_intro} can be used as a cost function for the network to minimize in order to minimize the KL divergence, and hence fit the mixture model parameterized by the InClass net to the data. Noting the similarity between the two terms in \eqref{eq:kl_split} and drawing inspiration from the naming of ``cross entropy'', we introduce the ``negative cross total correlation'' cost function (\texttt{neg\_ctc\_cost}) defined as
\begin{subequations}
\begin{align} \label{eq:neg_ctc_def}
 \texttt{neg\_ctc\_cost} &= \mathrm{KL}\left[\P^\ast~\big|\big|~\P \right] - C^\ast(x_1,\dots,x_V)\\
 &= - E_{\P^\ast}\left[~\log\left\{\frac{\displaystyle\sum_{i=1}^C~\left[\prod_{v=1}^V~\beta^{(i)}_v~\left(E_{\P^\ast}\left[\beta^{(i)}_v\right]\right)^{(1-V)/V}\right]}{\displaystyle\sum_{i=1}^C~\left[\prod_{v=1}^V\left(E_{\P^\ast}\left[\beta^{(i)}_v\right]\right)^{1/V}\right]}\right\}~\right]\,. \label{eq:neg_ctc}
\end{align}
\end{subequations}
Note that $\beta^{(i)}_v$ are functions of the corresponding input variate $x_v$. Despite the complicated appearance, this cost function provides a viable approach to learning the underlying mixture model from data. Let us make the following observations in the context of training InClass nets using this cost function, with outputs $\eta^{(i)}_v$ of the network identified with $\beta^{(i)}_v$.
\begin{enumerate}
 \item The cost function depends only on the outputs $\beta^{(i)}_v$ of the network. More precisely, the cost function depends on the distribution of the network output. It does not need the input data to be labelled to learn the mixture model, and the only supervisory signal exploited by the training process is the distribution of the input data.
 \item The cost function for a given state of the InClass net can be estimated using a (mini-)batch of training samples by approximating the expectation values $E_{\P^\ast}[\,\cdots]$ with sample means. The batch size should be large enough to perform a good estimation of $E_{\P^\ast}\left[\beta^{(i)}_v\right]$. 
\end{enumerate}
After training the InClass net, (\ref{eq:tilde_w_norm}--\ref{eq:IPC_representation}) and \eqref{eq:IPC_f} can be used to extract the fitted model, with the pseudo marginals $\Q_v$ set to the true marginals $\P^\ast_v$. The classifiers $\alpha^{(i)}_v(x_v)$ can be extracted from the pseudo classifiers $\beta^{(i)}_v(x_v)$ using \eqref{eq:true_classifier}. If one is interested in classifying the individual datapoints based on the full information $\X$, an aggregate classifier can be constructed, based on \eqref{eq:IPC_representation_b}, as
\begin{equation}
 \alpha^{(i)}_\mathrm{aggregate}(\X) = \frac{\displaystyle\tilde{w}_i^{1-V}~\prod_{v=1}^V~\beta^{(i)}_v(x_v)}{\displaystyle\sum_{j=1}^C~\tilde{w}_j^{1-V}~\prod_{v=1}^V~\beta^{(j)}_v(x_v)}\,.\label{eq:aggregate_classifier}
\end{equation}
Note that if there is a mismatch between the model learned by the InClass net and the true distribution the data is sampled from, then classifying the data using the aggregate classifier will not necessarily lead to components within which the $x_v$-s are independent.

\section{Bivariate Case}
When analyzing real data with conditional independence mixture models, a common difficulty is the identification of a suitable partitioning of the attributes of $\X$ into variates $x_v$ so that the distribution within each component would factorize to a good approximation. In this sense, a higher number of (conditionally independent) variates represents stronger assumptions about the underlying model. This makes the bivariate case ($V=2$) extremely important. The bivariate case is also difficult from an identifiability point of view---data distributed according to a conditional independence bivariate mixture model, in general, will not uniquely identify the model, since several different mixture models can lead to the same overall probability density $\P(\X)$. In \sref{sec:identifiability}, we will present some new results on the identifiability of conditional independence bivariate mixture models. In particular, we will provide the conditions under which bivariate mixture models are identifiable.

Despite being the most difficult case in terms of identifiability, the bivariate case lets us gain some useful intuition, as demonstrated with several examples in \Sref{sec:raindancesvi} below. But first, in preparation for \sref{sec:raindancesvi}, let us summarize the results from the previous sections for the bivariate case, in the order in which a typical analysis might use them.

\subsection{Notation}
The expressions from the previous sections become easier to follow if we explicitly write out the two variates, thus avoiding the product notation. To this end, let us simplify the notation by giving names $x$ and $y$ to our two variates $x_1$ and $x_2$, resulting in
\begin{subequations}
\begin{alignat}{7}
 &x \equiv x_1\,,\quad && \P_{\!x} \equiv \P_{\!1}\,,\quad && \Q_{x} \equiv \Q_{1}\,,\quad && \P^\ast_{\!x} \equiv \P^\ast_{\!1}\,,\quad && \alpha^{(i)}_x \equiv \alpha^{(i)}_1\,,\quad && \beta^{(i)}_x \equiv \beta^{(i)}_1\,,\quad && f^{(i)}_x \equiv f^{(i)}_1\,,\\
 &y \equiv x_2\,,\quad && \P_{\!y} \equiv \P_{\!2}\,,\quad && \Q_{y} \equiv \Q_{2}\,,\quad && \P^\ast_{\!y} \equiv \P^\ast_{\!2}\,,\quad && \alpha^{(i)}_y \equiv \alpha^{(i)}_2\,,\quad && \beta^{(i)}_y \equiv \beta^{(i)}_2\,,\quad && f^{(i)}_y \equiv f^{(i)}_2\,.
\end{alignat}
\end{subequations}
Under this notation, the conditional independence mixture model of \eqref{eq:cifmm} becomes simply
\begin{equation}
 \P(x, y) = \sum_{i=1}^C~w_i~f^{(i)}_x(x)~f^{(i)}_y(y)\,.\label{eq:bi_cifmm}
\end{equation}

\subsection{Cost Function}
Noting that total correlation is a generalization of mutual information for more than two variables, we will refer to the negative cross total correlation cost function of \eqref{eq:neg_ctc} in the bivariate special case as the ``negative cross mutual information'' cost function (\texttt{neg\_cmi\_cost}). Under our new notation, it is given by
\begin{equation}
 \texttt{neg\_cmi\_cost} = - E_{\P^\ast}\left[~\log\left\{\frac{\displaystyle\sum_{i=1}^C~\frac{\beta^{(i)}_x\,\beta^{(i)}_y}{\sqrt{\displaystyle E_{\P^\ast}\left[\beta^{(i)}_x\right]\,E_{\P^\ast}\left[\beta^{(i)}_y\right]}}}{\displaystyle\sum_{i=1}^C~\sqrt{\displaystyle E_{\P^\ast}\left[\beta^{(i)}_x\right]\,E_{\P^\ast}\left[\beta^{(i)}_y\right]}}\right\}~\right]\,,
 \label{eq:neg_cmi_cost}
\end{equation}
where, as before, $E_{\P^\ast}$ represents the expectation over the true distribution $\P^\ast(x, y)$ from which the data is sampled.

\subsection{Extracting the Learned Mixture Model From the Trained Network}
After training the InClass net, the trained $\beta^{(i)}_x$ and $\beta^{(i)}_y$ cannot directly be interpreted as classifiers based on $x$ and $y$ since they may correspond to different mixture weights. In order to extract the learned mixture model (and the corresponding classifiers), we can first estimate the pseudo weights $\varphi^{(i)}_x$ and $\varphi^{(i)}_y$ from the data as
\begin{equation}
 \varphi^{(i)}_x = E_{\P^\ast}\left[\beta^{(i)}_x\right]\,,\qquad\qquad \varphi^{(i)}_y = E_{\P^\ast}\left[\beta^{(i)}_y\right]\,. \label{eq:bi_pseudo}
\end{equation}
Now, using \eqref{eq:tilde_w} and \eqref{eq:true_classifier}, the marginals and classifiers for the model represented by the InClass net can be constructed as
\begin{subequations} \label{eq:bi_CIC}
\begin{alignat}{2}
 &\P_{\!x}(x) = \P^\ast_{\!x}(x)~\frac{\displaystyle\sum_{i=1}^C~\beta^{(i)}_x(x)~\displaystyle\sqrt{\displaystyle\frac{\varphi^{(i)}_y}{\displaystyle \varphi^{(i)}_x}}}{\displaystyle\sum_{i=1}^C~\sqrt{\displaystyle \varphi^{(i)}_x\,\varphi^{(i)}_y}}\,,\qquad\qquad
 &&\alpha^{(i)}_x(x) = \frac{\beta^{(i)}_x(x)~\displaystyle\sqrt{\displaystyle\frac{\varphi^{(i)}_y}{\displaystyle \varphi^{(i)}_x}}}{\displaystyle\sum_{j=1}^C~\beta^{(j)}_x(x)~\displaystyle\sqrt{\displaystyle\frac{\varphi^{(j)}_y}{\displaystyle \varphi^{(j)}_x}}}\,,\\
 &\P_{\!y}(y) = \P^\ast_{\!y}(y)~\frac{\displaystyle\sum_{i=1}^C~\beta^{(i)}_y(y)~\displaystyle\sqrt{\displaystyle\frac{\varphi^{(i)}_x}{\displaystyle \varphi^{(i)}_y}}}{\displaystyle\sum_{i=1}^C~\sqrt{\displaystyle \varphi^{(i)}_x\,\varphi^{(i)}_y}}\,,\qquad\qquad &&\alpha^{(i)}_y(y) = \frac{\beta^{(i)}_y(y)~\displaystyle\sqrt{\displaystyle\frac{\varphi^{(i)}_x}{\displaystyle \varphi^{(i)}_y}}}{\displaystyle\sum_{j=1}^C~\beta^{(j)}_y(y)~\displaystyle\sqrt{\displaystyle\frac{\varphi^{(j)}_x}{\displaystyle \varphi^{(j)}_y}}}\,.
\end{alignat}
\end{subequations}
From \eqref{eq:tilde_w_norm} and \eqref{eq:tilde_w}, the component weights of the learned model are given by
\begin{equation}
 w_i = \frac{\displaystyle\sqrt{\displaystyle \varphi^{(i)}_x\,\varphi^{(i)}_y}}{\displaystyle\sum_{j=1}^C~\sqrt{\displaystyle \varphi^{(j)}_x\,\varphi^{(j)}_y}} \label{eq:bi_w_final}
\end{equation}
and from \eqref{eq:IPC_f}, the distributions $f^{(i)}_x$ and $f^{(i)}_y$ within each component are given by
\begin{equation}
 f^{(i)}_x(x) = \frac{\P^\ast_{\!x}(x)~\beta^{(i)}_x(x)}{\displaystyle \varphi^{(i)}_x}\,,\qquad\qquad f^{(i)}_y(y) = \frac{\P^\ast_{\!y}(y)~\beta^{(i)}_y(y)}{\displaystyle \varphi^{(i)}_y}\,. \label{eq:bi_f_final}
\end{equation}
The corresponding joint distribution is given by
\begin{equation}
 \P(x, y) = \P^\ast_{\!x}(x)\,\P^\ast_{\!y}(y)~\frac{\displaystyle\sum_{i=1}^C~\frac{\beta^{(i)}_x(x)\,\beta^{(i)}_y(y)}{\sqrt{\displaystyle \varphi^{(i)}_x\,\varphi^{(i)}_y}}}{\displaystyle\sum_{i=1}^C~\sqrt{\displaystyle \varphi^{(i)}_x\,\varphi^{(i)}_y}}\,.
\end{equation}
Note that after estimating $\P^\ast_{\!x}$, $\P^\ast_{\!y}$, $\varphi^{(i)}_x$, and $\varphi^{(i)}_y$ from the dataset, the mixture model can be read off directly from the InClass net using \eqref{eq:bi_w_final} and \eqref{eq:bi_f_final}.

\subsection{Aggregate Classifier}
From \eqref{eq:aggregate_classifier}, the aggregate classifier that classifies the individual datapoints based on the full information $(x,y)$ is given by
\begin{equation}
 \alpha^{(i)}_\mathrm{aggregate}(x, y) = \frac{\displaystyle\frac{\beta^{(i)}_x(x)\,\beta^{(i)}_y(y)}{\sqrt{\displaystyle \varphi^{(i)}_x\,\varphi^{(i)}_y}}}{\displaystyle\sum_{j=1}^C~\frac{\beta^{(j)}_x(x)\,\beta^{(j)}_y(y)}{\sqrt{\displaystyle \varphi^{(j)}_x\,\varphi^{(j)}_y}}}\,. \label{eq:bi_aggregate}
\end{equation}

\section{Examples Using RainDancesVI (a Python Implementation of InClass Nets)}
\label{sec:raindancesvi}

We provide a public, \texttt{tensorflow}-based \cite{abadi2016tensorflow}, implementation of InClass nets as a Python 3 package called \href{\raindancesurl}{\raindances{}} \cite{rd6}. The package provides routines for wrapping the classifier networks of individual variates into InClass nets and cost functions for training them. It also provides utilities for extracting the model learned by the network post-training. In this section we will demonstrate the working of InClass nets using several toy examples analyzed using \raindances{}.

\subsection{Mixture of Two Independent Bivariate Gaussians \texorpdfstring{$(V=2, C=2)$}{(V=2, C=2)}} \label{subsec:bivariate_gaussian}
\begin{figure}[t!]
 \centering
 \includegraphics[height=.43\textwidth,trim=0 0 70 0,clip]{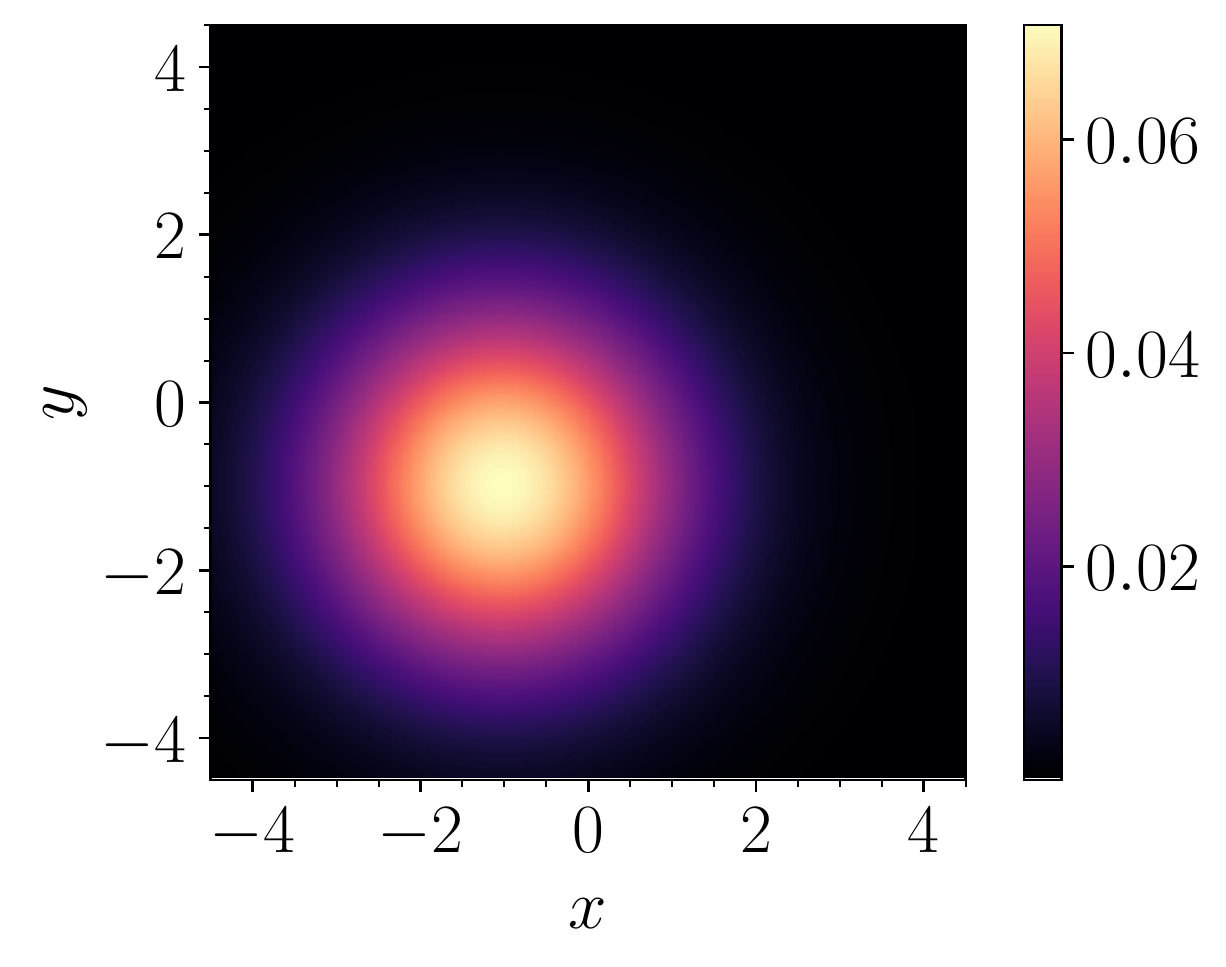}~
 \includegraphics[height=.43\textwidth]{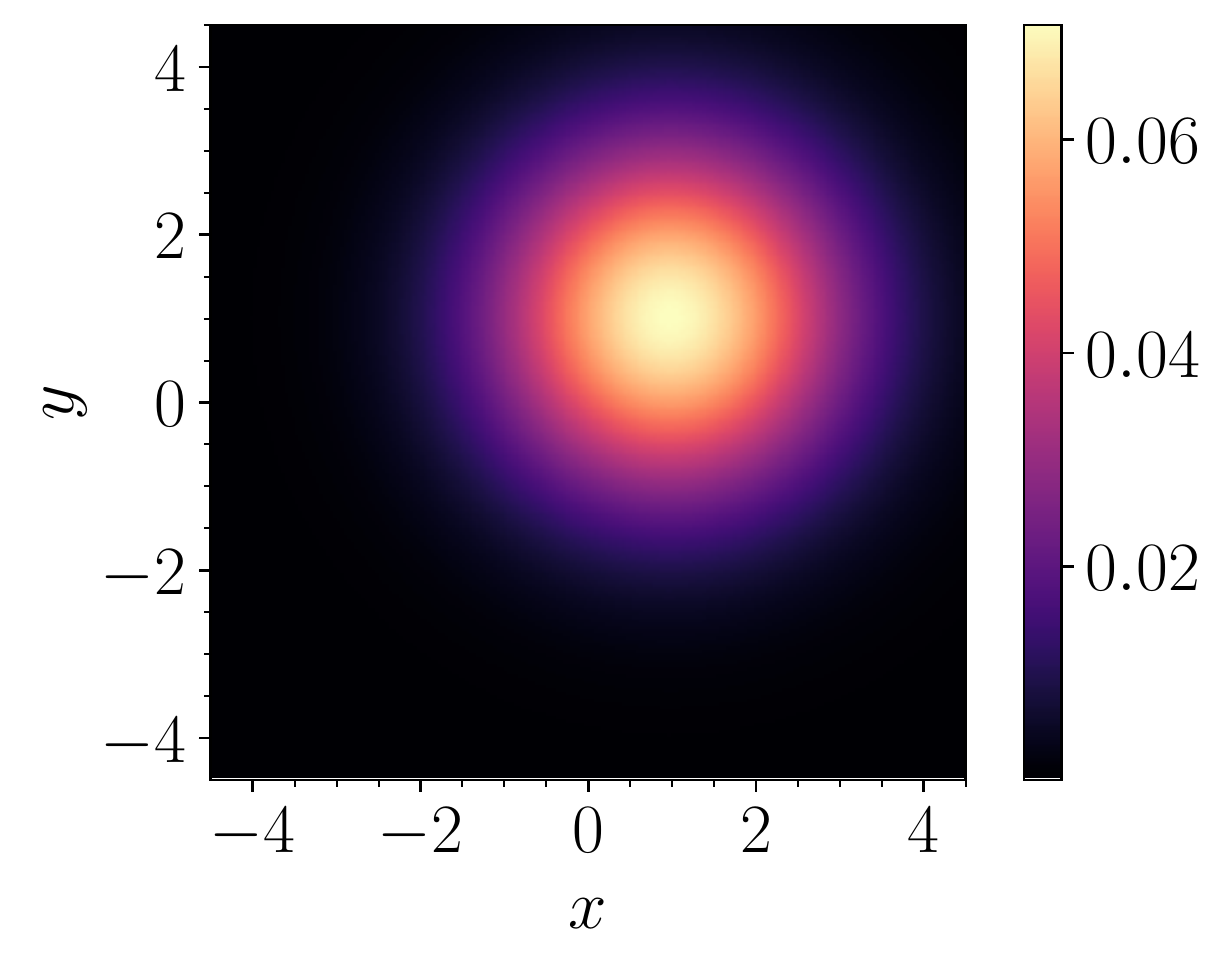}
 \caption{Heatmaps of the normalized joint distributions of $(x,y)$ under component 1 (left panel) and component 2 (right panel) for the example considered in \sref{subsec:bivariate_gaussian}.} \label{fig:bivariate_gaussian_comps}
\end{figure}

\begin{figure}[t!]
 \centering
 \includegraphics[width=.6\textwidth]{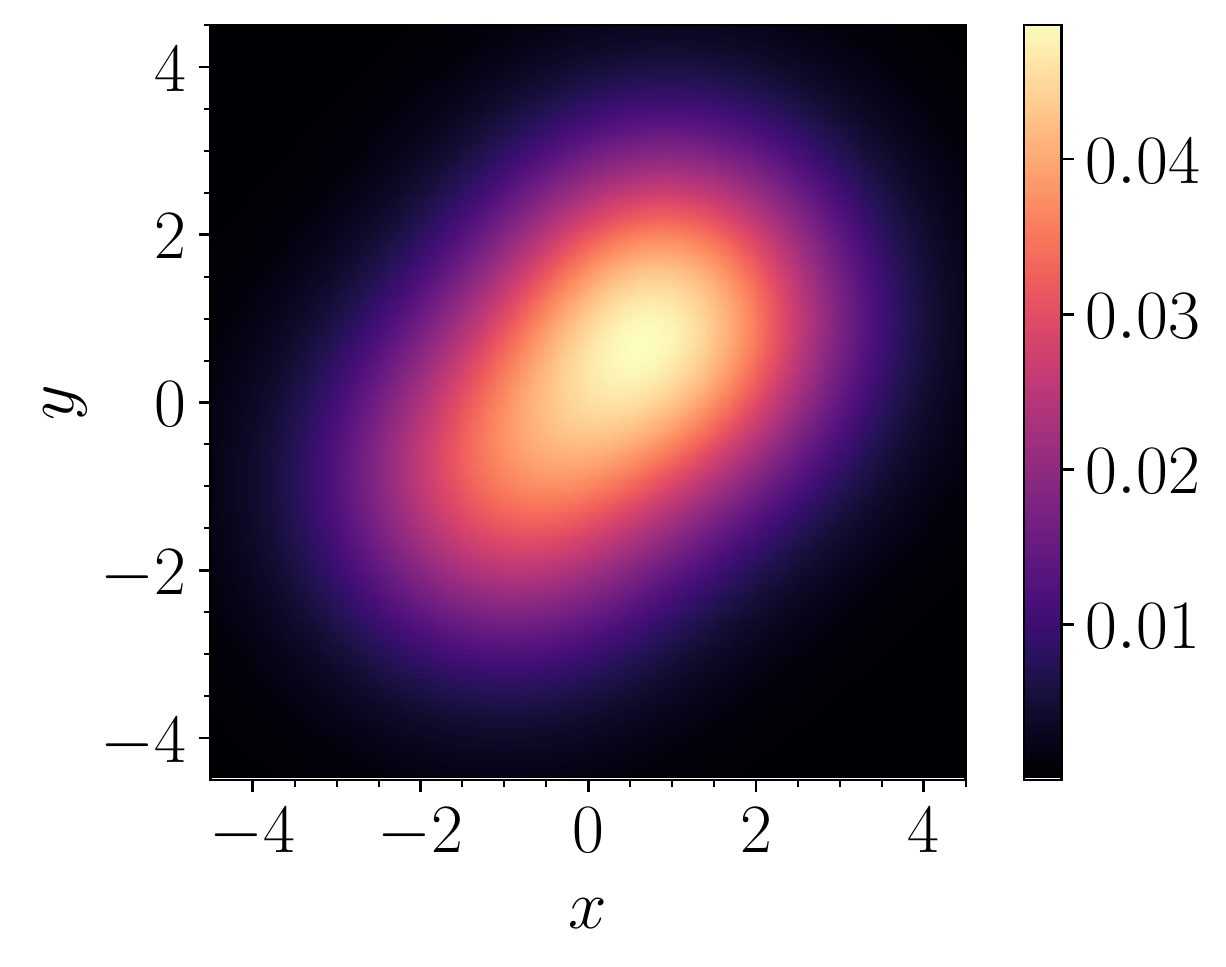}
 \caption{Heatmap of the normalized joint distribution of $(x,y)$ under the mixture model defined in \tref{tab:bivariate_gaussian}.} \label{fig:bivariate_gaussian_combined}
\end{figure}

In the first example, we consider the mixture of two independent bivariate Gaussians. In the first component, $x$ and $y$ are both (independently) normally distributed with mean $-1$ and standard deviation $1.5$. The second component is identical, except $x$ and $y$ both have mean $+1$. The mixture weights are taken to be $w_1 = 0.4, w_2 = 0.6$. \Tref{tab:bivariate_gaussian} summarizes the mixture model specification and \fref{fig:bivariate_gaussian_comps} shows the normalized joint distributions of $(x,y)$ under each of the two components as heatmaps. \Fref{fig:bivariate_gaussian_combined} shows the normalized joint distribution of $(x,y)$ under the mixture model and our InClass net will estimate the mixture model based on data generated as per this distribution.
\begin{table}[t!]
 \centering
 \begin{tabular}{c c c c}
 \toprule[1.5\heavyrulewidth]
 $i$ & $w_i$ & $f^{(i)}_x$ & $f^{(i)}_y$\\
 \midrule[1.5\heavyrulewidth]
 $1$ & $0.4$ & $\mathcal{N}(\text{mean}=-1, \text{SD}=1.5)$ & $\mathcal{N}(\text{mean}=-1, \text{SD}=1.5)$\\
 \cmidrule(lr){1-4}
 $2$ & $0.6$ & $\mathcal{N}(\text{mean}=+1, \text{SD}=1.5)$ & $\mathcal{N}(\text{mean}=+1, \text{SD}=1.5)$\\
 \bottomrule[1.5\heavyrulewidth]
 \end{tabular}
 \caption{The mixture model specification for the example considered in \sref{subsec:bivariate_gaussian}.} \label{tab:bivariate_gaussian}
\end{table}

The classifier networks $\beta^{(i)}_x$ and $\beta^{(i)}_y$ were constructed using \texttt{keras} with the \texttt{tensorflow} backend. The neural networks are distinct, but have identical architectures. The networks are fairly simple, consisting of three sequential dense layers of 32 nodes using the rectified linear unit (ReLU) \cite{10.5555/3104322.3104425} activation function. The output layer is a dense layer with 2 nodes (since $C=2$), with the softmax activation function. The individual classifier networks were then wrapped into an InClass net using the \raindances{} package. The resulting network has a total of 4,484 trainable parameters.

We trained the InClass net to minimize the negative cross mutual information cost function, using 100,000 datapoints sampled from the distribution depicted in \fref{fig:bivariate_gaussian_combined}. The optimization was performed for 15 epochs with the \texttt{Adam} \cite{adam} optimizer (with default hyperparameters) using a batch size of 50. After training the network, we used the same dataset to estimate the pseudo weights $\varphi^{(i)}_{x}$ and $\varphi^{(i)}_{y}$ and the mixture weights $w_i$ using \eqref{eq:bi_pseudo} and \eqref{eq:bi_w_final}. Note that the estimation of mixture models can only be performed up to permutations of the components indexed by $i$. For clarity of the presentation, unless otherwise stated, the components of the true mixture model will be matched with the respective closest candidates from the machine-learned components. The results of the estimation of the mixture weights are summarized in \Tref{tab:bivariate_gaussian_weights}, which demonstrates an excellent agreement between the true and estimated values. 
\begin{table}
 \centering
 \begin{tabular}{c c c c c}
 \toprule[1.5\heavyrulewidth]
 $i$ & Estimated $\varphi^{(i)}_x$ & Estimated $\varphi^{(i)}_y$ & Estimated $w_i$ & True $w_i$ \\
 \midrule[1.5\heavyrulewidth]
 $1$ & $0.4055$ & $0.4048$ & $0.4051$ & $0.4$ \\
 \cmidrule(lr){1-5}
 $2$ & $0.5945$ & $0.5952$ & $0.5949$ & $0.6$ \\
 \bottomrule[1.5\heavyrulewidth]
 \end{tabular}
 \caption{Results of the estimation of the mixture weights for the example considered in \sref{subsec:bivariate_gaussian}. } \label{tab:bivariate_gaussian_weights}
\end{table}

\begin{figure}[t]
 \centering
 \includegraphics[width=.47\textwidth]{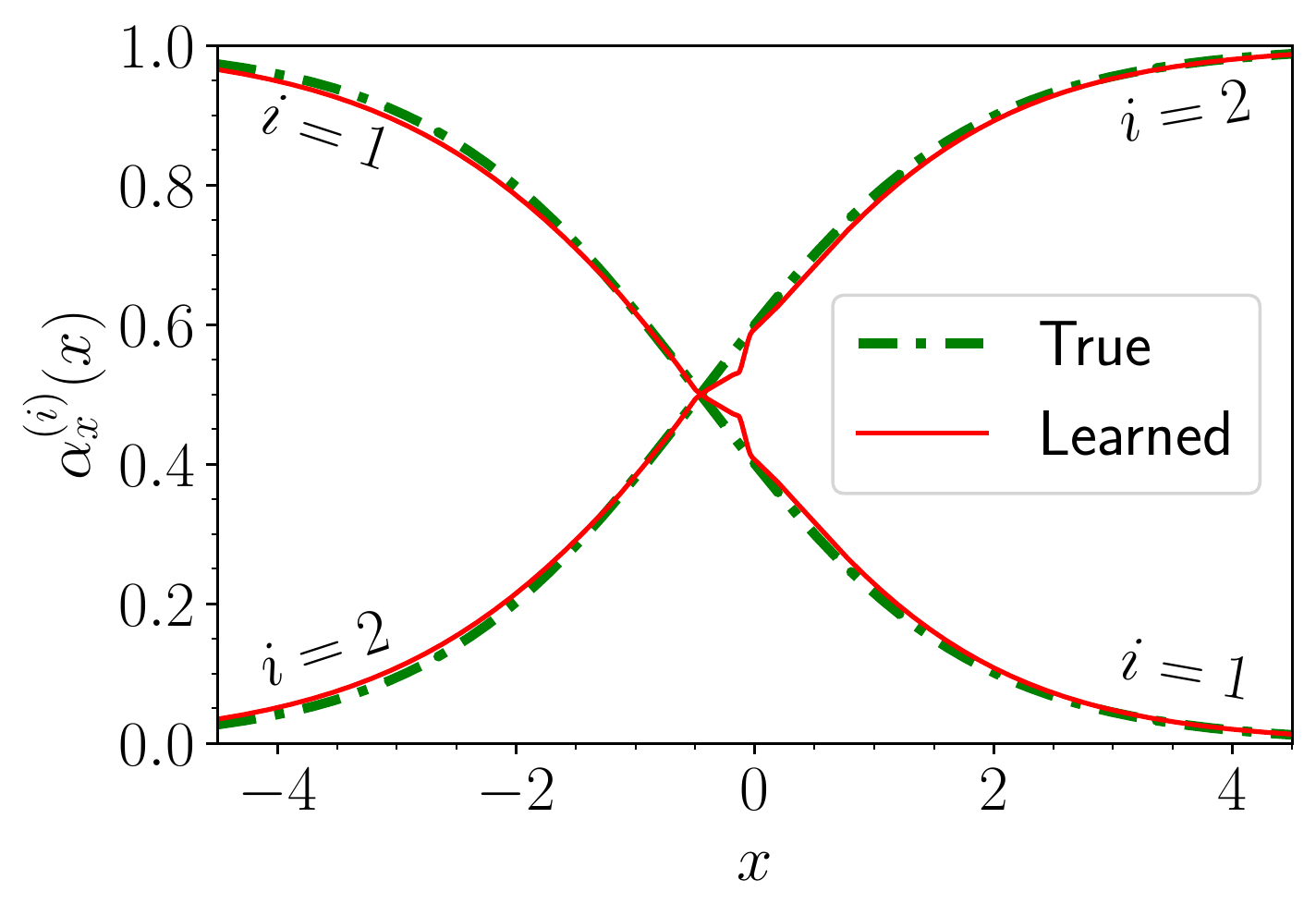}~~
 \includegraphics[width=.47\textwidth]{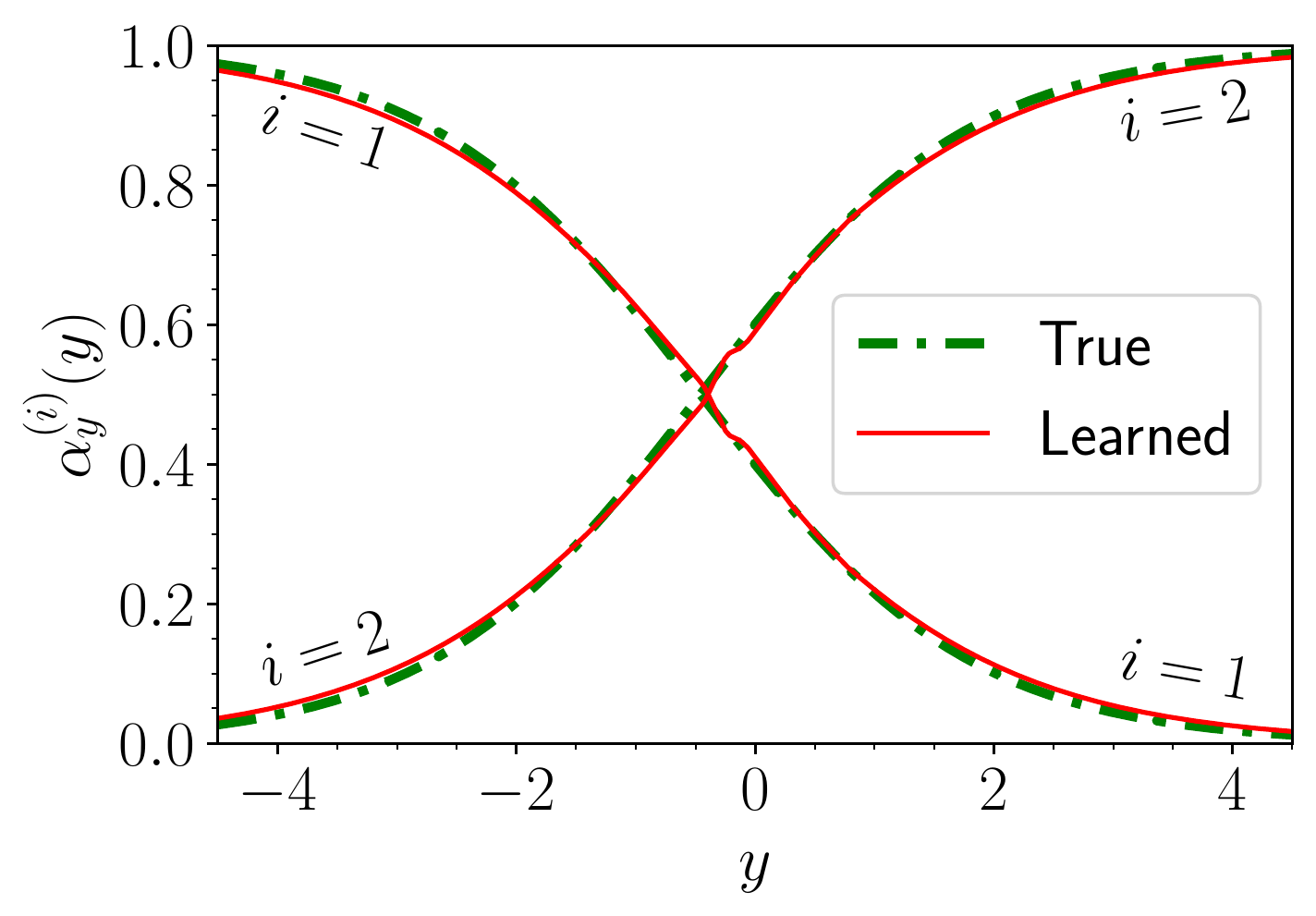}
 \caption{The classifiers $\alpha^{(i)}_{x}(x)$ (left panel) and $\alpha^{(i)}_{y}(y)$ (right panel) learned by the network (red solid lines) and
 the corresponding true classifiers (green dash-dot lines). } \label{fig:bivariate_gaussian_classifiers}
\end{figure}

The solid red curves in \fref{fig:bivariate_gaussian_classifiers} depict the classifiers $\alpha^{(i)}_{x}(x)$ (left panel) and $\alpha^{(i)}_{y}(y)$ (right panel) learned by the network---they are extracted from $\beta^{(i)}_{x}$ and $\beta^{(i)}_{y}$ with the help of \eqref{eq:bi_CIC}. For comparison, the true classifiers based on the exact functional forms of the component distributions are also shown as green dash-dot curves. The red solid lines and the green dash-dot lines almost coincide, which validates our method.

\begin{figure}[t!]
 \centering
 \includegraphics[width=.47\textwidth]{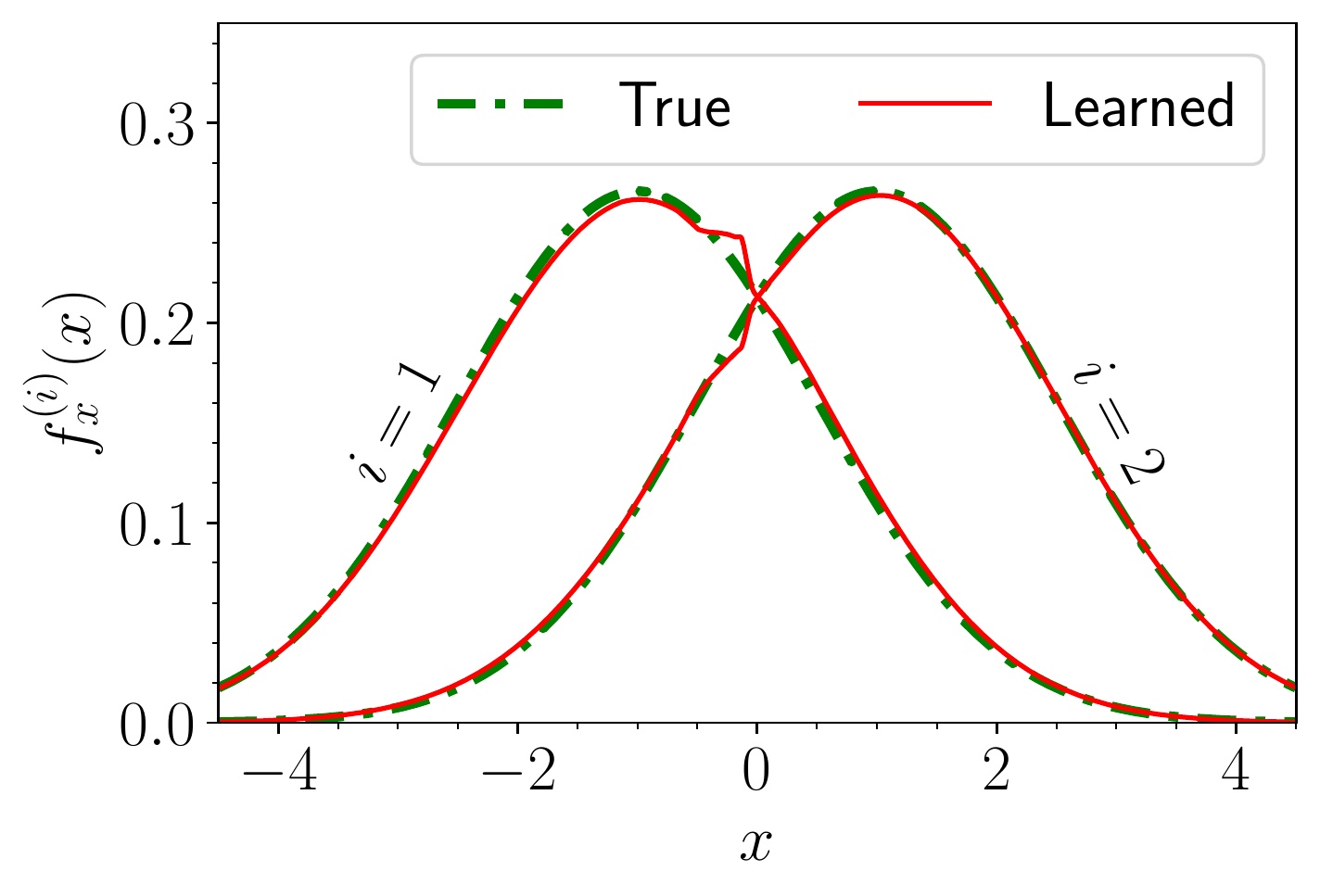}~~
 \includegraphics[width=.47\textwidth]{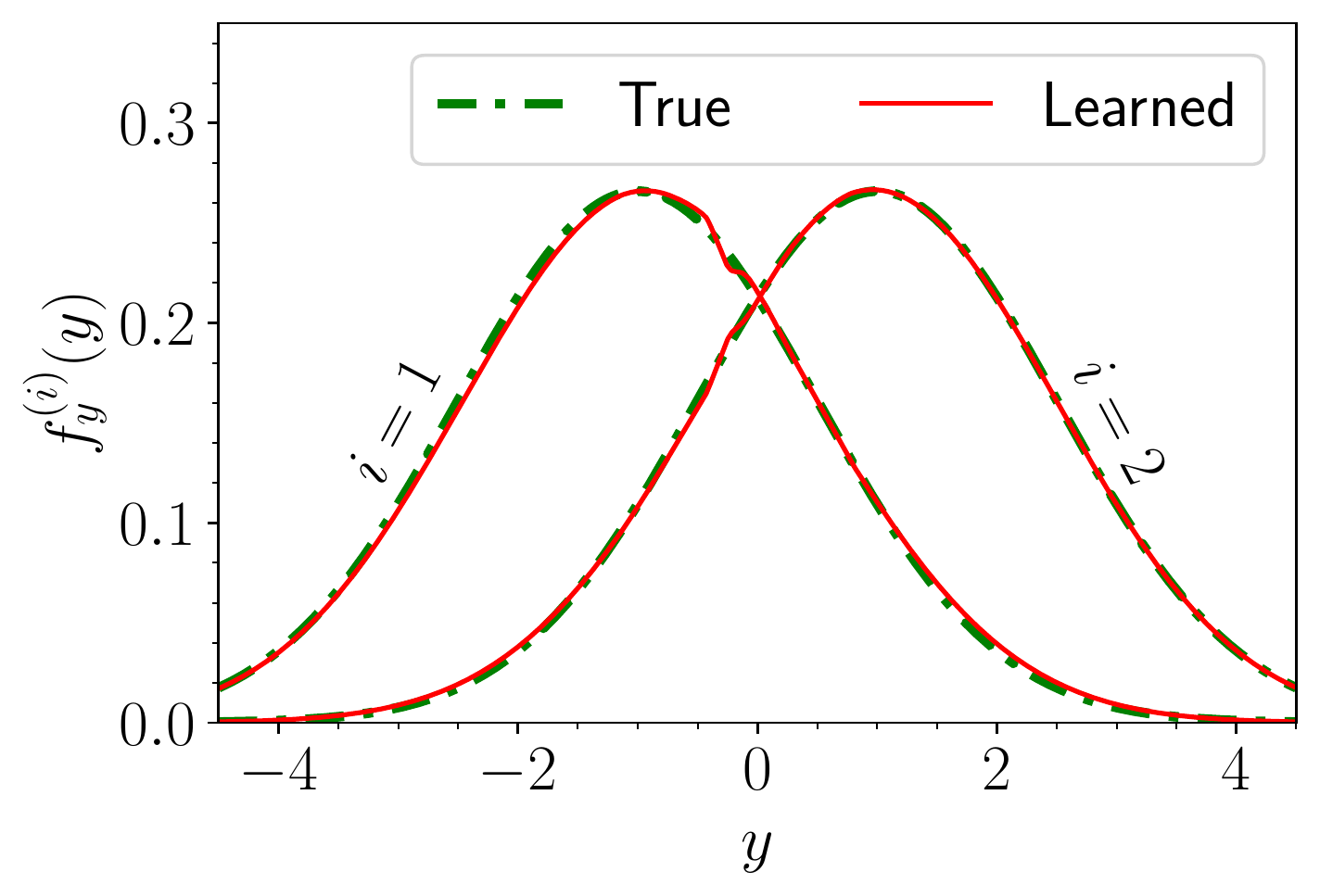}
 \caption{The distributions $f^{(i)}_x$ (left panel) and $f^{(i)}_y$ (right panel). The estimated (true) distributions are shown with red solid (green dash-dot) lines.} \label{fig:bivariate_gaussian_dists}
\end{figure}
\begin{figure}[t!]
 \centering
 \includegraphics[height=.4\textwidth,trim=0 0 65 0,clip]{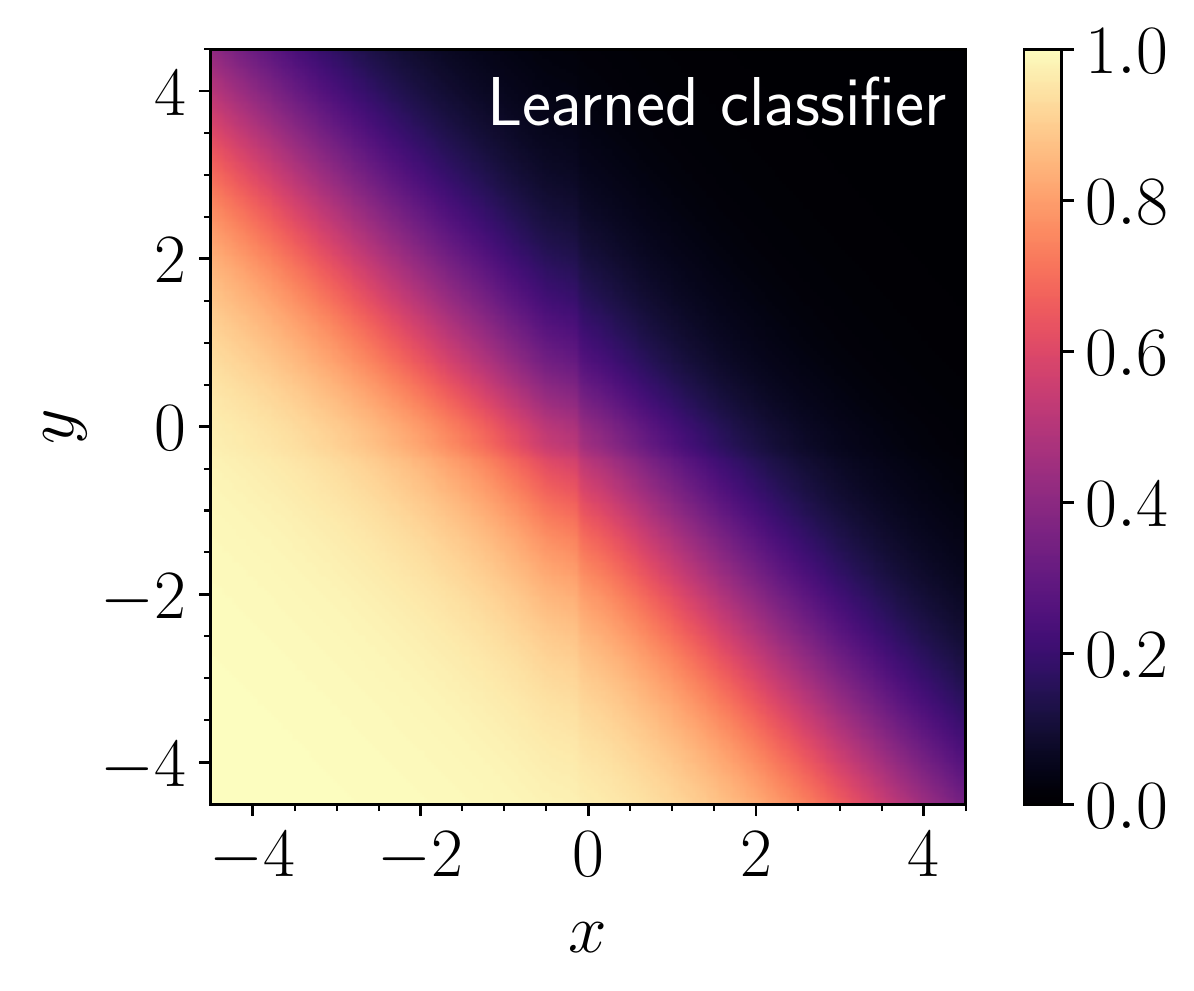}\qquad
 \includegraphics[height=.4\textwidth]{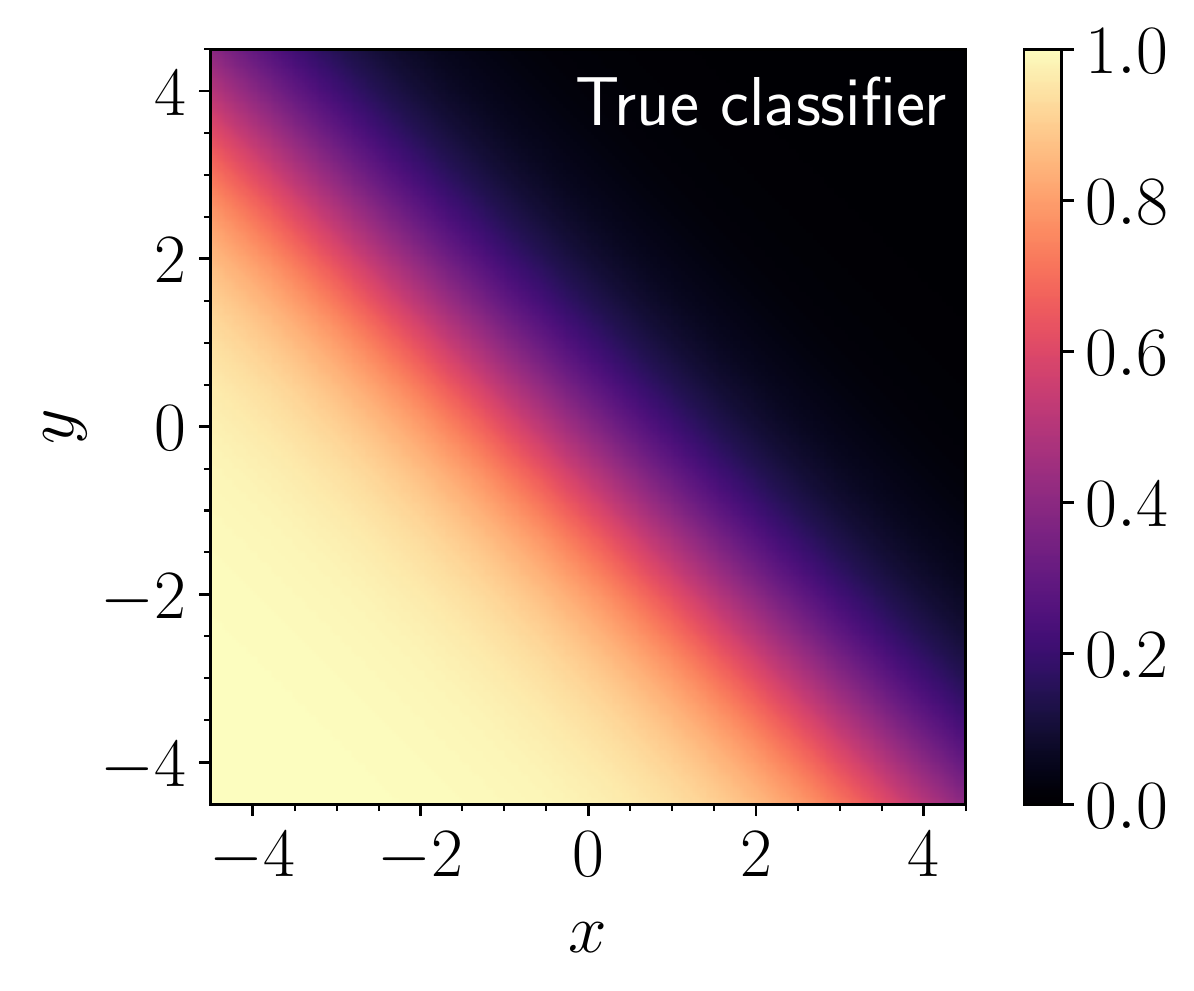}
 \caption{The estimated aggregate classifier $\alpha^{(i)}_\text{aggregate}(x,y)$ from \eqref{eq:bi_aggregate} (left panel) and the true aggregate classifier (right panel). } \label{fig:bivariate_gaussian_aggregate}
\end{figure}

Next, we used \eqref{eq:bi_f_final} to estimate the distributions $f^{(i)}_x$ and $f^{(i)}_y$. The resulting distributions are shown with red solid lines in the left and right panels of \fref{fig:bivariate_gaussian_dists}, respectively. In applying \eqref{eq:bi_f_final}, for simplicity we used the exact expressions for the marginal distributions of $x$ and $y$ in the mixture. In a typical example, the exact expressions for the marginals will not be available, but can be easily estimated from the data, say using a histogram or kernel density estimation \cite{rosenblatt1956,parzen1962}. In \fref{fig:bivariate_gaussian_dists}, we also show the true $f^{(i)}_x$ and $f^{(i)}_y$ as green dash-dot curves. The good agreement between the true $w_i$, $f^{(i)}_{x}$, and $f^{(i)}_{y}$ and their estimates shown in \tref{tab:bivariate_gaussian_weights} and \fref{fig:bivariate_gaussian_dists}, demonstrates that the InClass net has successfully estimated the mixture model. Finally, we use \eqref{eq:bi_aggregate} to estimate the aggregate classifier $\alpha^{(i)}_\text{aggregate}(x,y)$ which is shown as a heatmap in the left panel of \fref{fig:bivariate_gaussian_aggregate}. For comparison, in the right panel we show the true aggregate classifier based on the exact functional forms of the component distributions $f^{(i)}(x,y)$. As expected, the two heatmaps are in very good agreement.

\subsection{The Checkerboard Mixture \texorpdfstring{$(V=2, C=2)$}{(V=2, C=2)}}
\label{sec:checkerboard}

\begin{figure}[t]
 \centering
 \includegraphics[height=.4\textwidth]{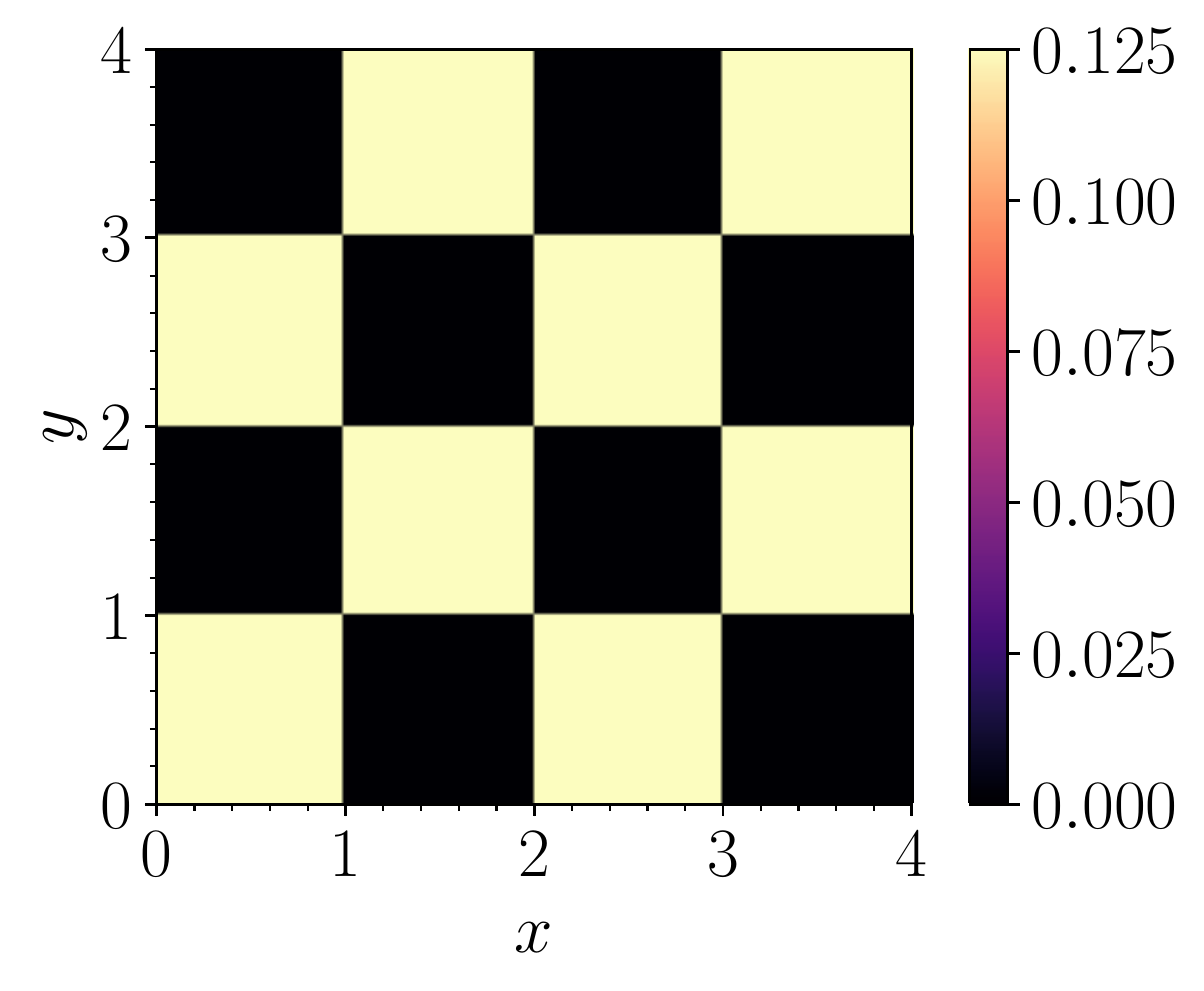}
 \caption{Heatmap illustrating the joint distribution of $(x,y)$ for the ``checkerboard'' mixture example considered in \sref{sec:checkerboard}. The datapoints are uniformly distributed on the bright squares of a $4\times4$ checkerboard spanning the region $0\leq x, y < 4$, while the dark squares have zero density.} \label{fig:checkerboard_combined}
\end{figure}

\begin{figure}[t!]
 \centering
 \includegraphics[height=.4\textwidth,trim=0 0 70 0,clip]{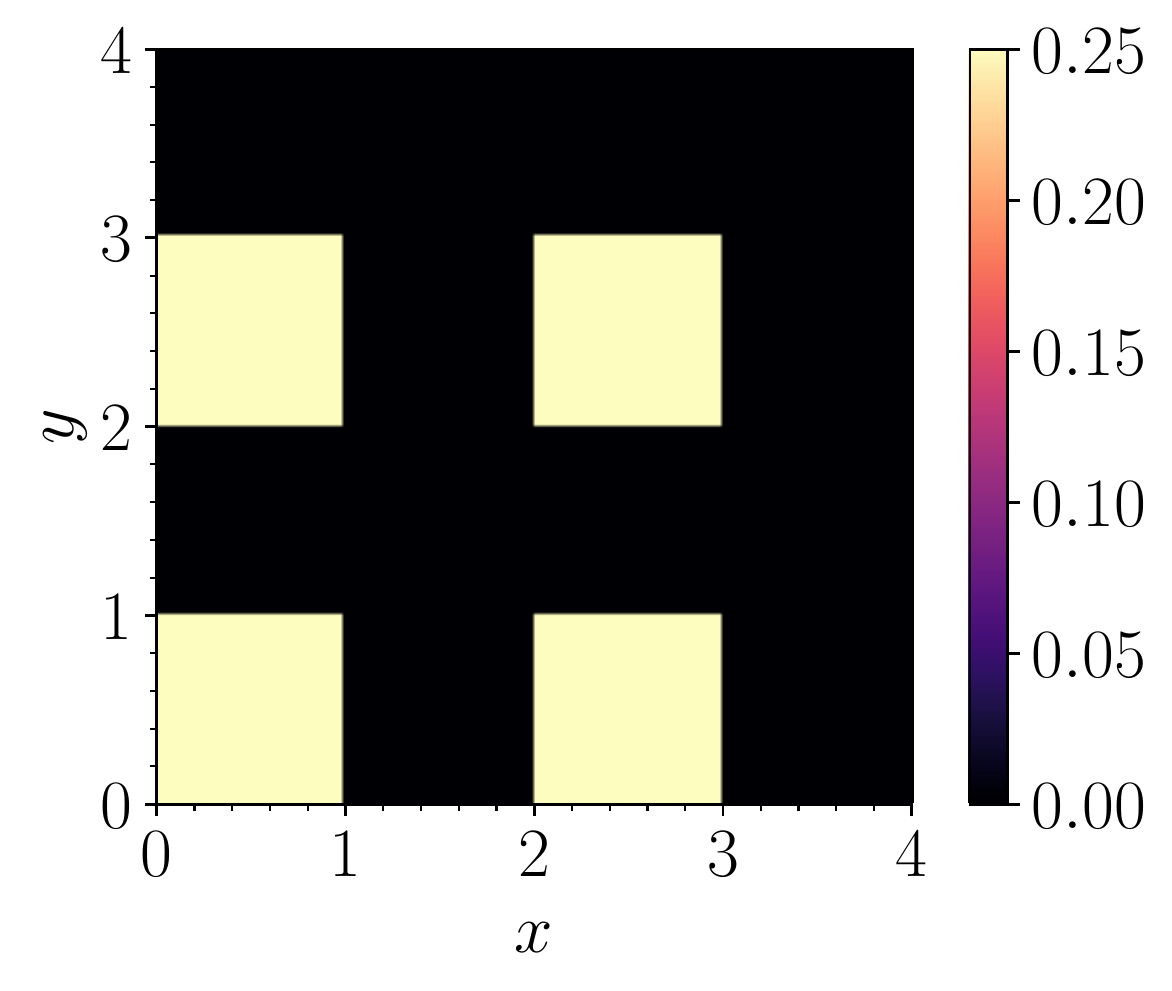}\qquad
 \includegraphics[height=.4\textwidth]{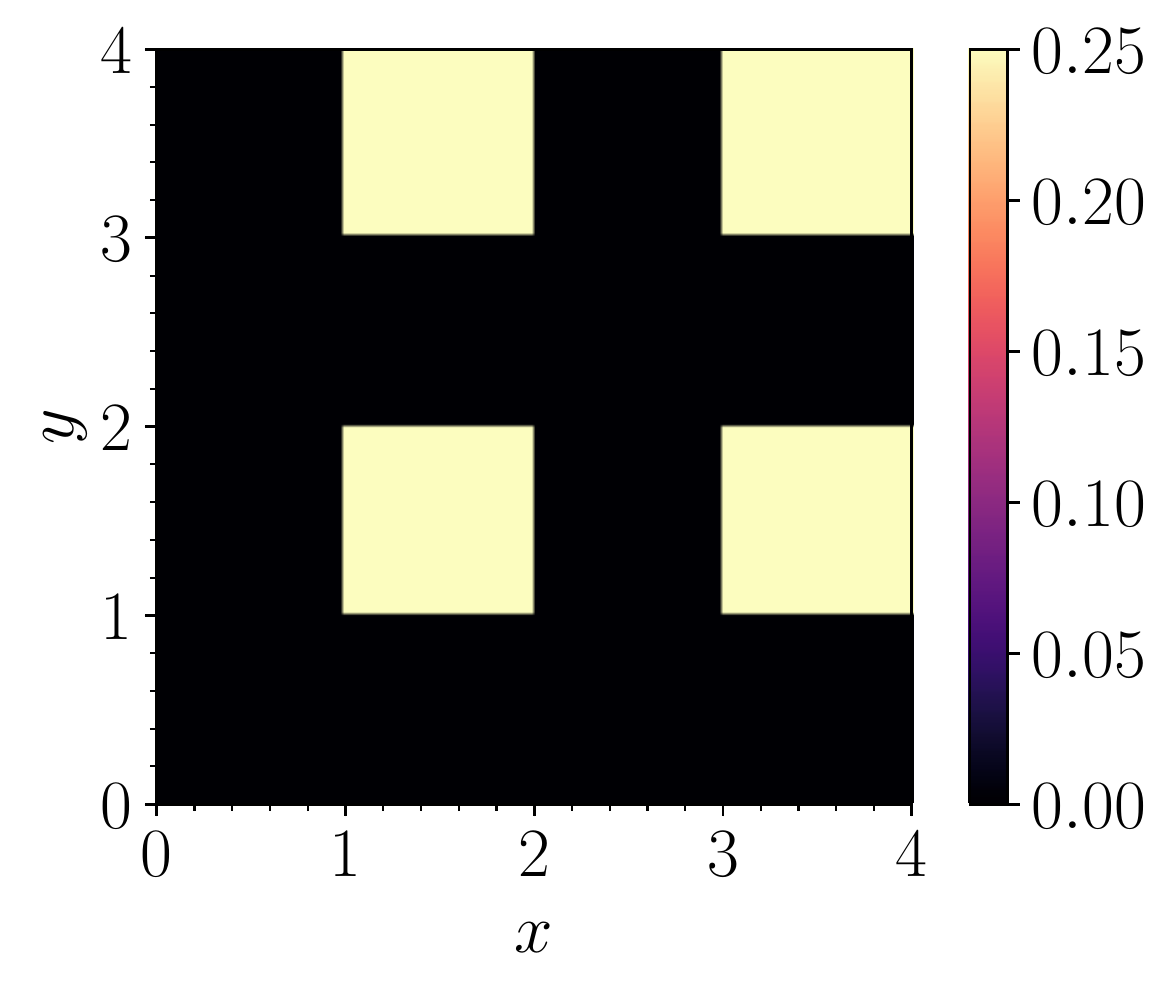}
 \caption{Heatmaps of the normalized joint distributions of $(x,y)$ under component 1 (left panel) and component 2 (right panel) for the ``checkerboard'' mixture shown in \fref{fig:checkerboard_combined}.} \label{fig:checkerboard_comps}
\end{figure}

Now we will look at a toy example which was instrumental in the conception and development of the InClass nets technique, see Figures~\ref{fig:checkerboard_combined} and \ref{fig:checkerboard_comps}. \Fref{fig:checkerboard_combined} shows the joint distribution of $(x,y)$ for a ``checkerboard'' mixture under which the datapoints are uniformly distributed on the bright squares of a $4\times4$ checkerboard spanning the region $0\leq x, y < 4$, while the dark squares have zero density. For concreteness, the vertical (horizontal) boundaries between cells are assigned to the cell on the right (top). It is easy to see that $x$ and $y$ are, individually, uniformly distributed between $0$ and $4$. It can also be seen that $x$ and $y$, despite being uncorrelated, are not mutually independent in the mixture, since x lies within $[0, 1) \cup [2, 3)$ if and only if $y$ does as well.

As shown in \fref{fig:checkerboard_comps}, the checkerboard mixture can be separated into two equally weighted components within which $x$ and $y$ are mutually independent. Under the first component, $x$ and $y$ both lie within $[0, 1) \cup [2, 3)$, and under the second component $x$ and $y$ both lie within $[1, 2) \cup [3, 4)$. Note that each of these components has four spatially disconnected regions---the classification cannot be achieved using spatial clustering techniques. This example also naturally evokes the intuition of the variates $x$ and $y$ serving as each other's supervisory signal, since the value of either $x$ or $y$ uniquely determines the component the datapoint belongs to.

\begin{figure}[t!]
 \centering
 \includegraphics[width=.45\textwidth]{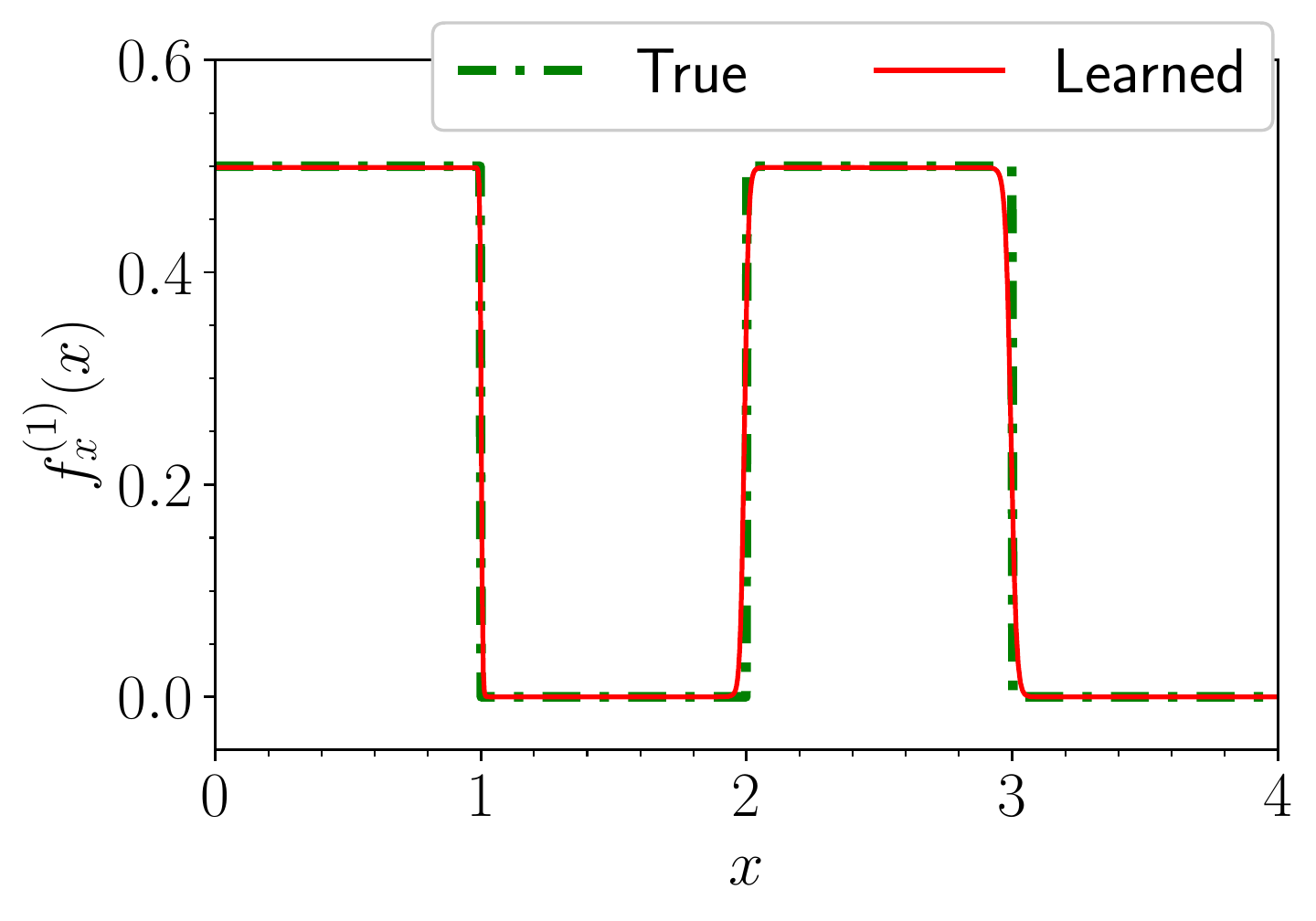}~~
 \includegraphics[width=.45\textwidth]{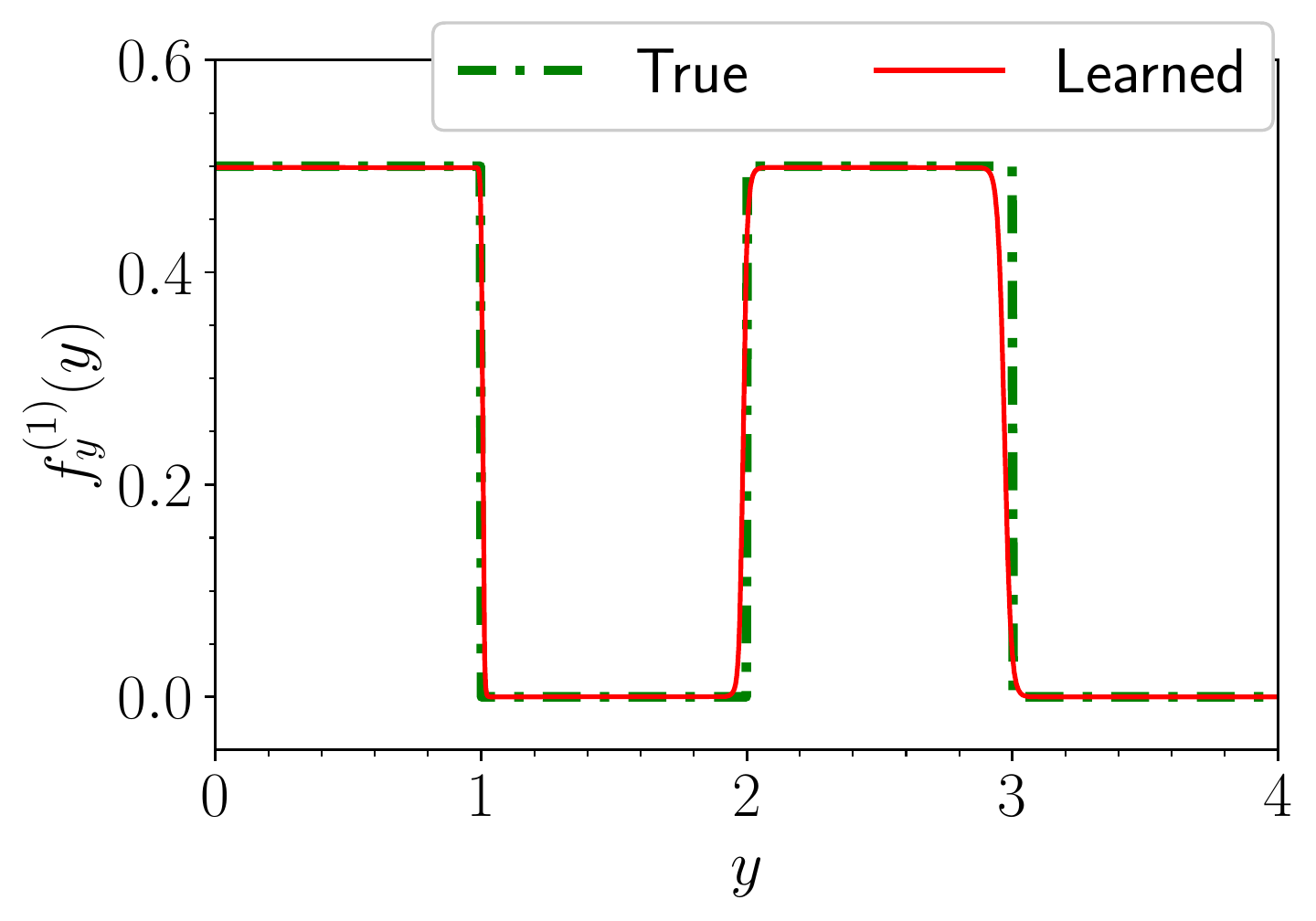}
 \caption{The distributions $f^{(1)}_x$ (left panel) and $f^{(1)}_y$ (right panel) for the ``checkerboard'' example considered in \sref{sec:checkerboard}. The estimated (true) distributions are shown with red solid (green dash-dot) lines. We only show the first component in this figure for the sake of visual clarity (see text).} \label{fig:checkerboard_dists}
\end{figure}

Let us now analyze this toy example using an InClass net. All the details of the network training process are identical to the analysis of the example in \sref{subsec:bivariate_gaussian}, including the network architectures, the size of the training dataset, the choice of optimizer, batch size and epoch count. The estimated mixture weights are $w_1=0.501, w_2=0.499$, which is in excellent agreement with their true values of $w_1 = w_2 = 0.5$. \Fref{fig:checkerboard_dists} shows, in solid red curves, the distributions of $x$ (left panel) and $y$ (right panel) under the first component learned by the network, using the same procedure as in \sref{subsec:bivariate_gaussian}. We only show the first component in this figure for the sake of clarity---the second component fills the gaps in the univariate distributions of $x$ and $y$ so that $w_1\,f^{(1)}_{x} + w_2\,f^{(2)}_{x}$ and $w_1\,f^{(1)}_{y} + w_2\,f^{(2)}_{y}$ are constant. For comparison, the true ``rectangular wave'' distributions are also shown as green dash-dot curves, which are also seen to agree with the estimates.
\subsection{Semi-Supervised Training on MNIST Data \texorpdfstring{$(V=2, C=10)$}{(V=2, C=10)}}
\label{sec:MNIST}

The biggest advantage offered by a machine learning based technique over existing non-machine-learning techniques for nonparametric mixture model estimation is the possibility of tackling high-dimensional data. As a proof of concept, in this section we will train an InClass net to classify images of handwritten digits from the MNIST database \cite{lecun2010mnist}, with the classes corresponding to the digits $0\text{--}9$. With this example, we will focus more on the data classification aspect of this paper than the mixture model estimation.

\begin{figure}
 \centering
 \includegraphics[width=\textwidth]{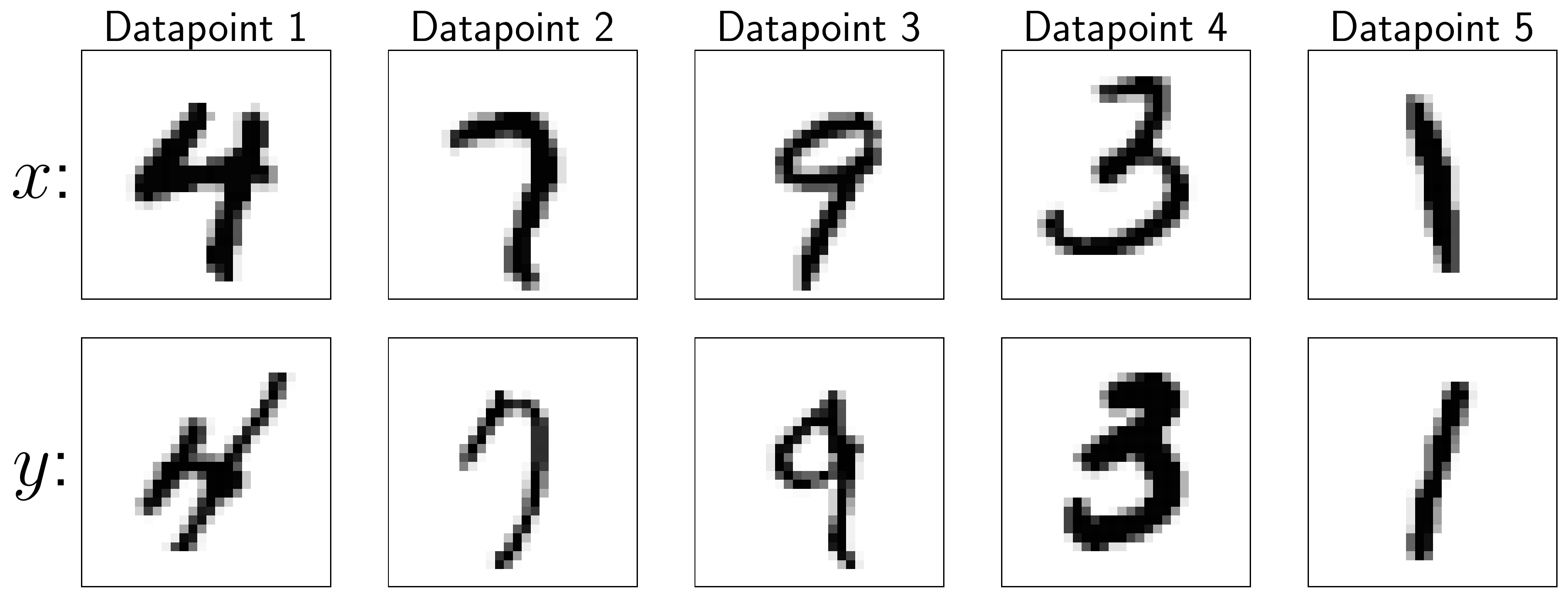}
 \caption{Five representative datapoints from the dataset used to train the InClass net in the example considered in \sref{sec:MNIST}. Each datapoint $(x, y)$ is a pair of images containing the same digit.} \label{fig:digits}
\end{figure}

The MNIST dataset contains $28\text{px}\times28\text{px}$ grayscale images of handwritten digits. Each image also has an associated label indicating the digit contained in the image. We will construct a bivariate mixture model out of the MNIST dataset, where each datapoint is a pair of images. A single datapoint of the dataset will be sampled by first choosing a class between $0\text{--}9$ uniformly at random, and then sampling two images\footnote{Alternatively, one can generate image pairs by sampling the first image from the MNIST dataset, and applying a random transformation on the sampled image to get the second, as in \cite{Xiang2019}.} containing that digit uniformly from the MNIST dataset (with replacement). This gives us a bivariate conditional independence mixture model with 10 classes of equal mixture weights---note that within each component (or class), the two images are mutually independent of each other. \Fref{fig:digits} illustrates the kind of data the InClass net will see, with 5 randomly chosen datapoints from the mixture model (one in each column).

For analyzing this dataset, instead of creating two different classifiers for the variates, we use the same neural network for classifying both $x$ and $y$. Viewed differently, the networks classifying $x$ and $y$ are identical in architecture and share their weights as well, and only differ in the input (output) they receive (return). The network uses a sequential architecture and contains, in order, a layer to flatten the $28\times28$ image data, 3 dense layers each with the ReLU activation function and 32 nodes, and finally a dense layer with the softmax activation function and 10 nodes (since $C=10$). This network has a total of 27,562 trainable parameters.

In principle, an InClass net can be trained without supervision to distinguish the digits. However, considering the large number of input dimensions and classes, without supervision, our network is expected to have difficulties ``discovering'' new classes in the data, and will end up in bad local minima of the cost function. We will discuss some ways of overcoming this difficulty in \aref{appendix:gradient}.

In this example, we addressed this issue by taking a semi-supervised approach: We ``seeded'' the classes (digits) in the network by performing a supervised training over a small dataset with noisy labels. For this purpose, we used a training dataset containing 2,000 images. The noisy label associated with each image matches its true label with probability $0.6$, and matches one of the other 9 incorrect labels (chosen uniformly) with probability $0.4$. The network was trained using the categorical cross-entropy loss function with the \texttt{Adam} optimizer for 30 epochs (batch size 20).

After this pre-training, we trained the network further using our \texttt{neg\_ctc\_cost} function on 100,000 pairs of images from our mixture model\footnote{The 200,000 images were all sampled with replacament from a set of 60,000 total images in the MNIST training dataset---repetitions will occur within the dataset.}. 10\% of the 100,000 datapoints were set aside as a validation dataset to monitor the evolution of the network performance, though no hyperparameter optimization was actively performed using the validation data. The training was done using the \texttt{Adam} optimizer with a batch size of 100 for 20 epochs.

\begin{figure}
 \centering
 \includegraphics[height=.45\textwidth,trim=0 0 65 0,clip]{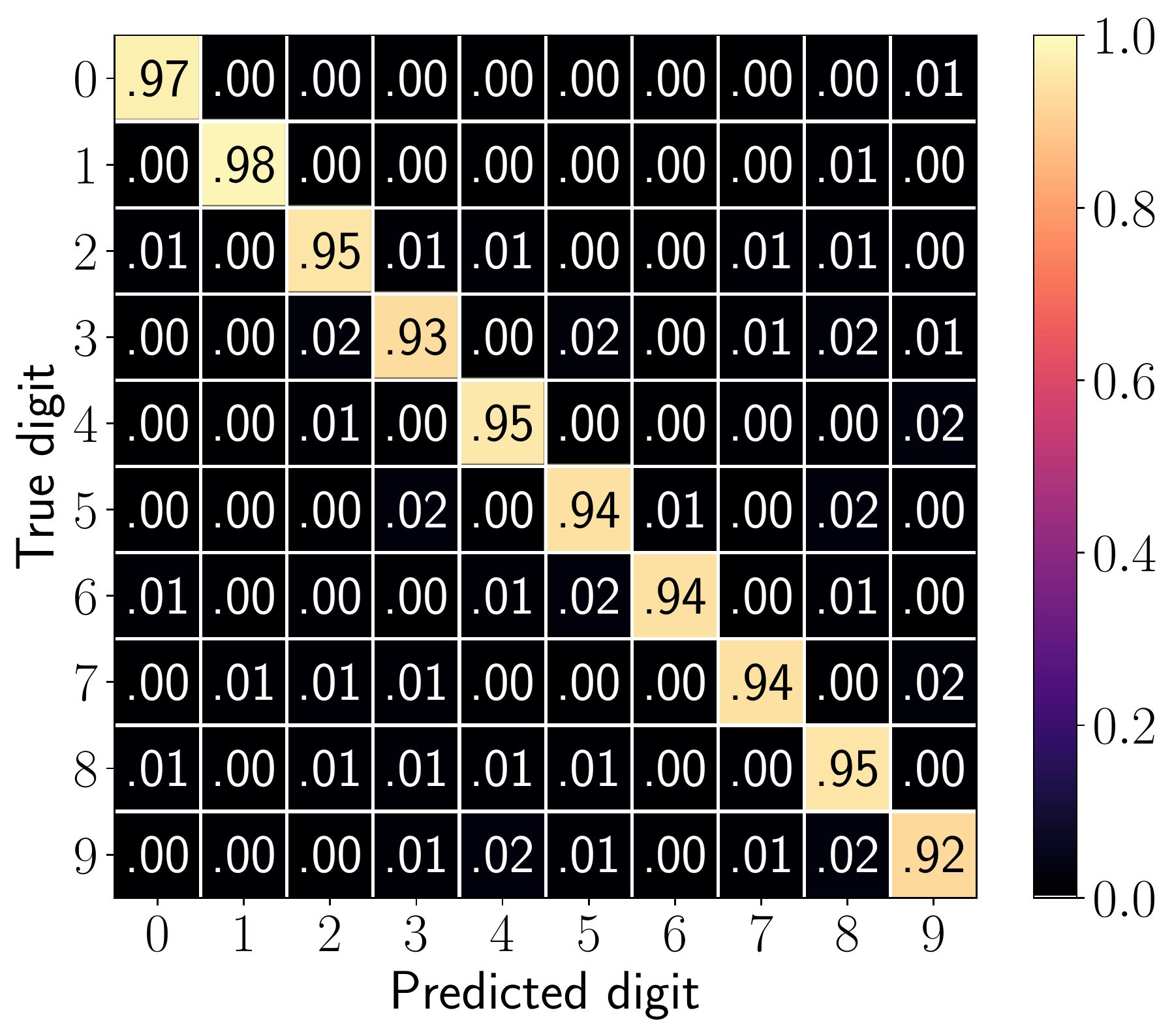}~~
 \includegraphics[height=.45\textwidth]{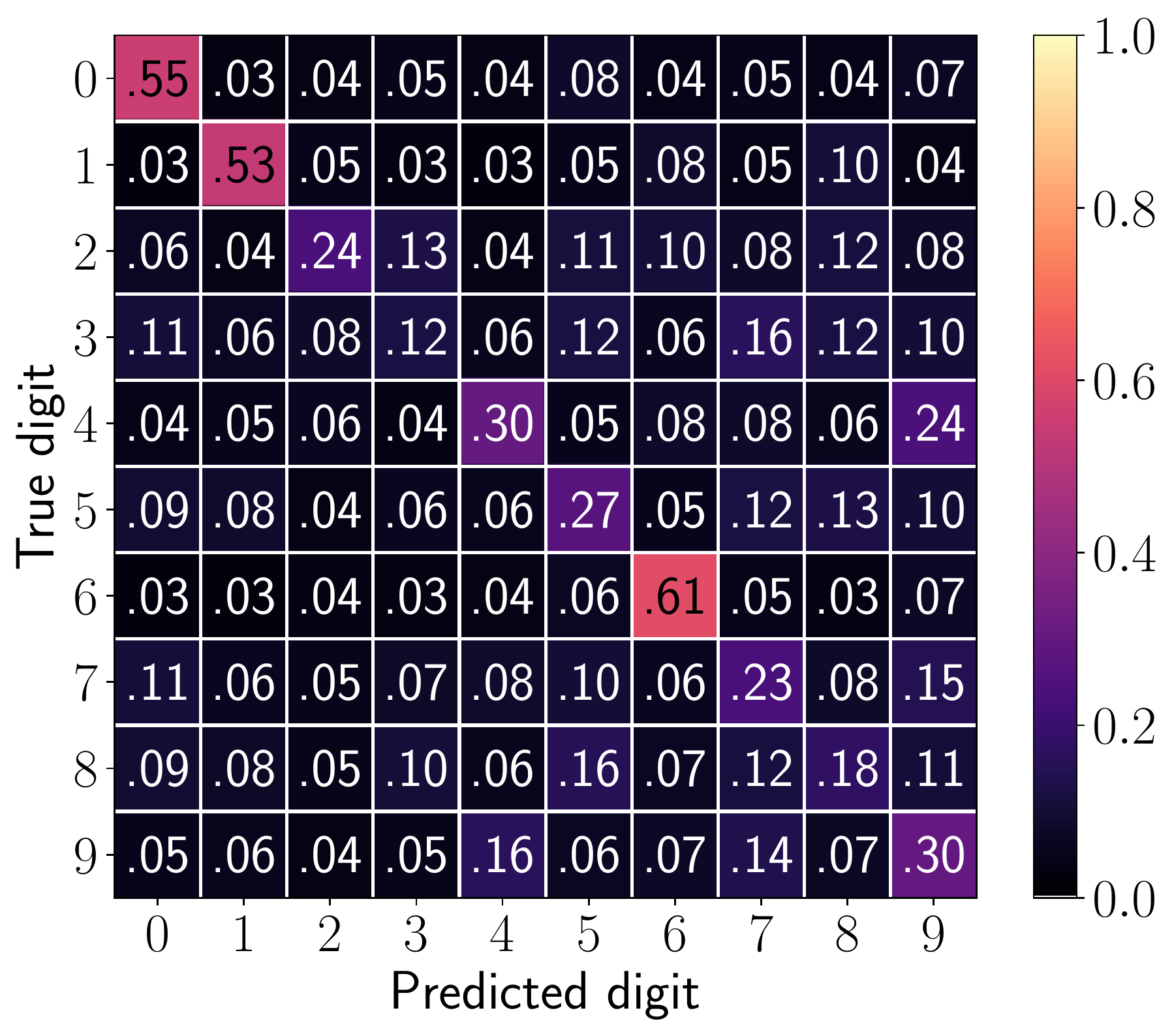}
 \caption{Confusion matrix of the neural network after full training (left panel) and after noisy pre-training (right panel). Each cell of the confusion matrix shows the average probability assigned by the network for images from a given true class ($y$-axis) to belong to a given predicted class ($x$-axis).} \label{fig:confusion}
\end{figure}

Finally, we evaluated the performance of the classifier on a testing dataset of 10,000 single images unseen by the network (either during training or during validation). The performance is illustrated as a confusion matrix in the left panel of \fref{fig:confusion}. Each row of the confusion matrix shows the output of the network averaged over test images containing a given digit (true label), both as a heatmap and as numerical values within each cell of the matrix. Recall that our network output, for each image, is 10-dimensional and can be interpreted as the probabilities assigned by the network to the different classes. Because the classes were pre-seeded into the network in a supervised manner, they matched with the true classes without requiring any manual reassignment.

For comparison, we also show the confusion matrix of the network after the supervised pre-training performed on the noisily labelled data in the right panel of \fref{fig:confusion}. Note that the training that resulted in the performance improvement from the right panel to the left panel was completely unsupervised. 
We will discuss this semi-supervised training approach in the context of real world applications in \sref{subsec:semisupervised}.

\subsection{Mixture of Four Independent Trivariate Gaussians \texorpdfstring{$(V=3, C=4)$}{(V=3, C=4)}} \label{subsec:four_gaussians}
\begin{figure}[t!]
\begin{minipage}[l]{0.47\textwidth}
 \includegraphics[width=\textwidth]{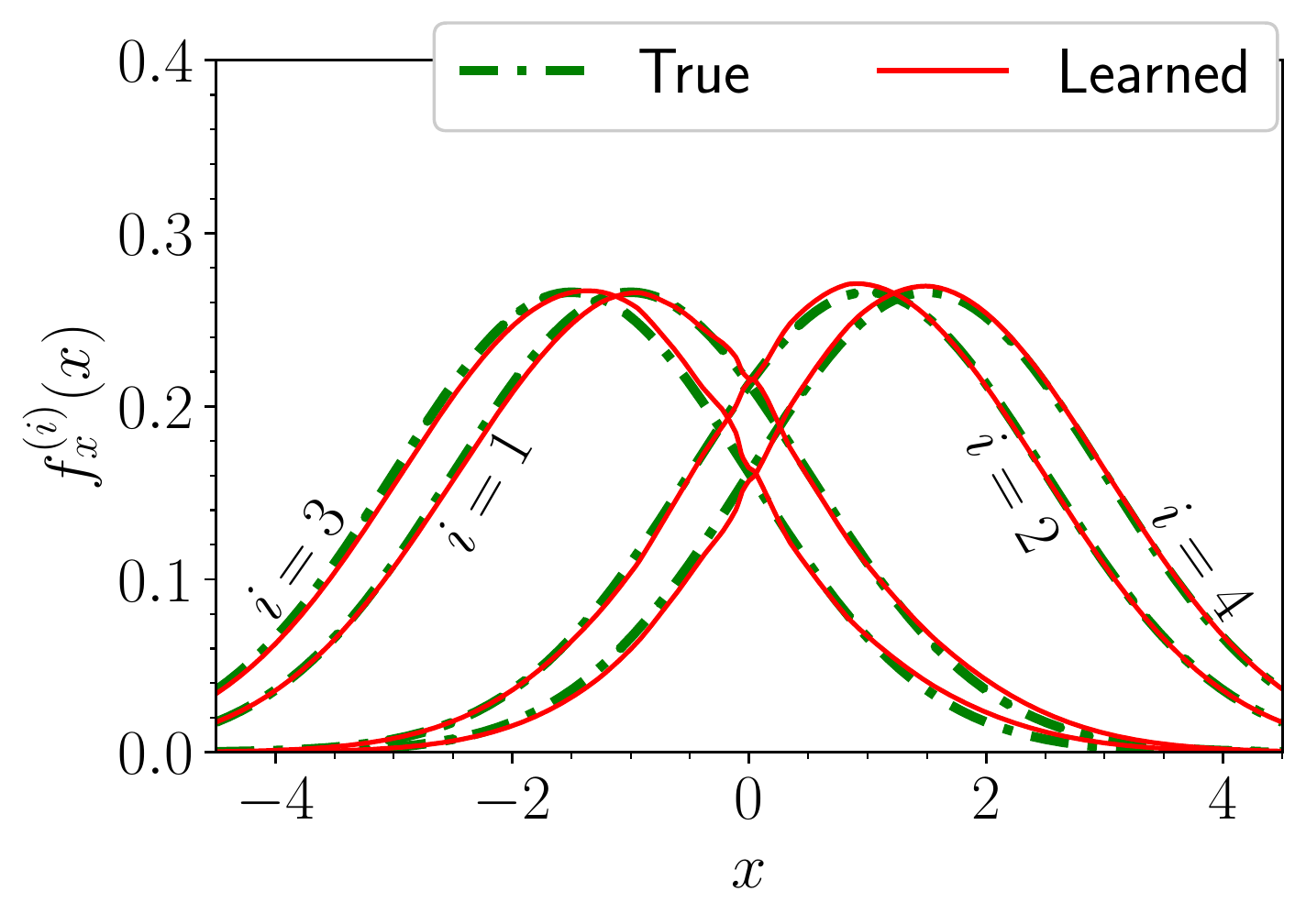}
\end{minipage}
\begin{minipage}[r]{0.47\textwidth}
 \includegraphics[width=\textwidth]{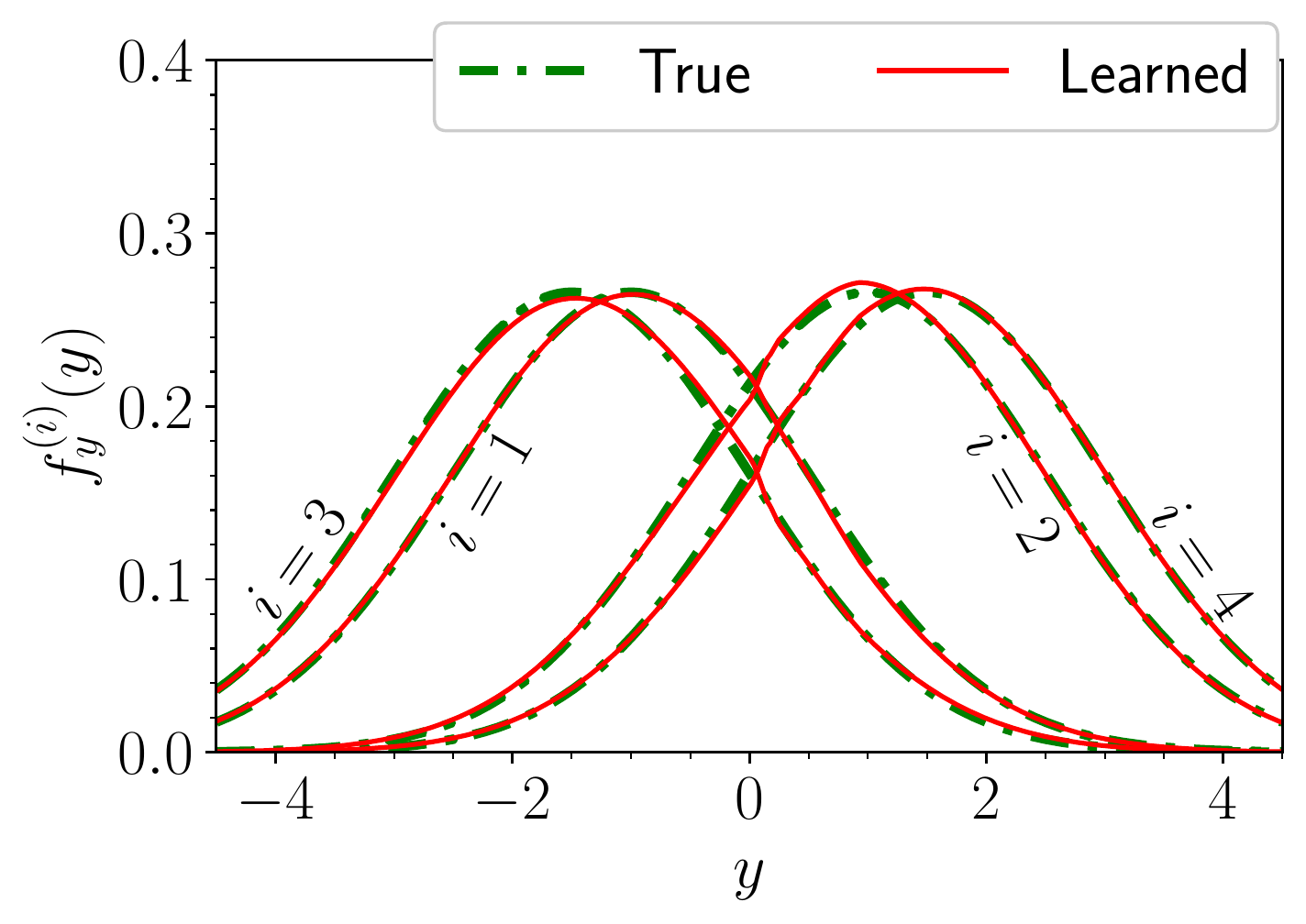}
\end{minipage}

\begin{minipage}[l]{0.47\textwidth}
 \includegraphics[width=\textwidth]{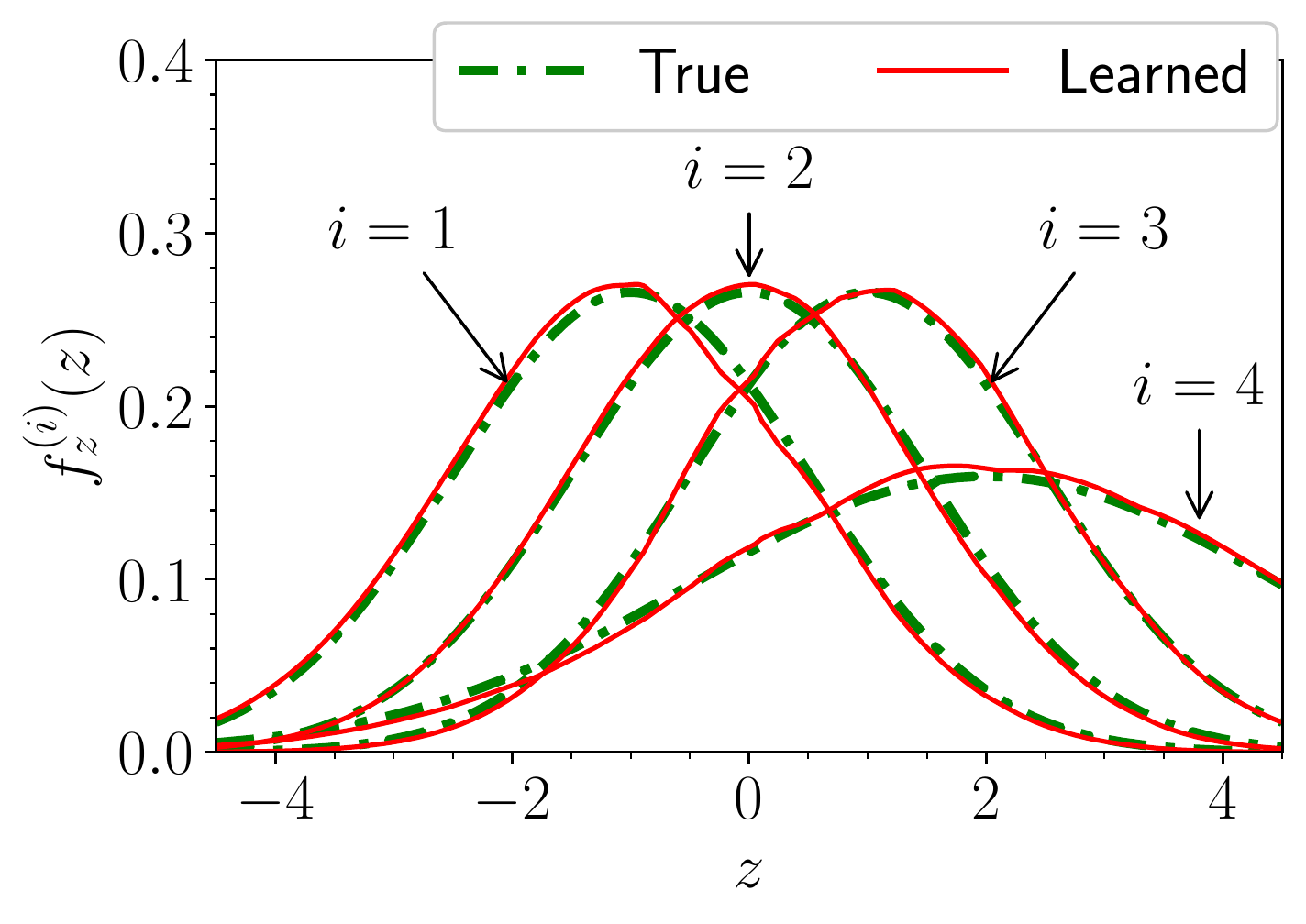}
\end{minipage}
\begin{minipage}[r]{0.47\textwidth}
 \centering
 \vskip -6mm
 \qquad\begin{tabular}{c c c}
 \toprule[1.5\heavyrulewidth]
 $i$ & True $w_i$ & Estimated $w_i$ \\
 \midrule[1.5\heavyrulewidth]
 $1$ & $0.22$ & $0.228$\\
 \cmidrule{1-3}
 $2$ & $0.28$ & $0.268$\\
 \cmidrule{1-3}
 $3$ & $0.18$ & $0.187$\\
 \cmidrule{1-3}
 $4$ & $0.32$ & $0.318$\\
 \bottomrule[1.5\heavyrulewidth]
 \end{tabular}
\end{minipage}
 \caption{The distributions $f^{(i)}_x$ (top-left panel), $f^{(i)}_y$ (top-right panel), and $f^{(i)}_z$ (bottom-left panel) for the example considered in \sref{subsec:four_gaussians}. The estimated (true) distributions are shown with red solid (green dash-dot) lines.} \label{fig:four_gaussians_dists}
\end{figure}

In this example, we will demonstrate that the InClass nets technique works for the estimation of mixture models with more than two variates as well. We consider the mixture of four independent trivariate Gaussians, with the third variate denoted by $z$. \Tref{tab:four_gaussians} summarizes the mixture model specification.
\begin{table}[t!]
 \centering
 \resizebox{\textwidth}{!}{
 \begin{tabular}{c c c c c c}
 \toprule[1.5\heavyrulewidth]
 $i$ & $w_i$ & $f^{(i)}_x$ & $f^{(i)}_y$ & $f^{(i)}_z$ & Estimated $w_i$\\
 \midrule[1.5\heavyrulewidth]
 $1$ & $0.22$ & $\mathcal{N}(\text{mean}=-1, \text{SD}=1.5)$ & $\mathcal{N}(\text{mean}=-1, \text{SD}=1.5)$ & $\mathcal{N}(\text{mean}=-1, \text{SD}=1.5)$ & $0.228$\\
 \cmidrule(lr){1-6}
 $2$ & $0.28$ & $\mathcal{N}(\text{mean}=+1, \text{SD}=1.5)$ & $\mathcal{N}(\text{mean}=+1, \text{SD}=1.5)$ & $\mathcal{N}(\text{mean}=0, \text{SD}=1.5)$ & $0.268$\\
 \cmidrule(lr){1-6}
 $3$ & $0.18$ & $\mathcal{N}(\text{mean}=-1.5, \text{SD}=1.5)$ & $\mathcal{N}(\text{mean}=+1.5, \text{SD}=1.5)$ & $\mathcal{N}(\text{mean}=+1, \text{SD}=1.5)$ & $0.187$\\
 \cmidrule(lr){1-6}
 $4$ & $0.32$ & $\mathcal{N}(\text{mean}=+1.5, \text{SD}=1.5)$ & $\mathcal{N}(\text{mean}=-1.5, \text{SD}=1.5)$ & $\mathcal{N}(\text{mean}=+2, \text{SD}=2.5)$ & $0.318$\\
 \bottomrule[1.5\heavyrulewidth]
 \end{tabular}}
 \caption{The mixture model specification for the example considered in \sref{subsec:four_gaussians}. The last column shows the mixture weights estimated by the InClass net technique.} \label{tab:four_gaussians}
\end{table}
The classifier networks $\beta^{(i)}_x$, $\beta^{(i)}_y$, and $\beta^{(i)}_z$ have a similar architecture to the classifier architectures used in \sref{subsec:bivariate_gaussian}, except that the output layer has 4 nodes, since $C=4$. The InClass net constructed out of the classifiers has a total of 6,924 trainable parameters. We trained the InClass net using 1,000,000 datapoints for 15 epochs with a batch size of 500 (the other details of the training process remained the same as in \sref{subsec:bivariate_gaussian}), and estimated the mixture model. The estimated mixture weights, shown in the last column of \tref{tab:four_gaussians}, are in good agreement with the true weights of the components (second column in \tref{tab:four_gaussians}). The estimated distributions $f^{(i)}_x$, $f^{(i)}_y$, and $f^{(i)}_z$ of the variates $x$, $y$, and $z$, respectively, are shown in \fref{fig:four_gaussians_dists} as red solid curves, along with the true distributions depicted as green dash-dot curves. In all twelve cases (4 components $\times$ 3 variates) we observe good agreement between the true and estimated distribution.

 

\section{Identifiability of Conditional Independence Bivariate Mixture Models}
\label{sec:identifiability}
Identifiability of a statistical model is concerned with whether the parameters and functions that describe the model are uniquely identifiable from an infinite sample of datapoints produced from the model. The identifiability of mixture models is an important concept, especially in the context of using the techniques introduced in this paper for science applications. For the sake of precision, in this section we will distinguish between a statistical model and an instance of a statistical model as follows:
\begin{itemize}
 \item A \textbf{statistical model} captures the set of assumptions imposed on the form of the underlying distribution (e.g. exact functional forms, conditional independence, etc). It encompasses all the distributions which satisfy the assumptions, and the distributions are described by certain parameters and functions (in nonparameteric situations).
 \item An \textbf{instance} of a statistical model refers to a particular choice for the parameters and functions, and corresponds to one of the distributions within the model.
\end{itemize}
There are two related notions of identifiability in the literature. An instance of a statistical model is said to be identifiable if it is observationally distinguishable from every other instance of the same model, i.e., if no other instance leads to an equivalent\footnote{In this context, two distributions are considered equivalent if they are equal \emph{almost surely}.} probability distribution of the observed data. A statistical model is said to be identifiable if every instance of the model is observationally distingushable from every other instance.


For a given statistical model, the definition of what it means to estimate the model is usually chosen to be practically useful. In the context of estimating nonparametric CIMMs, the definition allows for the following ``leniencies'':
\begin{itemize}
 \item The number of components $C$ is assumed to be known \textit{a priori}. Otherwise, any CIMM will be unidentifiable since a given component $i$ can always be split into several new components which share the same distribution $f^{(i)}(\X)$ (with weights adding up to the weight of the ``parent'' component). Similarly, zero weight components can always be added without affecting the data distribution.
 \item The model only needs to be (and can only ever be) estimated up to permutations of the component indices.
 \item The distribution $f^{(i)}_v$ is considered to be the same as the distribution $g^{(i)}_v$ if $f^{(i)}_v(x_v) = g^{(i)}_v(x_v)$ \emph{almost surely}. In other words, $f^{(i)}_v$ and $g^{(i)}_v$ are allowed to be different over a set of probability measure $0$. This is required for the nonparametric case which allows arbitrary $f^{(i)}_v$-s.
\end{itemize}
Even with these leniencies, in general, nonparametric CIMMs are not identifiable. The results in the literature are usually concerned with the identifiability of \emph{specific instances of CIMMs}---they provide conditions under which instances of CIMMs are identifiable \cite{Hall2003,Hall2005,Elmore2005,Allman2009,Tahmasebi2018}.

The $(V=1, C\geq 2)$ case (univariate) is always unidentifiable nonparametrically. For the $(V\geq 3, C=2)$ case, \cite{Hall2003} provided certain regularity conditions under which instances of CIMMs are identifiable, and \cite{Allman2009} generalized the result to the $(V\geq 3, C\geq 2)$ case. The result from \cite{Allman2009} states that an instance of a CIMM with $V\geq 3$ and $C\geq 2$ is identifiable if the functions $\left\{f^{(1)}_v,\dots, f^{(C)}_v\right\}$ are linearly independent, for all $v=1,\dots,V$.

This leaves the bivariate case $(V=2)$, which is the main focus of this section. Ref.~\cite{Hall2003} showed that in the $(V=2, C=2)$ case, instances of nonparameteric CIMMs are not identifiable in general. In particular it was shown that for any instance of a two component bivariate nonparametric CIMM, there exists a two-parameter family of instances which leads to the same distribution of the observed variables $(x, y)$. The authors also noted that non-negativity conditions will introduce constraints on the allowed values for the two parameters. Extending this result from \cite{Hall2003}, we derive the following two theorems which provide a sufficient and a (different) necessary condition for instances of nonparametric CIMMs with $(V=2, C\geq 2)$ to be identifiable. The two conditions coincide for the $C=2$ case. We relegate the proof of the theorems to \aref{appendix:proof}.
\begin{theorem} \label{th:necessary}
\textbf{(Necessary condition)} A nonparametric conditional independence bivariate ($V=2$) mixture model with $C \geq 2$ components of the form given in \eqref{eq:bi_cifmm} is uniquely identifiable up to permutations of the component-identities only if the following necessary condition is satisfied:
\begin{equation}
\begin{split}
 \esssup\left[\frac{w_i f^{(i)}_t(t)}{w_i f^{(i)}_t(t) + w_j f^{(j)}_t(t)}\right] \equiv &\esssup\left[\frac{\alpha^{(i)}_t(t)}{\alpha^{(i)}_t(t) + \alpha^{(j)}_t(t)}\right] = 1\,,\\
 &\forall (i,j) \in \{1,\dots,C\}^2~:~i\neq j\,,~~\forall t\in \{x, y\}\,, \label{eq:necessary}
\end{split}
\end{equation}
where $\esssup[\texttt{func}(t)]$ represents the essential supremum of $\texttt{func}(t)$.
\hfill\openbox
\end{theorem}

\begin{theorem} \label{th:sufficient}
\textbf{(Sufficient condition)} A nonparametric conditional independence bivariate ($V=2$) mixture model with $C \geq 2$ components of the form given in \eqref{eq:bi_cifmm} is uniquely identifiable up to permutations of the component-identities if the following sufficient condition is satisfied:
\begin{equation}
 \esssup\left[\frac{w_i f^{(i)}_t(t)}{\P_{\!t}(t)}\right] \equiv \esssup\left[\alpha^{(i)}_t(t)\right] = 1\,,\qquad\qquad \forall i\in \{1,\dots,C\}\,, \forall t\in \{x, y\}\,, \label{eq:sufficient}
\end{equation}
where, as before, $\esssup[\texttt{func}(t)]$ represents the essential supremum of $\texttt{func}(t)$.
\hfill\openbox
\end{theorem}
The essential supremum can be thought of as an adaptation of the notion of supremum of a function, allowing for ignoring the behaviour of the function over regions with a total probability measure\footnote{It is understood that the relevant probability measure in \eqref{eq:necessary} and \eqref{eq:sufficient} is the one that corresponds to the mixture model itself.} of zero. These conditions can \emph{roughly} be interpreted as follows: The sufficient condition \eqref{eq:sufficient} will be satisfied if, for every component $i$ and variate $x$ or $y$, there exists some region in the phase space of the variate where component $i$ \emph{completely} dominates the mixture, i.e., all the datapoints in that region are from component $i$. The necessary condition \eqref{eq:necessary} will be satisfied if, for every pair of components $i\neq j$ and variate $x$ or $y$, there exists some region in the phase space of the variate where component $i$ \emph{completely} dominates the mixture of components $i$ and $j$.

Let us now revisit the examples considered earlier in \sref{sec:raindancesvi} from the point of view of identifiablilty. Considering the successful estimation of mixture models and/or classifier training in those examples, we can expect them to be identifiable. For the mixture of two independent bivariate Gaussians, \fref{fig:bivariate_gaussian_classifiers} shows how, for both $x$ and $y$, the true and reconstructed classifier output for the first (second) component approaches $1$ for increasingly negative (positive) values. This ensures that the sufficient condition for identifiability \eqref{eq:sufficient} is satisfied. Similarly, for the checkerboard mixture, by construction, there are regions in $x$ and $y$ which contain points from only component 1 or only component 2, see \fref{fig:checkerboard_comps}.

\begin{figure}[t]
 \centering
 \includegraphics[width=.47\textwidth]{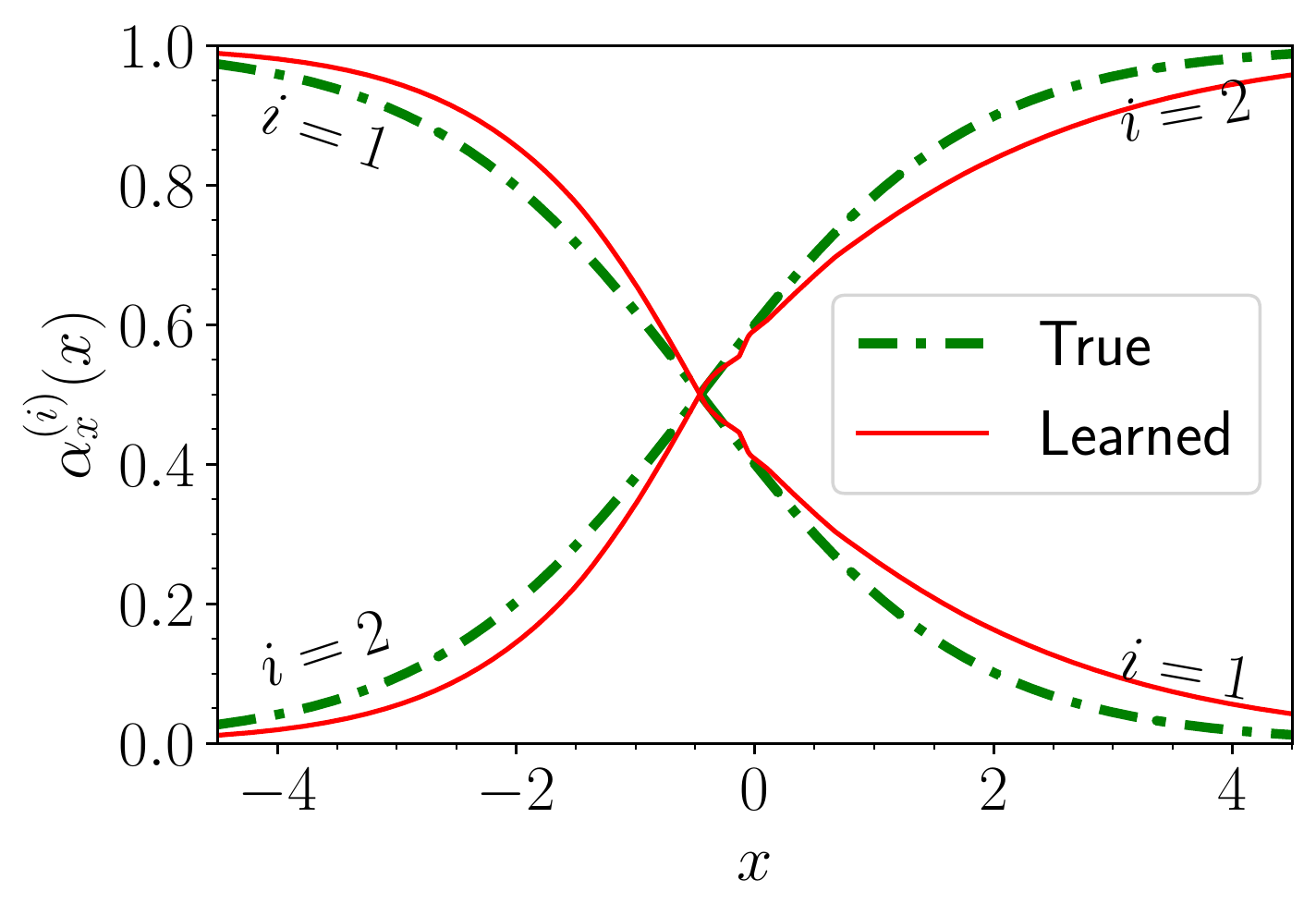}~~
 \includegraphics[width=.47\textwidth]{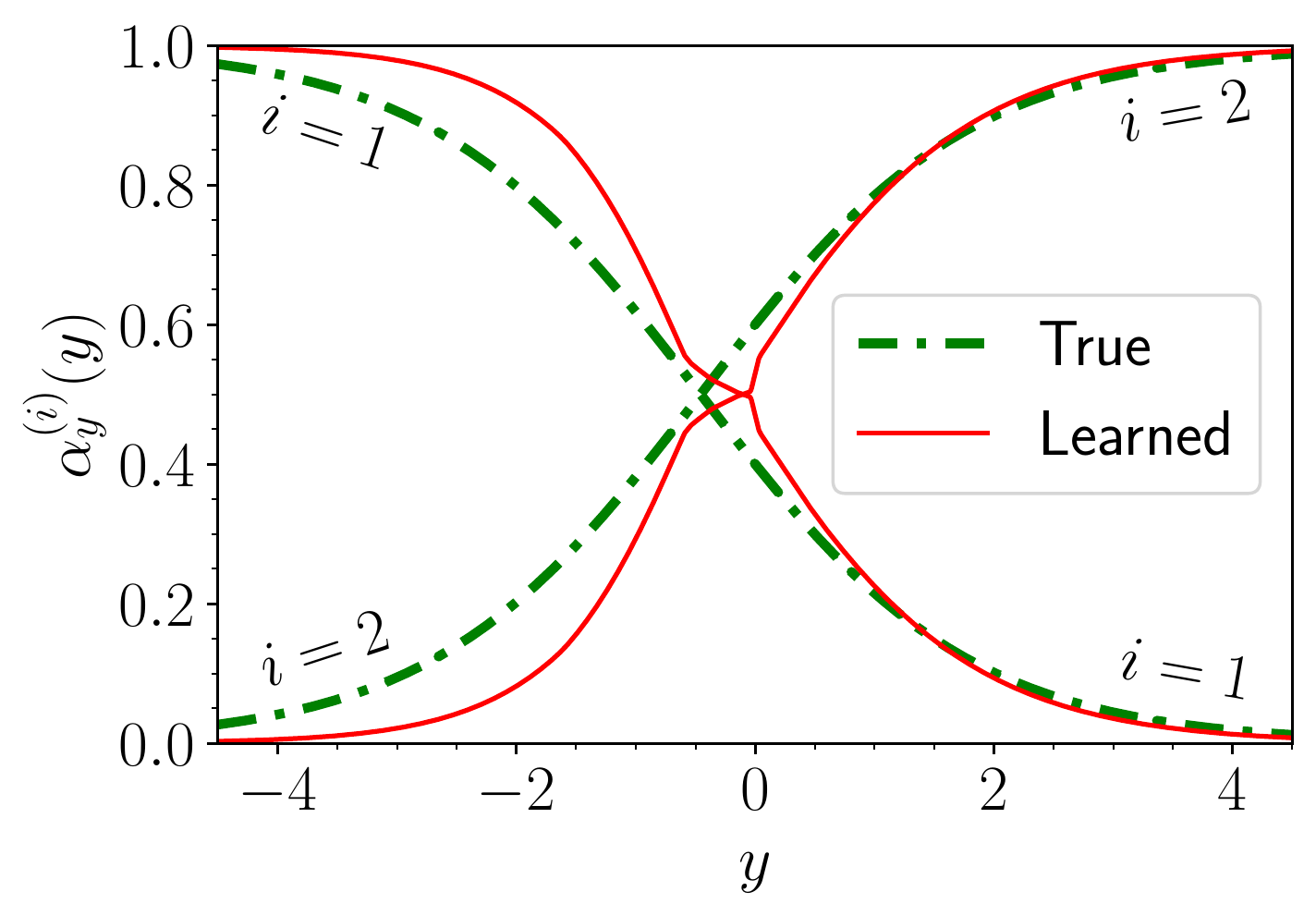}
 \caption{
 The same as \fref{fig:bivariate_gaussian_classifiers}, but using only 5,000 events for the estimation of the mixture model considered in \sref{subsec:bivariate_gaussian}.
 } \label{fig:bivariate_gaussian_unidentifiable_classifiers}
\end{figure}

\begin{figure}[t!]
 \centering
 \includegraphics[width=.47\textwidth]{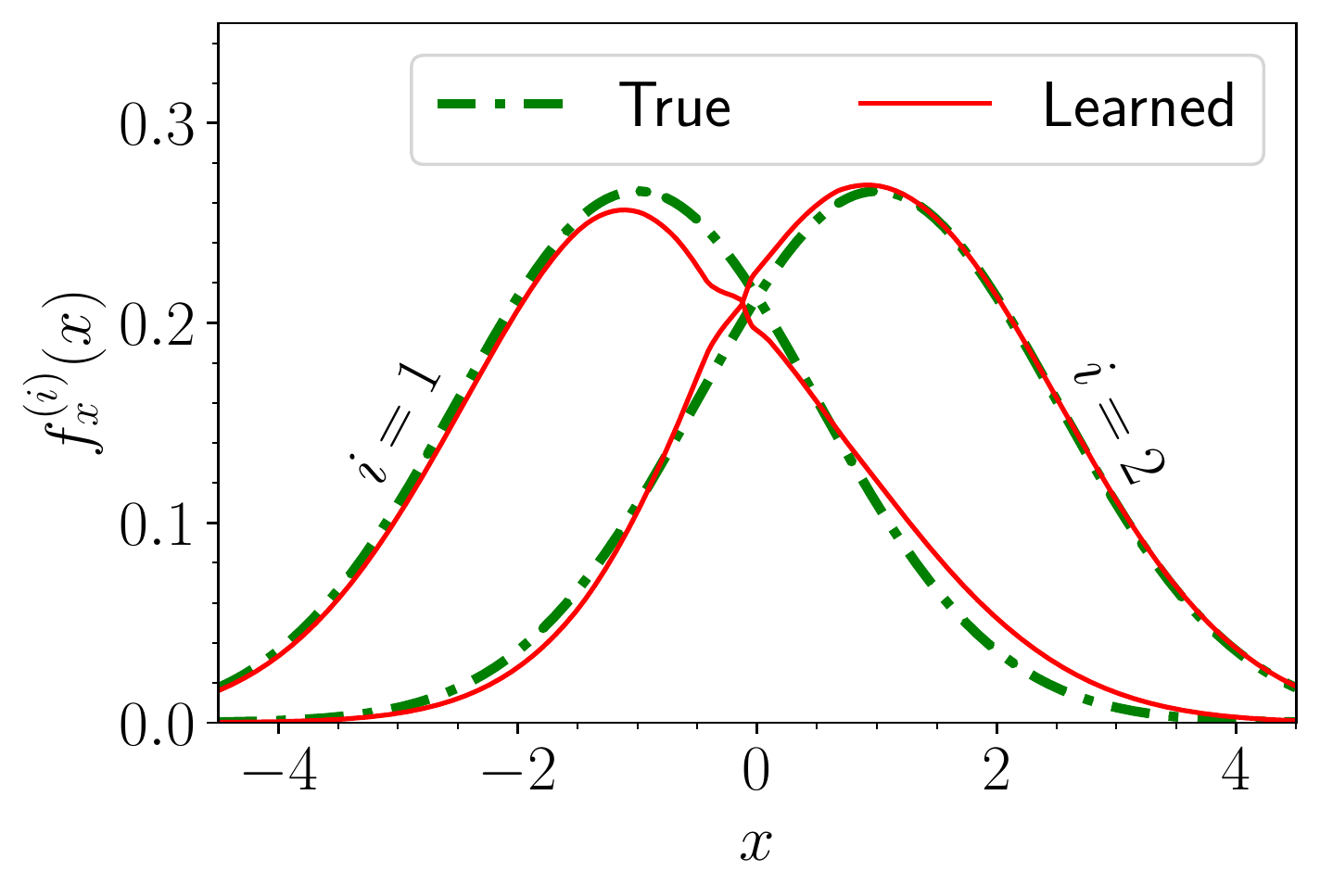}~~
 \includegraphics[width=.47\textwidth]{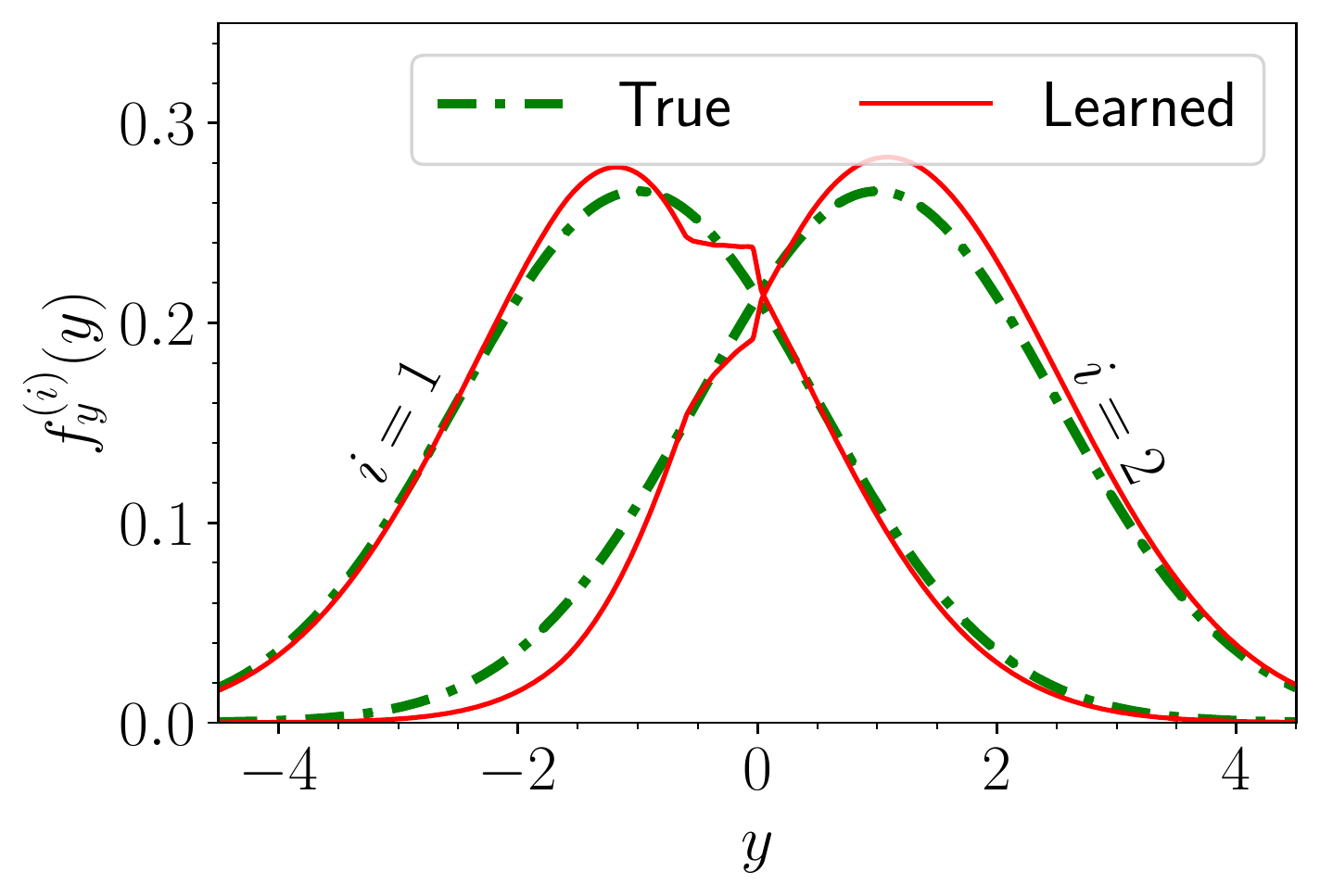}
 \caption{The same as \fref{fig:bivariate_gaussian_dists}, but using only 5,000 events for the estimation of the mixture model considered in \sref{subsec:bivariate_gaussian}.
 } \label{fig:bivariate_gaussian_unidentifiable_dists}
\end{figure}

Recall that in our treatment, the individual variates $x$ and $y$ are themselves allowed to be multi-dimensional. In the special case of one-dimensional variates $x$ and $y$, typically a component will only dominate the mixture in either the left tail or the right tail of the other components. This means that for most natural examples with one-dimensional $x$ and $y$, it is unlikely for the mixture to be identifiable for more than two components. On the other hand, this limitation does not apply to higher dimensional variates $x$ and $y$ which our InClass nets specialize in. For instance, the sufficient condition \eqref{eq:sufficient} for the mixture model constructed out of the MNIST dataset becomes: ``For every digit $d$, there must exist some region in the space of images, within which the images look unmistakably like the digit $d$''. This condition is naturally expected to be satisfied, considering the reliability of good handwritten communication.
\vskip 2mm
\noindent\textbf{Reduced identifiability due to limited statistics.}
The unique estimation of mixture model instances guaranteed by \thref{th:sufficient} can only be achieved with an infinite dataset. There will always be an uncertainty associated with estimation performed using finite datasets \cite{Matchev:2020jqz}. The level of this uncertainty is related to (among other things) how close the conditions \eqref{eq:necessary} and \eqref{eq:sufficient} are to being satisfied within the region of sample space covered sufficiently by the finite dataset at hand. In this sense, the result in \thref{th:sufficient} is useful from a practical point of view. To illustrate this, we repeated the two Gaussians example from \sref{subsec:bivariate_gaussian}, with the same setup, but with much fewer datapoints, namely 5,000 instead of 100,000. With fewer datapoints, the dataset is less likely to probe the tails of the $x$ and $y$ distributions, where a single component dominates. As expected, this time the estimation of the weights is slightly worse --- we obtained $w_1=0.44$ and $w_2=0.56$, to be compared with the true values of $w_1=0.4$ and $w_2=0.6$. The results for the classifiers $\alpha^{(i)}_{x}(x)$ and $\alpha^{(i)}_{y}(y)$ and for the component distributions $f^{(i)}_x$ and $f^{(i)}_y$ are shown in Figures~\ref{fig:bivariate_gaussian_unidentifiable_classifiers}
and \ref{fig:bivariate_gaussian_unidentifiable_dists}, respectively. Comparing to the analogous high statistics Figures~\ref{fig:bivariate_gaussian_classifiers} and \ref{fig:bivariate_gaussian_dists}, we see that the estimation has generally succeeded (after all, the model was identifiable), but is not perfect and suffers from statistical uncertainties.

\vskip 2mm
\noindent\textbf{Unidentifiable situations.}
When a CIMM instance is not identifiable, our technique will yield one of the parameterizations (weights and functions) that best fits the available data. Note that the unidentifiability of an instance of a nonparametric CIMM is not a weakness of our InClass nets approach, but rather a statement on the impossibility of the task of unique estimation.

\section{Discussion}
\label{sec:discussion}

In this section, we will discuss some considerations which might be relevant in the context of science applications of the InClass nets technique introduced in this paper.

\subsection{Uncertainty Quantification}
An important aspect of estimating a model (or equivalently, fitting a model to the available data), is providing an uncertainty on the estimate. Uncertainties in parametric estimation are conceptually straightforward---they correspond to the (possibly correlated) uncertainties in the estimated values of the parameters. The corresponding approach in the context of nonparameteric models (which allow arbitrary functions), would be to treat either the neural network outputs $\beta^{(i)}_v$ or the estimated distributions $f^{(i)}_v$ as Gaussian processes \cite{10.5555/1162254}. For a Gaussian process $G(\texttt{input})$, the value of $G$ at any finite set of $\texttt{input}$ values is taken to be randomly distributed according to a multivariate normal distribution. This allows us to assign uncertainty estimates on the value of $G$ at individual $\texttt{input}$ points, while also accounting for the correlations in the uncertainties between the values at different $\texttt{input}$ points. It has been shown that Gaussian processes can be modeled using Bayesian neural networks with wide layers \cite{neal_priors_1996,Lee2018DeepNN,g.2018gaussian}. Using wide Bayesian neural networks as the individual classifiers of the InClass net, one can obtain robust uncertainties on the estimated mixture model. In many scenarios, one is simply interested in visualizing a band of uncertainty around the estimated $\beta^{(i)}_v$-s, $\alpha^{(i)}_v$-s, or $f^{(i)}_v$-s and even narrow Bayesian neural networks may be sufficient for this purpose.

Note that the Gaussian process approach will work when the instance of the CIMM at hand is identifiable, and the uncertainty in the estimated model arises only from the finiteness of the dataset being analyzed. It is presently unclear whether Bayesian neural networks can capture the degree(s) of freedom in the model specification which are introduced by the unidentifiability of the CIMM instance.

\subsection{Estimating the Number of Components \texorpdfstring{$C$}{C}}
\label{subsec:C_estimation}
In many applications, one does not \textit{a priori} know the number of components in the CIMM \cite{Kasahara2014,mbakop}. In other situations, the assumption that the distribution of the data can be written as a CIMM may not necessarily be valid. In such situations, by training different InClass nets with different values of $C$, one may be able to a) verify the validity of the conditional independence assumption, and b) estimate $C$.

Note that increasing the number of components increases the fitting ability of a CIMM. More concretely, every CIMM instance with $C$ components can be thought of as a CIMM instance with $C'>C$ components (with $C'-C$ additional zero-weight components). As a result, an InClass net with more components should strictly perform better (in terms of the minimum cost value achieved), up to network training deficiencies and statistical fluctuations due to the finiteness of the training dataset. However, the improvement (in the minimum cost achieved) resulting from increasing $C$ is expected to diminish beyond a certain point.

In particular, if the true probability distribution of the data $\P^\ast(\X)$ can be modeled as a CIMM, then there exists a minimum number of components $C_\text{min}$ required to express $\P^\ast(\X)$ in the form
\begin{equation}
 \P^\ast(\X) = \sum_{i=1}^{C_\text{min}}~w_i~\prod_{v=1}^V~f^{(i)}_v(x_v)\,.
\end{equation}
Increasing $C$ from $1$ to $C_\text{min}$ will show an improvement in $\texttt{neg\_ctc\_cost}$ value, but beyond $C_\text{min}$, the performance is expected to saturate. This feature, if observed, can simultaneously a) confirm that the data is consistent with the conditional independence assumption, and b) provide an estimate of $C_\text{min}$. Note that the $C_\text{min}$ value identified in this way is only an estimate---inferring the presence of a component with a small mixing weight, or the presence of two components with very similar distributions $f^{(i)}(\X)$ may be statistically limited by the amount of data available. If the actual number of components is \textit{a priori} unknown, then the $C_\text{min}$ estimate can serve as an Occam's razor estimate of $C$.

On the other hand, if such a sharp saturation of network performance is not observed at a particular value of $C$, and the saturation is more gradual, this could be a sign of a Latent Factor Model---the underlying latent variable that explains the dependence of the different variates could be continuous instead of being the discrete category label $i$.

\subsection{Minimum Possible Value of \texorpdfstring{$\texttt{neg\_ctc\_cost}$}{neg\_ctc\_cost}}
When estimating $C_\text{min}$ using the method described in \sref{subsec:C_estimation}, one relies on observing a saturation in the value of the minimum cost achieved. However, such a saturation could also result from deficiencies in the architecture and/or training of the network. It it therefore useful to have an estimate of the minimum possible $\texttt{neg\_ctc\_cost}$ achievable by the best fitting model. Recall from \eqref{eq:neg_ctc_def}, that
\begin{equation}
 \texttt{neg\_ctc\_cost} = \mathrm{KL}\left[\P^\ast~\big|\big|~\P \right] - C^\ast(x_1,\dots,x_V)\,,
\end{equation}
where $\mathrm{KL}\left[\P^\ast~\big|\big|~\P \right]$is the KL divergence from the distribution represented by the InClass net $\P$ to the true distribution $\P^\ast$ and $C^\ast(x_1,\dots,x_V)$ is the total correlation of the variates under the true distribution. Since, the KL divergence is manifestly non-negative and equals 0 only when $\P^\ast$ is equivalent to $\P$, we have the following inequality
\begin{equation}
 \texttt{neg\_ctc\_cost} \geq -C^\ast(x_1,\dots,x_V)\,,
\end{equation}
where the equality is achieved when $\P^\ast$ matches $\P$ \emph{almost surely}. Thus, the negative total correlation $-C^\ast$ provides a (theoretically achievable) lower-bound on the negative cross total correlation $\texttt{neg\_ctc\_cost}$. From \eqref{eq:TC} the total correlation $C^\ast$ is given by
\begin{equation}
 C^\ast(x_1,\dots,x_V) = \int d\X~\P^\ast(\X)~\log{\left[\frac{\P^\ast(\X)}{\P^\ast_{\!1}(x_1) ~\P^\ast_{\!2}(x_2)~\dots~\P^\ast_{\!V}(x_V)}\right]}\,,
\end{equation}
where $\P_{\!v}^\ast$ represents the marginal distribution of $x_v$ in the data. For low dimensional data, $C^\ast$ can be estimated directly using this formula, after first estimating the distributions $\P^\ast(\X)$ and $\P_{\!v}^\ast(x_v)$.

Alternatively, for both low and high dimensional data, one can estimate $C^\ast$ using supervised machine learning as follows. Let the distribution $\Q^\ast(\X)$ be defined as
\begin{equation}
 \Q^\ast(\X) \equiv \prod_{v=1}^V~\P^\ast_{\!v}(x_v).
\end{equation}
Note that $C^\ast$ is simply the Kullback--Leibler divergence $\mathrm{KL}\left[\P^\ast~\big|\big|~\Q^\ast \right]$ from $\Q^\ast$ to $\P^\ast$. One can produce datapoints as per the distribution $\Q^\ast(\X)$ by independently sampling the variates $x_1,\dots,x_V$ from the available dataset. This gives us two datasets: the original one distributed as per $\P^\ast$ and a resampled one distributed as per $\Q^\ast$. One can train a machine in a supervised manner to distinguish between these two datasets and estimate the KL divergence, and hence $C^\ast$, from the trained classifier.

\subsection{Incorporating Prior Knowledge} \label{subsec:prior_knowledge}
In some applications, one may have additional prior knowledge about the mixture model, beyond the conditional independence assumption. It may be possible to incorporate this knowledge into the InClass net directly. For example, in the MNIST image classification example considered in \sref{sec:MNIST}, we used the information that the variates $x$ and $y$ are both images of digits to use the same classifier neural network for both variates.

As a different example, if the distribution of a given variate $x_v$ is known under a given component $i$, then the value of $\beta^{(i)}_v$ can be set to $f^{(i)}_v(x_v) / \P_{\!v}^\ast(x_v)$ up to a multiplicative weight factor which will constitute a single, trainable parameter. For the special case where the distribution of a given variate $x_v$ is known under every component, the classifier for the $v$-th variate can be parameterized by only the mixture weights of the components---in this way the InClass nets technique can be applied in the situations where the ${}_s\mathcal{P}lots$ technique is currently being used in high energy physics.

As another example, consider the case where the weights of the different components are \textit{a priori} known (but not the distributions of the variates within the components). Then an extra term can be added to the cost function to force the mixture weights $w_i$ estimated by the InClass net towards the true known weights $w_i^\text{true}$. One possible form of the extra term is inspired by the cross entropy:
\begin{equation}
  -\lambda \sum_{i=1}^C~w_i^\text{true}~\log\left(w_i\right)\,,
\end{equation}
where $\lambda$ is a parameter that controls the relative importance of the new term in the cost function. The additional term could be added either at the beginning, or after training the network for a few epochs (and identifying the map from the true component indices to the learned component indices). The additional term may be particularly useful in estimating unidentifiable CIMMs, where the additional knowledge of the mixture weights could help rule out the observationally indistinguishable ``fake'' CIMM instances.

\section{Possible Variations and Extensions}
\label{sec:extensions}

In this section we will discuss some potential variations and extensions of the InClass nets technique introduced in this paper. A detailed exploration of these ideas is beyond the scope of this work.

\subsection{Surrogate Cost Functions}
For the cost function $\texttt{neg\_ctc\_cost}$ from \eqref{eq:neg_ctc}, we can define a \emph{surrogate} cost function $\texttt{unnorm\_neg\_ctc\_cost}$ as
\begin{equation}
 \texttt{unnorm\_neg\_ctc\_cost} = - E_{\P^\ast}\left[~\log\left\{\displaystyle\sum_{i=1}^C~\left[\prod_{v=1}^V~\beta^{(i)}_v~\left(E_{\P^\ast}\left[\beta^{(i)}_v\right]\right)^{(1-V)/V}\right]\right\}~\right]\,. \label{eq:unnorm_neg_ctc}
\end{equation}
This surrogate cost function is an alternative cost function whose minimization will also lead to the minimization of the $\texttt{neg\_ctc\_cost}$. More concretely, if the true distribution $\P^\ast$ does correspond to a conditional independence mixture model, then the surrogate cost function will be minimized only when the network outputs (pseudo classifiers) $\beta^{(i)}_v$ match the classifiers $\alpha^{(i)}_v$ that correspond to a best fitting CIMM instance. This can be proved as follows: From \eqref{eq:neg_ctc} and \eqref{eq:unnorm_neg_ctc}, we have
\begin{subequations}
\begin{align}
 \texttt{unnorm\_neg\_ctc\_cost} &= \texttt{neg\_ctc\_cost} - E_{\P^\ast}\left[~\log\left\{\displaystyle\sum_{i=1}^C~\left[\prod_{v=1}^V\left(E_{\P^\ast}\left[\beta^{(i)}_v\right]\right)^{1/V}\right]\right\}~\right]\\
 &\geq \texttt{neg\_ctc\_cost} - E_{\P^\ast}\left[~\log\left\{\displaystyle\frac{1}{V}\sum_{i=1}^C~\sum_{v=1}^V\left(E_{\P^\ast}\left[\beta^{(i)}_v\right]\right)\right\}~\right] \label{eq:unnorm_proof_b}\\
 &= \texttt{neg\_ctc\_cost}\,. \label{eq:unnorm_proof_c}
\end{align}
\end{subequations}
In \eqref{eq:unnorm_proof_b}, we have used the inequality of arithmetic and geometric means and in step \eqref{eq:unnorm_proof_c}, we have used the constraint $\displaystyle\sum_{i=1}^C~\beta^{(i)}_v(x_v) = 1$ satisfied by the neural network outputs. Note that setting the pseudo classifiers $\beta^{(i)}_v$ to be equal to the classifiers $\alpha^{(i)}$ corresponding to a best fitting CIMM instance both a) minimizes $\texttt{neg\_ctc\_cost}$, and b) satisfies the condition for equality in \eqref{eq:unnorm_proof_b}. This completes the proof that the \texttt{unnorm\_neg\_ctc\_cost} is a suggogate cost function for the \texttt{neg\_ctc\_cost} when the data does correspond to some CIMM. The bivariate special case \texttt{unnorm\_neg\_cmi\_cost} which is a surrogate cost function for the \texttt{neg\_cmi\_cost} can be explictly written as
\begin{equation}
 \texttt{unnorm\_neg\_cmi\_cost} = - E_{\P^\ast}\left[~\log\left\{\displaystyle\sum_{i=1}^C~\frac{\beta^{(i)}_x\,\beta^{(i)}_y}{\sqrt{\displaystyle E_{\P^\ast}\left[\beta^{(i)}_x\right]\,E_{\P^\ast}\left[\beta^{(i)}_y\right]}}\right\}~\right]\,.
\end{equation}
These surrogate cost functions are also implemented in the \raindances{} package.

\subsection{Regularizers}
Recall that the dataset at hand could be consistent with multiple CIMM instances, either due to the unidentifiability of the instance or due to the finiteness of the dataset. This allows us to add additional properties for the learned model to satisfy. For example, depending on the application at hand, one might be interested in reducing the number of dominant (high mixture weight) components in the model identified by the network. This can be encouraged by adding (to the cost function) additional regularization terms like
\begin{subequations} \label{eq:reg}
\begin{align}
 \texttt{neg\_tikhonov\_reg} &= -\lambda~\sum_{i=1}^C w_i^2,\\
 \texttt{neg\_shannon\_reg} &= \lambda~\sum_{i=1}^C w_i \log w_i,
\end{align}
\end{subequations}
where $\lambda$ is a positive constant. If one is interested in more evenly weighted components, the same regularizer terms can be used with $\lambda$ set to a negative value.

\subsection{Unsupervised Classification With Multi-Label InClass Nets} 
\label{sec:multilabel}
The InClass nets architecture introduced in this paper can have more general data mining applications beyond the estimation of CIMMs. If the datapoints in a dataset are comprised of the (possibly multi-dimensional) variates $x_1,\dots,x_V$, the joint distribution of these variates may be understandable in terms of the classes of datapoints within the dataset, even if it does not fall under a CIMM. Furthermore, the set of classes corresponding to one variate need not necessarily be the same as the set of classes corresponding to another. In the literature, the existence of different sets of classes within the dataset falls under the realm of \emph{multi-label classification}.

For example, consider a dataset containing paired data: each datapoint contains the identities of a book and a movie liked by a person. The working assumption could be that there exists a classification of books and a (different) classification of movies, such that the class of books liked by a person is related to the class of movies liked by the same person. In such cases, it may be possible to simultaneously train a book and a movie classifier using the InClass nets architecture, by simply maximizing the mutual information between the classes predicted by the network.

To this end, we define the ``negative total correlation function'' $\texttt{neg\_tc\_cost}$ and its bivariate special case ``negative mutual information'' cost function $\texttt{neg\_mi\_cost}$ as
\begin{subequations}
\begin{align}
 \texttt{neg\_tc\_cost} &= -\sum_{i_1=1}^{C_1}~\sum_{i_2=1}^{C_2}~\cdots~\sum_{i_V=1}^{C_V}~E_{\P^\ast}\left[\prod_{v=1}^V~\alpha^{(i_v)}_v\right]~\log{\left[\frac{\displaystyle E_{\P^\ast}\left[\prod_{v=1}^V~\alpha^{(i_v)}_v\right]}{\displaystyle \prod_{v=1}^V~E_{\P^\ast}\left[\alpha^{(i_v)}_v\right]}\right]}, \label{eq:neg_tc}\\
 \texttt{neg\_mi\_cost} &= -\sum_{i=1}^{C_x}~\sum_{j=1}^{C_y}~E_{\P^\ast}\left[\alpha^{(i)}_x\,\alpha^{(j)}_y\right]~\log{\left[\frac{\displaystyle E_{\P^\ast}\left[\alpha^{(i)}_x\,\alpha^{(j)}_y\right]}{\displaystyle E_{\P^\ast}\left[\alpha^{(i)}_x\right]\,E_{\P^\ast}\left[\alpha^{(j)}_y\right]}\right]},
\end{align}
\end{subequations}
where the outputs of the InClass net $\eta^{(i)}_v$ are directly interpreted as the classifier output $\alpha^{(i)}_v$, and $C_v$ is the number of classes for the classifier corresponding to the $v$-th variate---note that the $C_v$-s need not all be equal. We point out that the \texttt{neg\_tc\_cost} of \eqref{eq:neg_tc} is a generalization of the cost function used in \cite{ji2019invariant} for the case where a) there can be more than two variates in the data, b) the classifiers $\alpha^{(i)}_v$ are not necessarily the same for different variates, and c) the number of classes $C_v$ could be different for different variates. Also, in this formulation, it is not required that the inputs $x_v$ to the different classifiers have only non-overlapping attributes of the datapoint.

\subsection{Semi-supervised classification with InClass nets}
\label{subsec:semisupervised}
In the MNIST image classification example considered in \sref{sec:MNIST}, we seeded the categories into the classifier network via supervised learning using a small, noisily labeled dataset. After the categories were seeded in, we used the unsupervised training of the InClass nets technique to further train the network.

This strategy has straightforward applications in semi-supervised learning scenarios where only a subset of the datapoints in the training dataset are labeled. For example, in the training of neural networks to perform medical diagnosis \cite{2005.07377}, generating labeled datasets requires manual annotation by experts, and only a small number of labeled samples may be available. On the other hand, a large number of unlabeled samples are typically available for training purposes. If, say, two different aspects (or variates) of the medical records are expected to only be weakly dependent on each other, but a confounding factor like the presence or absence of a disease can influence both variates, then we can train a neural network to perform the diagnosis leveraging both the labeled and unlabeled datasets. A hybrid cost function that incorporates a supervised classification cost function (for the labeled datapoints), as well an unsupervised cost function introduced in this paper (for the unlabeled datapoints) may be appropriate for the task.

Note that the medical diagnosis example considered here will not strictly be a conditional independence mixture model. For instance, in addition to the presence or absence of the disease, the severity of a particular case is also likely to influence the medical record. It may be possible to accommodate this particular effect by having multiple labels for different severity levels. Despite not strictly being an example of conditional indepedence mixture model, training using the $\texttt{neg\_ctc\_cost}$ or the $\texttt{neg\_tc\_cost}$ can still potentially yield useful diagnostic tools.


\section{Summary}

In this paper we introduced a novel approach for the \emph{nonparameteric} estimation of conditional independence mixture models defined by \eqref{eq:cifmm}. In this approach, the estimation of a CIMM is treated as a multi-class classification problem, which we solve with machine-learning methods. The main results of the paper are as follows.
\begin{itemize}
 \item We develop a specific machine-learning technique which we call the InClass nets technique. The basic architecture of InClass nets is illustrated in \fref{fig:inclass} and consists of a number of classifiers (one for each variate), which are realized as artificial neural networks. 
 \item In \sref{sec:inclass}, we show how CIMMs can be represented using InClass nets. The ability of neural networks to approximate arbitrary functions allows for the \emph{nonparametric} modeling of the CIMM.
 \item We recast the problem of estimation of a CIMM as a classification problem, and construct suitable cost functions for training the individual NNs without supervision. We also provide the prescription for extracting the learned CIMM from the trained InClass nets. The efficacy of our procedure is demonstrated with several toy examples in \sref{sec:raindancesvi}, including a high-dimensional image classification problem. 
 \item For easy adoption of the InClass nets technique, we provide a public implementation of our method as a Python package called \href{\raindancesurl}{\raindances{}} \cite{rd6}.
 \item In \sref{sec:identifiability} we derive some new results on the nonparametric identifiability of bivariate CIMMs, in the form of a necessary and a (different) sufficient condition for a bivariate CIMM to be identifiable. The proofs of the theorems can be found in \aref{appendix:proof}.
\end{itemize} 

As discussed in Sections \ref{sec:discussion} and \ref{sec:extensions}, the InClass nets technique has many potential applications beyond the narrow focus of CIMM. Specifically, the use of machine learning opens new avenues for addressing old-standing problems in nonparametric statistics.

\section*{Acknowledgements}
The authors would like to thank M.~Lisanti and A.~Roman for useful discussions. The work of PS was supported in part by the University of Florida CLAS Dissertation Fellowship (funded by the Charles Vincent and Heidi Cole McLaughlin Endowment) and the Institute of Fundamental Theory Fellowship. This work was supported in part by the United States Department of Energy under Grant No. DE-SC0010296.

\section*{Code and data availability}
The code and data that support the findings of this study are openly available at the following URL: \url{https://gitlab.com/prasanthcakewalk/code-and-data-availability/}.

\appendix

\section{Proof of Theorems \ref{th:necessary} and \ref{th:sufficient}} \label{appendix:proof}
Here we will prove Theorems \ref{th:necessary} and \ref{th:sufficient}. Let us begin by noting that:
\begin{itemize}
 \item A nonparametric CIMM can be identifiable only if all the mixture weights are non-zero---if one of the mixture components has zero weight, it can be removed from the mixture and different component can be split into two.
 \item A nonparametric CIMM with $V=2$ cannot be identifiable if there exists a pair of components $i,j$ for which $f^{(i)}_x(x) = f^{(j)}_x(x)$ almost surely. Otherwise, the sub-mixture of the components $i$ and $j$, $w_i\,f^{(i)}_x(x)\,f^{(i)}_y(y) + w_j\,f^{(j)}_x(x)\,f^{(j)}_y(y)$, can be rewritten as a different combination of two components of total weight $w_i + w_j$ which have the same distribution of the variate $x$ as the original components, but different mixture weights and distributions of the variate $y$.
 \item Similarly, a nonparametric CIMM with $V=2$ cannot be identifiable if there exists a pair of components $i,j$ for which $f^{(i)}_y(y) = f^{(j)}_y(y)$ almost surely.
\end{itemize} 
Congruently, neither the necessary nor the sufficient condition from Theorems \ref{th:necessary} and \ref{th:sufficient} can be satisfied if one of the mixing weights is zero, or if $\exists (i, j, t): f^{(i)}_t(t) = f^{(j)}_t(t)$ \emph{almost surely}. Henceforth, we will only consider instances of nonparametric bivariate CIMMs for which
\begin{subequations}
\begin{align}
 w_i > 0\,,\qquad &\forall i\in\{1,\dots,C\}\,,\\
 \Big(f^{(i)}_x - f^{(j)}_x = 0 \text{ \emph{almost surely}} \Big)~~ &\Rightarrow~~\big(i=j\big)\,,\\
 \Big(f^{(i)}_y - f^{(j)}_y = 0 \text{ \emph{almost surely}} \Big)~~ &\Rightarrow~~\big(i=j\big)\,.
\end{align}
\end{subequations}

\subsection{Two component case \texorpdfstring{$(C=2)$}{(C=2)}}
Let us first tackle the $C=2$ case of Theorems \ref{th:necessary} and \ref{th:sufficient}. Throughout this section, equality of distributions will refer to their equality \emph{almost surely}. From Theorem 4.1 and 4.2 of \cite{Hall2003}, for every instance of parametric bivariate CIMM of the form given in \eqref{eq:bi_cifmm}, all the instances with the same distribution of observed data form a two-parameter family. This family of instances identified in \cite{Hall2003} can be parameterized in terms of $\gamma\in\mathbb{R}$ and $0\leq w'_1\leq 1$, and can be written as
\begin{equation} 
 \sum_{i=1}^2~w'_i\,g^{(i)}_x(x)\,g^{(i)}_y(y) = \sum_{i=1}^2~w_i\,f^{(i)}_x(x)\,f^{(i)}_y(y)\,,
\end{equation}
where

\begin{subequations} \label{eq:twoparam}
\begin{alignat}{2}
 w'_2 &= 1 - w'_1\,,\\
 g^{(1)}_x(x) &= \P_x(x) + \gamma w'_2 &&\sqrt{\frac{w_1\,w_2}{w'_1\,w'_2}}~\left(f^{(1)}_x(x) - f^{(2)}_x(x)\right)\,,\\
 g^{(2)}_x(x) &= \P_x(x) - \gamma w'_1 &&\sqrt{\frac{w_1\,w_2}{w'_1\,w'_2}}~\left(f^{(1)}_x(x) - f^{(2)}_x(x)\right)\,,\\
 g^{(1)}_y(y) &= \P_y(y) + ~\frac{w'_2}{\gamma} &&\sqrt{\frac{w_1\,w_2}{w'_1\,w'_2}}~\left(f^{(1)}_y(y) - f^{(2)}_y(y)\right)\,,\\
 g^{(2)}_y(y) &= \P_y(y) - ~\frac{w'_1}{\gamma} &&\sqrt{\frac{w_1\,w_2}{w'_1\,w'_2}}~\left(f^{(1)}_y(y) - f^{(2)}_y(y)\right)\,.
\end{alignat}
\end{subequations}
Note, that the transformation $\gamma \longleftrightarrow -\gamma, w'_1 \longleftrightarrow 1-w'_1$ is equivalent to a permutation of the component indices $(1) \longleftrightarrow (2)$. Since, we are only interested in the identifiability of the CIMM instance upto this permutation, we can restrict $\gamma$ to be non-negative. The only additional constraints on $\gamma$ and $w'_1$ are provided by the non-negativity of the distribution functions $g^{(i)}_x$ and $g^{(i)}_y$.

It can be verified that $\gamma = 1$ and $w'_1 = w_1$ corresponds to the original CIMM instance with $g^{(i)}_x = f^{(i)}_x$ and $g^{(i)}_y = f^{(i)}_y$. Furthermore, any other set of values for $\gamma$ and $w'_1$ corresponds to a different instance, since the $w_1, w_2 > 0$ and the differences $f^{(1)}_x - f^{(2)}_x$ and $f^{(1)}_y - f^{(2)}_y$ are not identically zero. This leads us to the following lemma: The CIMM instance will be identifiable if and only if the non-negativity constraints on $g^{(i)}_x$ and $g^{(i)}_y$ only allow $\gamma$ and $w'_1$ to be $1$ and $w_1$, respectively.

The non-negativity conditions on the functions $g^{(i)}_x$ and $g^{(i)}_y$ can be written using \eqref{eq:twoparam} as
\begin{subequations} \label{eq:muintro}
\begin{alignat}{2}
 \frac{1}{\gamma}\sqrt{\frac{w'_1}{w'_2}} &\geq \sqrt{w_1\,w_2}~~\esssup\left[\frac{f^{(2)}_x(x) - f^{(1)}_x(x)}{\P_{\!x}(x)}\right] &&= \frac{\mu^{(2)}_x - w_2}{\sqrt{w_1\,w_2}}\,, \label{eq:muintro_a}\\
 \frac{1}{\gamma}\sqrt{\frac{w'_2}{w'_1}} &\geq \sqrt{w_1\,w_2}~~\esssup\left[\frac{f^{(1)}_x(x) - f^{(2)}_x(x)}{\P_{\!x}(x)}\right] &&= \frac{\mu^{(1)}_x - w_1}{\sqrt{w_1\,w_2}}\,, \label{eq:muintro_b}\\
 \gamma\sqrt{\frac{w'_1}{w'_2}} &\geq \sqrt{w_1\,w_2}~~\esssup\left[\frac{f^{(2)}_y(y) - f^{(1)}_y(y)}{\P_{\!y}(y)}\right] &&= \frac{\mu^{(2)}_y - w_2}{\sqrt{w_1\,w_2}}\,, \label{eq:muintro_c}\\
 \gamma\sqrt{\frac{w'_2}{w'_1}} &\geq \sqrt{w_1\,w_2}~~\esssup\left[\frac{f^{(1)}_y(y) - f^{(2)}_y(y)}{\P_{\!y}(y)}\right] &&= \frac{\mu^{(1)}_y - w_1}{\sqrt{w_1\,w_2}}\,, \label{eq:muintro_d}
\end{alignat}
\end{subequations}
where
\begin{equation}
 \mu^{(i)}_t = \esssup\left[\frac{w_i\,f^{(i)}_t(t)}{w_1\,f^{(1)}_t(t)+w_2\,f^{(2)}_t(t)}\right]\,,\qquad\qquad \forall i\in\{1,2\}\,,\forall t\in\{x,y\}\,.
\end{equation}
It can seen from \eqref{eq:muintro} that the $\mu^{(i)}_t$-s satisfy the constraints $w_i \leq \mu^{(i)}_t \leq 1$, since the essential supremum of the difference between two normalized distributions is non-zero---normalized equations have to cross or be equal almost surely. Now, multiplying \eqref{eq:muintro_a} with \eqref{eq:muintro_c}, and \eqref{eq:muintro_b} with \eqref{eq:muintro_d} we get the following constraint in the ratio $w'_1/w'_2$
\begin{equation}
 \frac{\left(\mu^{(2)}_x - w_2\right)\left(\mu^{(2)}_y - w_2\right)}{w_1\,w_2} \leq \frac{w'_1}{w'_2} \leq \frac{w_1\,w_2}{\left(\mu^{(1)}_x - w_1\right)\left(\mu^{(1)}_y - w_1\right)}\,. \label{eq:ratio_constraint}
\end{equation}
Note that all values $w'_1/w'_2$ allowed by this constraint are allowed by \eqref{eq:muintro} (e.g., by setting $\gamma=1$). This implies that the CIMM instance will be identifiable if and only if the \eqref{eq:ratio_constraint} only allows $w'_1 = w_1$. The upper and lower bounds on the ratio $w'_1/w'_2$ from \eqref{eq:ratio_constraint} both equal $w_1/w_2$ iff $\mu^{(1)}_x = \mu^{(2)}_x = \mu^{(1)}_y = \mu^{(2)}_y = 1$. This completes the proof of Theorems \ref{th:necessary} and \ref{th:sufficient} for the two component case.
\subsection{Necessary condition for the \texorpdfstring{$C > 2$}{C > 2} case}
The necessary condition from \thref{th:necessary} for the $C > 2$ case can be seen as a corollary of the same \thref{th:necessary} for the $C = 2$ case, since a nonparametric CIMM instance with more than two components can be identifiable only if for every pair of components, the two component mixture formed by the pair (after appropriately scaling their weights to add up to 1) is identifiable.
\subsection{Sufficient condition for the \texorpdfstring{$C > 2$}{C > 2} case}
Let $\Omega_x$ and $\Omega_y$ be the sample spaces of $x$ and $y$, respectively, and let $\mathbb{P}[\,\cdots]$ represent the probability of an event. Let us consider a bivariate CIMM instance with $C>2$ components which satisfies condition \eqref{eq:sufficient}, i.e., the sufficient condition for identifiability according to \thref{th:sufficient} (which is to be proved here)\footnote{The following proof also works for the $C=2$ case.}. Let $w_i$, $f^{(i)}_x$, $f^{(i)}_y$, $\alpha^{(i)}_x$, and $\alpha^{(i)}_y$ have the same meanings as in the rest of the paper.

From the definition of $\esssup$, we can see that for all $0 < \epsilon < 1$, there exist disjoint sets $X_1, \dots, X_C \subset \Omega_x$ and disjoint sets $Y_1, \dots, Y_C \subset \Omega_y$ such that\footnote{For notational convenience, the $\epsilon$-dependence of the sets $X_i$ and $Y_i$ is not indicated explicitly.}
\begin{subequations} \label{eq:suff_cond_alternative}
\begin{alignat}{3}
 \mathbb{P}[x\in X_i] &> 0\,, &&\qquad &&\forall i\in\{1,\dots,C\}\,, \\
 \mathbb{P}[y\in Y_i] &> 0\,, &&\qquad &&\forall i\in\{1,\dots,C\}\,, \\
 (1-\epsilon) \leq \alpha^{(i)}_x(x) &\leq 1\,, &&\qquad \forall x\in X_i\,, &&\forall i\in\{1,\dots,C\}\,, \label{eq:suff_cond_alternative_c}\\
 (1-\epsilon) \leq \alpha^{(i)}_y(y) &\leq 1\,, &&\qquad \forall y\in Y_i\,, &&\forall i\in\{1,\dots,C\}\,. \label{eq:suff_cond_alternative_d}
\end{alignat}
\end{subequations}
From \eqref{eq:suff_cond_alternative_c} and \eqref{eq:suff_cond_alternative_d}, we can see that
\begin{subequations} \label{eq:ulim_noneq}
\begin{alignat}{3}
 \alpha^{(i)}_x(x) &\leq \epsilon\,, &&\qquad \forall x\in X_j\,, &&~~\forall (i,j) \in \{1,\dots,C\}^2~:~i\neq j\,, \label{eq:ulim_noneq_x}\\
 \alpha^{(i)}_y(y) &\leq \epsilon\,, &&\qquad \forall y\in Y_j\,, &&~~\forall (i,j) \in \{1,\dots,C\}^2~:~i\neq j\,. \label{eq:ulim_noneq_y}
\end{alignat}
\end{subequations}
As $\epsilon$ is made arbitrarily small, the region $x\in X_i$ and the region $y\in Y_i$ become arbitrarily close to being populated exclusively by the component $i$. This induces a block diagonal structure, with the probability $\mathbb{P}_{ij}\equiv\mathbb{P}\left[(x,y)\in X_i\times Y_j\right]$ becoming arbitrarily small if $i\neq j$. More concretely, from \eqref{eq:IC_pdf}, we can write
\begin{equation} \label{eq:prob_xy}
 \mathbb{P}_{ij} = \mathbb{P}[x\in X_i]~\mathbb{P}[y\in Y_j]~\sum_{k=1}^C~\frac{E_{x\in X_i}\left[\alpha^{(k)}_x(x)\right]~E_{y\in Y_j}\left[\alpha^{(k)}_y(y)\right]}{w_k}\,.
\end{equation}
Using \eqref{eq:suff_cond_alternative_c}, \eqref{eq:suff_cond_alternative_d}, \eqref{eq:ulim_noneq}, and \eqref{eq:prob_xy}, we can show that
\begin{equation} \label{eq:off_diagonal}
 \mathbb{P}_{ij} \leq \epsilon~~\mathbb{P}[x\in X_i]~\mathbb{P}[y\in Y_j]~\sum_{k=1}^C~w_k^{-1}\,,\qquad \forall i\neq j\,.
\end{equation}
Similarly, using \eqref{eq:suff_cond_alternative_c}, \eqref{eq:suff_cond_alternative_d}, and \eqref{eq:prob_xy} we can show that
\begin{equation} \label{eq:on_diagonal}
 \mathbb{P}_{ii} \geq (1-\epsilon)^2~~\frac{\mathbb{P}[x\in X_i]~\mathbb{P}[y\in Y_i]}{w_i}\,, \qquad \forall i\,.
\end{equation}
Now, let us consider a different CIMM instance with weights $w'_i$, distributions $f'^{\,(i)}_x$ and $f'^{\,(i)}_y$, and classifiers $\alpha'^{\,(i)}_x$ and $\alpha'^{\,(i)}_y$ which has an observationally equivalent distribution $\P(x,y)$ as the original CIMM instance. We will refer to this as the ``primed CIMM instance''. We will prove that the original CIMM is identifiable by showing that the primed CIMM instance must be equivalent to the original, up to permutations of the component index $i$.

The key observation is that in the small $\epsilon$ limit, no component of primed CIMM instance can have non-vanishing contributions in the region $(x,y)\in X_i\times Y_i$ for more than one $i$. If some component of the primed instance, say the $k$-th component, has non-vanishing contributions to $\mathbb{P}_{ii}$ and $\mathbb{P}_{jj}$ for $i\neq j$, then the ``off-diagonal probabilities'' $\mathbb{P}_{ij}$ and $\mathbb{P}_{ji}$ will also receive non-vanishing contributions (due to conditional independence), which is not allowed by \eqref{eq:off_diagonal}. To make this argument more carefully, it can be shown, from \eqref{eq:prob_xy}, that for all $i,j,k\in\{1,\dots,C\},$
\begin{equation}\label{eq:primed}
\begin{split}
 \mathbb{P}_{ij}\,\mathbb{P}_{ji} &\geq \mathbb{P}[x\in X_i]~\mathbb{P}[y\in Y_i]~\mathbb{P}[x\in X_j]~\mathbb{P}[y\in Y_j]\\
 &\qquad\quad \times \frac{E_{x\in X_i}\left[\alpha'^{\,(k)}_x(x)\right]\,E_{y\in Y_i}\left[\alpha'^{\,(k)}_y(y)\right]\,E_{x\in X_j}\left[\alpha'^{\,(k)}_x(x)\right]\,E_{y\in Y_j}\left[\alpha'^{\,(k)}_y(y)\right]}{w'^{\,2}_k}\,.
\end{split}
\end{equation}
Using \eqref{eq:off_diagonal} and \eqref{eq:primed}, we can show that for all $i,j,k\in\{1,\dots,C\}$ with $i\neq j$,
\begin{equation}
 \frac{E_{x\in X_i}\left[\alpha'^{\,(k)}_x(x)\right]\,E_{y\in Y_i}\left[\alpha'^{\,(k)}_y(y)\right]}{w'_k}~\frac{E_{x\in X_j}\left[\alpha'^{\,(k)}_x(x)\right]\,E_{y\in Y_j}\left[\alpha'^{\,(k)}_y(y)\right]}{w'_k} \leq \left[\epsilon \sum_{k=1}^C~w_k^{-1}\right]^2\,. \label{eq:primed_ulim}
\end{equation}
This equation captures the constraint that no component $k$ of the primed CIMM instance can have non-vanishing contributions in two different regions $(x,y)\in X_i\times Y_i$ and $(x,y)\in X_j\times Y_j$ with $i\neq j$. On the other hand, each of the ``diagonal regions'' must receive a non-vanishing contribution from at least of the components of the primed instance. More concretely, from \eqref{eq:prob_xy} and \eqref{eq:on_diagonal}, one can see that for all $i\in \{1,\dots,C\}$, there exists a $k\in \{1,\dots,C\}$ such that
\begin{equation}
 \frac{E_{x\in X_i}\left[\alpha'^{\,(k)}_x(x)\right]\,E_{y\in Y_i}\left[\alpha'^{\,(k)}_y(y)\right]}{w'_k} \geq \frac{(1-\epsilon)^2}{C\,w_i}\,. \label{eq:primed_llim}
\end{equation}
From \eqref{eq:primed_ulim} and \eqref{eq:primed_llim} and the fact that both the original and the primed CIMM instances have the same number of components, one can see that as $\epsilon$ is made arbitrarily small, there exists a permutation $\sigma$ of the component indices such that $f^{(i)}_x$-s are observationally equivalent to the corresponding $f'^{\,\sigma(i)}_x$-s in the region $x\in \bigcup\limits_{i=1}^C\, X_i$, and similarly $f^{(i)}_y$-s are observationally equivalent to $f'^{\,\sigma(i)}_y$-s in the region $y\in \bigcup\limits_{i=1}^C\, Y_i$.

The equality of the weights $w_i$ and $w'_{\sigma(i)}$ and the equivalence of the distributions $f^{(i)}_x$ and $f'^{\,\sigma(i)}_x$ in the entire sample space $\Omega_x$ follows from the fact that both the original and primed CIMM instances have observationally equivalent distribution $\P(x,y)$ in the region $(x,y)\in \Omega_x\times Y_i$---note that $Y_i$ has a non-zero measure for all $\epsilon > 0$. A symmetric argument establishes the equivalence of the distributions $f^{(i)}_y$ and $f'^{\,\sigma(i)}_y$ in the entire sample space $\Omega_y$. This completes the proof of \thref{th:sufficient} for $C\geq 2$ components.





\section{Functional Gradient of \texttt{neg\_ctc\_cost}} \label{appendix:gradient}
In this section we will discuss a strategy that can speed-up and improve the training of InClass nets using the $\texttt{neg\_ctc\_cost}$ of \eqref{eq:neg_ctc}
\begin{equation}
 \texttt{neg\_ctc\_cost} = - E_{\P^\ast}\left[~\log\left\{\frac{\displaystyle\sum_{i=1}^C~\left[\prod_{v=1}^V~\beta^{(i)}_v~\left(E_{\P^\ast}\left[\beta^{(i)}_v\right]\right)^{\frac{1-V}{V}}\right]}{\displaystyle\sum_{i=1}^C~\left[\prod_{v=1}^V\left(E_{\P^\ast}\left[\beta^{(i)}_v\right]\right)^{1/V}\right]}\right\}~\right]\,.
\end{equation}
Note that there are multiple expectations $E_{\P^\ast}$ in the expression for the cost function. The outermost expectation is similar to the one in a cost function which can be written as an expectation over a per-datapoint loss function. For such cost functions (which only have an outermost expectation), one can use stochastic or mini-batch gradient descent for faster or more efficient training of the network. However, the presence of the inner expectations $\varphi^{(i)}_v \equiv E_{\P^\ast}\left[\beta^{(i)}_v\right]$ in our cost function means that the batch size used in the training should be large enough to estimate the pseudo weights $\varphi^{(i)}_v$ well. In particular, the batch size should be large enough to pick up subtle changes in the value of $\varphi^{(i)}_v$ caused by changes to the network weights $\vec{\theta}$. The need for large batch sizes will only be exacerbated as the number of components increases.

However, we can overcome this difficulty, and facilitate the use of stochastic gradient descent to optimize the $\texttt{neg\_ctc\_cost}$ as shown below. We will begin by deriving the expression for the functional derivative of the cost function with respect to the neural network outputs. For convenience, let is define $N$ and $D$ as
\begin{subequations}
\begin{align}
 N &\equiv \sum_{i=1}^C~\left[\prod_{v=1}^V~\beta^{(i)}_v~\left(E_{\P^\ast}\left[\beta^{(i)}_v\right]\right)^{\frac{1-V}{V}}\right]\,,\\
 D &\equiv \sum_{i=1}^C~\left[\prod_{v=1}^V\left(E_{\P^\ast}\left[\beta^{(i)}_v\right]\right)^{1/V}\right]\,.
\end{align}
\end{subequations}
This lets us write
\begin{equation}
 \texttt{neg\_ctc\_cost} = -E_{\P^\ast}\left[\log\left(\frac{N}{D}\right)\right]\,.
\end{equation}
Taking the functional derivative with respect to $\beta^{(j)}_u(x'_u)$, one gets
\begin{align}
 &\frac{\delta\,\texttt{neg\_ctc\_cost}}{\delta\,\beta^{(j)}_u(x'_u)} = - E_{\P^\ast}\left[~\frac{1}{N}\frac{\delta\,N}{\delta\,\beta^{(j)}_u(x'_u)}~\right] + \frac{1}{D}\frac{\delta\,D}{\delta\,\beta^{(j)}_u(x'_u)}\\
\begin{split}
 \qquad\qquad&= -~\frac{\P^\ast_u(x'_u)}{\beta^{(j)}_u(x'_u)}~E_{\P^\ast}\left[\left.~\frac{~\displaystyle\prod_{v=1}^V~\beta^{(j)}_v(x_v)~\left(E_{\P^\ast}\left[\beta^{(j)}_v\right]\right)^{(1-V)/V}~}{~\displaystyle\sum_{i=1}^C~\left[\prod_{v=1}^V~\beta^{(i)}_v(x_v)~\left(E_{\P^\ast}\left[\beta^{(i)}_v\right]\right)^{(1-V)/V}\right]~} ~\right|~ x_u = x'_u ~\right]\\
 &\qquad\qquad -~\frac{\P^\ast_u(x'_u)}{E_{\P^\ast}\left[\beta^{(j)}_u\right]}~E_{\P^\ast}\left[~\frac{~\displaystyle\prod_{v=1}^V~\beta^{(j)}_v~\left(E_{\P^\ast}\left[\beta^{(j)}_v\right]\right)^{(1-V)/V}~}{~\displaystyle\sum_{i=1}^C~\left[\prod_{v=1}^V~\beta^{(i)}_v~\left(E_{\P^\ast}\left[\beta^{(i)}_v\right]\right)^{(1-V)/V}\right]~}~\right]\\
 &\qquad\qquad +~\frac{\P^\ast_u(x'_u)}{E_{\P^\ast}\left[\beta^{(j)}_u\right]}~\frac{~\displaystyle\prod_{v=1}^V\left(E_{\P^\ast}\left[\beta^{(j)}_v\right]\right)^{1/V}~}{~\displaystyle\sum_{i=1}^C~\left[\prod_{v=1}^V\left(E_{\P^\ast}\left[\beta^{(i)}_v\right]\right)^{1/V}\right]~}\,.
\end{split}
\end{align}
Using this, we can write the gradient of the cost function with respect to the neural network weights $\vec{\theta}$ as
\begin{align}
 \nabla_{\mathbf{\theta}}\,&\texttt{neg\_ctc\_cost} = \sum_{j=1}^C~\sum_{u=1}^V~\int d x'_u~ \frac{\delta\,\texttt{neg\_ctc\_cost}}{\delta\,\beta^{(j)}_u(x'_u)}~\nabla_{\mathbf{\theta}}\,\beta^{(j)}_u(x'_u)\\
\begin{split}
 &= -\sum_{j,u}~E_{\P^\ast}\left[~\frac{\nabla_{\mathbf{\theta}}\,\beta^{(j)}_u}{\beta^{(j)}_u}~\frac{~\displaystyle\prod_{v=1}^V~\beta^{(j)}_v~\left(E_{\P^\ast}\left[\beta^{(j)}_v\right]\right)^{(1-V)/V}~}{~\displaystyle\sum_{i=1}^C~\left[\prod_{v=1}^V~\beta^{(i)}_v~\left(E_{\P^\ast}\left[\beta^{(i)}_v\right]\right)^{(1-V)/V}\right]~}~\right]\\
 &\qquad\quad -\sum_{j,u}~\frac{E_{\P^\ast}\left[\nabla_{\mathbf{\theta}}\,\beta^{(j)}_u\right]}{E_{\P^\ast}\left[\beta^{(j)}_u\right]}~E_{\P^\ast}\left[~\frac{~\displaystyle\prod_{v=1}^V~\beta^{(j)}_v~\left(E_{\P^\ast}\left[\beta^{(j)}_v\right]\right)^{(1-V)/V}~}{~\displaystyle\sum_{i=1}^C~\left[\prod_{v=1}^V~\beta^{(i)}_v~\left(E_{\P^\ast}\left[\beta^{(i)}_v\right]\right)^{(1-V)/V}\right]~}~\right]\\
 &\qquad\quad +\sum_{j,u}~\frac{E_{\P^\ast}\left[\nabla_{\mathbf{\theta}}\,\beta^{(j)}_u\right]}{E_{\P^\ast}\left[\beta^{(j)}_u\right]}~\frac{~\displaystyle\prod_{v=1}^V\left(E_{\P^\ast}\left[\beta^{(j)}_v\right]\right)^{1/V}~}{~\displaystyle\sum_{i=1}^C~\left[\prod_{v=1}^V\left(E_{\P^\ast}\left[\beta^{(i)}_v\right]\right)^{1/V}\right]~}\,.
\end{split}
\end{align}
This expression allows us to approximate the gradient of the cost function as
\begin{align}
 \begin{split}
 \nabla_{\mathbf{\theta}}\,\texttt{neg\_ctc\_cost} &\approx -\sum_{j,u}~E_{\P^\ast}\left[~\frac{\nabla_{\mathbf{\theta}}\,\beta^{(j)}_u}{\beta^{(j)}_u}~\frac{~\displaystyle\prod_{v=1}^V~\beta^{(j)}_v~\left(\hat{\varphi}^{(j)}_v\right)^{(1-V)/V}~}{~\displaystyle\sum_{i=1}^C~\left[\prod_{v=1}^V~\beta^{(i)}_v~\left(\hat{\varphi}^{(i)}_v\right)^{(1-V)/V}\right]~}~\right]\\
 &\qquad\qquad -\sum_{j,u}~\frac{E_{\P^\ast}\left[\nabla_{\mathbf{\theta}}\,\beta^{(j)}_u\right]}{\hat{\varphi}^{(j)}_u}~\times~\texttt{aux}^{(j)}\\
 &\qquad\qquad +\sum_{j,u}~\frac{E_{\P^\ast}\left[\nabla_{\mathbf{\theta}}\,\beta^{(j)}_u\right]}{\hat{\varphi}^{(j)}_u}~\frac{~\displaystyle\prod_{v=1}^V\left(\hat{\varphi}^{(j)}_v\right)^{1/V}~}{~\displaystyle\sum_{i=1}^C~\left[\prod_{v=1}^V\left(\hat{\varphi}^{(i)}_v\right)^{1/V}\right]~}\,, \label{eq:sgd}
\end{split}
\end{align}
where $\hat{\varphi}^{(i)}_v$ represents a moving estimate of $E_{\P^\ast}\left[\beta^{(i)}_v\right]$ maintained throughout the network training process and $\texttt{aux}^{(j)}$ represents a moving estimate of
\begin{equation}
 E_{\P^\ast}\left[~\frac{~\displaystyle\prod_{v=1}^V~\beta^{(j)}_v~\left(E_{\P^\ast}\left[\beta^{(j)}_v\right]\right)^{(1-V)/V}~}{~\displaystyle\sum_{i=1}^C~\left[\prod_{v=1}^V~\beta^{(i)}_v~\left(E_{\P^\ast}\left[\beta^{(i)}_v\right]\right)^{(1-V)/V}\right]~}~\right]\, .
\end{equation}
Maintaining the moving estimates $\hat{\varphi}^{(i)}_v$ and $\texttt{aux}^{(j)}$ is comparable to maintaining a discriminator in the training of a Generative Adversarial Network (GAN). The discriminator can be used to evaluate (and improve) the generator using mini-batches of data, instead of evaluating the (gradient of the) statistical distance between the training dataset and the GAN-dataset from scratch at every training step. Likewise, $\hat{\varphi}^{(i)}_v$ and $\texttt{aux}^{(j)}$ faciliate the use of stochastic or mini-batch gradient descent for the $\texttt{neg\_ctc\_cost}$ using \eqref{eq:sgd}---all the expectations in that expression are amenable to replacement with stochastic or mini-batch estimates. We note that this strategy was not needed for the studies performed in this paper.

The strategy employed in this section to facilitate the use to stochastic gradient descent for optimizing $\texttt{neg\_ctc\_cost}$ is applicable to a number of cost functions which cannot be written as expectations of per-datapoint loss functions. We will expand on this idea in future publications, and may implement it in future versions of \raindances{}.

  \bibliography{references}

\end{document}